\documentclass[review]{elsarticle}
\usepackage{lineno,hyperref}
\modulolinenumbers[5]

\usepackage{amsthm}
\usepackage{amsmath}
\usepackage{dsfont}
\usepackage{amssymb}
\usepackage{graphicx,color}
\usepackage{extarrows}
\usepackage{enumitem}
\usepackage{mathrsfs}
\usepackage{subcaption}
\usepackage{booktabs}
\usepackage{xcolor} 
\usepackage{float}
\usepackage{multirow} 
\usepackage{threeparttable}
\usepackage[justification=raggedright]{caption}
\usepackage{threeparttable}
\usepackage[misc]{ifsym}

\newtheorem{thm}{Theorem}[section]
\newtheorem{lem}[thm]{Lemma}

\newtheorem{assumption}{Assumption}[section]

\makeatletter
\@addtoreset{equation}{section}
\makeatother

\biboptions{numbers,sort&compress}

\usepackage{xcolor}

\begin{document}
\begin{frontmatter}

\title{\bf\Large Reconstruction and Prediction of Volterra Integral Equations Driven by Gaussian Noise}

\author{Zhihao Xu \fnref{addr1}}
\author{Saisai Ding \fnref{addr1}}
\author{Zhikun Zhang \fnref{addr1}}
\author{Xiangjun Wang \fnref{addr1}\corref{mycorrespondingauthor}}
\cortext[mycorrespondingauthor]{Corresponding author}
\ead{xjwang@hust.edu.cn}

\address[addr1]{School of Mathematics and Statistics, \\ Huazhong University of Science and Technology, Wuhan 430074, China}

\begin{abstract}
Integral equations are widely used in fields such as applied modeling, medical imaging, and system identification, providing a powerful framework for solving deterministic problems. While parameter identification for differential equations has been extensively studied, the focus on integral equations, particularly stochastic Volterra integral equations, remains limited. This research addresses the parameter identification problem, also known as the equation reconstruction problem, in Volterra integral equations driven by Gaussian noise. We propose an improved deep neural networks framework for estimating unknown parameters in the drift term of these equations. The network represents the primary variables and their integrals, enhancing parameter estimation accuracy by incorporating inter-output relationships into the loss function. Additionally, the framework extends beyond parameter identification to predict the system’s behavior outside the integration interval. Prediction accuracy is validated by comparing predicted and true trajectories using a 95\% confidence interval. Numerical experiments demonstrate the effectiveness of the proposed deep neural networks framework in both parameter identification and prediction tasks, showing robust performance under varying noise levels and providing accurate solutions for modeling stochastic systems.

\end{abstract}


\end{frontmatter}


\section{Introduction}
Integral equations, as a crucial branch of modern mathematics, are widely employed in fields such as applied modeling \cite{ahmed2015parameter, fermo2016scattering}, medical imaging \cite{mohammadian2018integral}, and system identification \cite{song2005integral}, offering a powerful framework for solving deterministic problems. While much research has been devoted to parameter identification for differential equations \cite{li2024parameter}, the study of parameter identification for integral equations has received comparatively less attention. This gap in the literature motivates our focus on the parameter identification problem for stochastic integral equations. The challenge we address is particularly general, as we explore the identification of parameters in stochastic integral equations driven by Gaussian noise. These stochastic integral equations arise in various real-world applications, such as biological populations, metabolic systems, medical chemotherapy, genetic mechanisms, and communication systems, where uncertainty plays a key role. In contrast to deterministic models, traditional integral equations are not sufficient to describe these phenomena, leading to the need for stochastic integral equations. Among these, stochastic Volterra integral equations \cite{li2022numerical} have found applications in fields like mathematical finance, physics, and biology. Since Berger \cite{ref4} first examined stochastic Volterra equations, this form has garnered significant attention from researchers. 
\begin{equation}
    X_t = x + \int_0^t b(t,s,X_s)\:\mathrm{d}s + \int_0^t \sigma(t,s,X_s)\:\mathrm{d}B_s, \qquad t \geq 0,
\end{equation}
where $x\in \mathbb{R}^d$, $b$ and $\sigma$ are Borel measurable functions satisfying certain conditions, $\{B_t,t \geq 0\}$ is $r$--dimensional standard Brownian motion. The Japanese mathematician Itô \cite{17} introduced the stochastic It\^o-Volterra integral equation, the solution of which is a Markov process, in the study of diffusion problems. \par
Volterra integral equations have been extensively studied, yielding significant results. Burton \cite{dingxing2} provides a comprehensive synthesis of the relevant theory, while Islam \cite{dingxing3} extends results from linear to nonlinear equations, examining conditions for bounded, nonnegative, and integrable solutions. Pedas \cite{shuzhi8} introduced the smooth transformation method for second-kind equations with weakly singular kernels. Spectral collocation methods \cite{shuzhi16,shuzhi17} have been employed to solve weakly singular Volterra equations. \par
To address the parameter identification problem, various methods are commonly employed, including two-stage methods \cite{canshu1}, nonlinear least squares \cite{canshu2}, mixed effects models \cite{canshu3}, and machine learning techniques \cite{canshu4}. The ultimate goal of these methods is to solve the equation reconstruction problem, where accurate parameter identification is crucial for obtaining reliable solutions. In engineering, parameters and internal variables are often unknown or unmeasured, and sensor signals are frequently distorted by noise. To extract useful information, accurate parameter estimation is essential. To enhance parameter estimation from noisy data and improve the sensitivity of numerical differentiation, Boulier \cite{boulier2014algorithm} introduces algorithms that convert differential equations into integral equations. \par
Deep learning, particularly deep neural networks, has become a powerful tool in solving complex mathematical equation problems. One important approach is to integrate physical laws with the neural network framework, which has proven to be effective in solving partial differential equations (PDEs) \cite{pinn2} and integral equations \cite{pinn20}. In addition to this framework, advancements in deep neural networks, such as specialized activation functions, adaptive weights, and uncertainty quantification, have also been explored to address complex problems, such as fractional differential equations and integro-differential equations \cite{pinn28,pinn29,pinn32}. Physics-informed neural networks (PINNs) \cite{pinn1} have found applications in various fields, including fluid mechanics \cite{pinn4}, medical diagnosis \cite{pinn9}, and heat transfer analysis \cite{pinn10,pinn11}. In this context, our method adopts a neural network architecture inspired by PINNs, the improved deep neural networks (DNNs). However, our focus is on leveraging the technique as DNNs more broadly to enhance the accuracy and efficiency of the solution process. By incorporating physical information directly into the training process, we improve the approximation of the solution to the data-driven parameter identification problem of stochastic integral equations. The network utilizes automatic differentiation \cite{pinn3} to compute the derivatives required in the governing equations and employs optimization methods to minimize the residuals from the governing equations, measurement data, and output conditions.

In this study, we address the challenge of learning the drift term coefficients in stochastic integral equations in the presence of perturbation noise. Unlike deterministic scenarios, where the dynamics can be directly modeled with a specific integral equation, the incorporation of Brownian motion introduces inherent randomness into the system. To mitigate the impact of this noise, we simulate a large number of trajectories to capture the stochastic behavior of the system. By taking the expectation of these simulated paths, we obtain a more representative and smooth trajectory that reflects the underlying dynamics while effectively filtering out the noise \cite{pinn34}. This averaged trajectory serves as a critical reference point for training the improved DNNs. This approach contrasts with deterministic cases, where the parameters can be directly inferred from a single, noise-free trajectory. By leveraging the expectation of multiple trajectories, we aim to enhance the robustness and accuracy of our parameter identification in the stochastic context. 

In addition to parameter identification for reconstructing the equations, we also focus on predicting the equation’s evolution beyond the integration interval. By forecasting the system’s future behavior, we further validate the accuracy of the reconstructed equation. This approach contrasts with deterministic scenarios, where a single, noise-free trajectory suffices for parameter inference. By utilizing multiple trajectories, our method enhances both the robustness and accuracy of parameter identification and prediction in the stochastic context.\par

The paper is organized as follows. In Section \ref{sec:problem setup}, we describe the general form of the parameter identification problem of the Volterra integral equation disturbed by Gaussian noise. In Section \ref{sec:method}, we propose the improved DNNs framework. In Section \ref{sec:experiment}, we present the results of the improved DNNs for solving the parameter identification problem of Volterra integral equations disturbed by Gaussian noise, with both linear and nonlinear kernel functions. Additionally, we predict the evolution of the equations beyond the integration interval and present the corresponding experimental results. Finally, we conclude with a summary and discussion.

\section{Problem Setup}
\label{sec:problem setup}
Let $(\Omega,\mathcal{F},\mathfrak{F},\mathbb{P})$  be a complete filtered probability space satisfying the usual conditions on which a one-dimensional standard Brownian motion $B(\cdot)$ is defined with $\mathfrak{F}=\{\mathcal{F}_t\}_{t \geq 0}$ being its natural filtration augmented by all the $\mathbb{P}$--null sets. Suppose $0 \leq S < T$ and $f(t,\omega)$ is given, where $f:[0,\infty) \times \Omega \rightarrow \mathbb{R}$, we want to define
\begin{equation*}
    \int_S^T f(t,\omega) \:\mathrm{d}B_t(\omega).
\end{equation*}
Referring to Oksendal \cite{oksendal2013stochastic}, we define the class of functions $\mathcal{V} = \mathcal{V}(S, T)$ to consist of all functions 
$f: [0, \infty] \times \Omega \rightarrow \mathbb{R}$ that satisfy the following conditions:
\begin{enumerate}
    \item $(t, \omega) \rightarrow f(t, \omega)$ is $\mathcal{B} \times \mathcal{F}$--measurable, where $\mathcal{B}$ denotes the Borel $\sigma$--algebra on $[0, \infty)$.
    
    \item $f(t, \omega)$ is $\mathcal{F}_t$--adapted.
    
    \item $E\left[\int_S^T f(t, \omega)^2 \, \mathrm{d}t\right] < \infty$.
\end{enumerate}
For functions $f \in \mathcal{V}$, we will now show how to define the It\^o integral. Let $f \in \mathcal{V}(S,T)$, then the It\^o integral of $f$ (from $S$ to $T$) is defined by
\[
\int_S^T f(t,\omega) \:\mathrm{d}B_t(\omega) = \lim_{n \to \infty} \int_S^T \varphi_n(t,\omega) \:\mathrm{d}B_t(\omega),
\]
where $\{\varphi_n\}$ is a sequence of simple functions such that
\[
E\left[\int_S^T (f(t,\omega)-\varphi_n(t,\omega))^2 \:\mathrm{d}t\right] \rightarrow 0 \quad \text{as} \quad n \rightarrow \infty.
\]
An important property of the It\^o integral is that it is a martingale. First, we define a filtration (on $(\Omega,\mathcal{F})$) as a family $\mathcal{M} = \{\mathcal{M}_t\}_{t \geq 0}$ of $\sigma$--algebras $\mathcal{M}_t \subset \mathcal{F}$ such that for any $0 \leq s < t$, we have $\mathcal{M}_s \subset \mathcal{M}_t$ (i.e., $\{\mathcal{M}_t\}$ is increasing). Next, we say that an $n$--dimensional stochastic process $\{M_t\}_{t \geq 0}$ on $(\Omega,\mathcal{F},\mathbb{P})$ is a martingale with respect to a filtration $\{\mathcal{M}_t\}_{t \geq 0}$ if it satisfies the following conditions:
\begin{enumerate}
    \item $M_t$ is $\mathcal{M}_t$--measurable for all $t$.
    
    \item $E[\|M_t\|] < \infty$ for all $t$.
    
    \item $E[M_s | \mathcal{M}_t] = M_t$ for all $s \geq t$.
\end{enumerate}

After introducing the It\^o integrals, we present the form of stochastic differential equations based on ordinary differential equations
\begin{equation}
    \:\mathrm{d}X_t = b(t,X_t)\:\mathrm{d}t + \sigma(t,X_t)\:\mathrm{d}B_t,
\end{equation}
or in integral form
\begin{equation}
    X_t = X_0 + \int_0^t b(s,X_s)\:\mathrm{d}s + \int_0^t \sigma(s,X_s)\:\mathrm{d}B_s, \qquad t\in [0,T]\,.
\end{equation}

Consider the following Volterra integral equation disturbed by Gaussian noise:
\begin{equation}
    X(t) = f(t) + \int_{0}^{t}k(t,s;\theta)X(s)\:\mathrm{d}s + \int_{0}^{t}h(t,s;\lambda)\:\mathrm{d}B_s, \qquad t\in [0,T],
    \label{disturbed}
\end{equation}
where $f:[0,T] \rightarrow \mathbb{R}$ is a continuous function; $\theta$ and $\lambda$ are real constant parameters; $k(t,s;\theta)$ and $h(t,s;\lambda)$ are kernel functions defined on the domain $S=\{(t,s):0 \leq s \leq t \leq T\}$, modified by the constants $\theta$ and $\lambda$, respectively.



By a solution of Eq.\,\ref{disturbed} we mean an $\mathcal F_t$--adapted stochastic process
$X=\{X(t)\}_{t\in[0,T]}$ with continuous sample paths a.s., satisfying Eq.\,\ref{disturbed} for every
$t\in[0,T]$ and having finite second moment, i.e.
\[
  \sup_{t\in[0,T]}\mathbb{E}\big[|X(t)|^2\big]<\infty.
\]


We state below a set of standard sufficient conditions (in the spirit of \cite{ito1979existence,jin2022existence})
under which the Volterra integral equation Eq.\,\ref{disturbed} is well posed.

\begin{enumerate}[label=(A\arabic*)]
  \item \textbf{Measurability / adaptedness.} For each fixed $(t,s)\in S$ the mappings
    $(t,s,x)\mapsto k(t,s;\theta)x$ and $(t,s)\mapsto h(t,s;\lambda)$ are measurable; moreover for each $s$
    the function $h(t,s;\lambda)$ is $\mathcal F_s$--adapted.
  \item \textbf{Lipschitz in the state.} There exists $L>0$ such that for a.e.\ $(t,s)\in S$ and all $x,y\in\mathbb R$
    \[
      |k(t,s;\theta)x-k(t,s;\theta)y|\le L\,|x-y|.
    \]
  \item \textbf{Linear growth / boundedness.} There exists $K>0$ such that for all $(t,s,x)\in S\times\mathbb R$
    \[
      |k(t,s;\theta)x|\le K(1+|x|),\qquad |h(t,s;\lambda)|\le K.
    \]
  \item \textbf{Integrability (well-posedness of integrals).} The kernels satisfy the integrability conditions
    \[
      \int_0^T\sup_{t\in[0,T]}|k(t,s;\theta)|\,\mathrm{d}s<\infty,\qquad
      \int_0^T\sup_{t\in[0,T]}|h(t,s;\lambda)|^2\,\mathrm{d}s<\infty,
    \]
    so that the deterministic integral and the It\^o integral in Eq.\,\ref{disturbed} are well defined in the mean-square sense.
\end{enumerate}

\begin{lem}
\label{lem:exist-uniq}
Assume (A1)--(A4) above. Then Eq.\,\ref{disturbed} admits a unique $\mathcal F_t$--adapted solution
$X=\{X(t)\}_{t\in[0,T]}$ with continuous sample paths a.s.\ and $\sup_{t\in[0,T]}\mathbb{E}|X(t)|^2<\infty$.
\end{lem}

\paragraph{Remark}
The result in Lemma~\ref{lem:exist-uniq} follows from the standard Picard successive-approximation argument
for stochastic Volterra equations together with It\^o isometry, suitable maximal inequalities and a Gronwall-type
estimate. For detailed formulations and complete estimates, we refer the reader to It\^o \cite{ito1979existence} and Jin \cite{jin2022existence}, and omit the lengthy technical details here for brevity.


We aim to investigate the parameter identification problem for the drift term in the disturbed Volterra integral equation, in which the disturbed Eq.\,\ref{disturbed} is definite and boundary and initial conditions may be known or not, in the governing equation, the parameter $\theta$ is unknown while the other parameter $\lambda$ is given. We will use some measurement data to discover the unknown parameter $\theta$.

\section{Methodology}
\label{sec:method}
\subsection{Numerical approximation method}
To better solve Eq.\,\ref{disturbed} with generality, we adopt numerical approximation methods. The finite difference method (FDM) \cite{fdm} is a simple and intuitive approach for solving Volterra integral equations. We divide the interval $[0,T]$ into $n$ equal parts, each of length $\Delta t=\frac{T}{n}$. To simplify the expression, we set the Itô integral term to the most common case, namely $\lambda \int_{0}^{t}\:\mathrm{d}B_s$, and let
\begin{align*}
    k(i\Delta t,j\Delta t;\theta)&=k_{ij},\\
    X(j\Delta t)&=X_j, 
\end{align*}
where $0 \leq j < i \leq n$.
Then the integral term in Eq.\,\ref{disturbed} can be approximated by an algebraic sum $\Sigma_{i=1}^{n}\Sigma_{j=0}^{i-1}k_{ij}X_j\Delta t$. Here, we use the left endpoint of each subinterval for the approximation when discretizing the integral. The detailed approximation process is as follows:
\begin{equation}
  \begin{aligned}
    X(i\Delta t) &= f(i\Delta t) + \int_0^{i\Delta t} k(i\Delta t, s;\theta)X(s)\:\mathrm{d}s + \lambda B_{i\Delta t} \\
    &\approx f(i\Delta t) + k(i\Delta t, 0;\theta)X(0)\Delta t \\
    &+ k(i\Delta t, \Delta t;\theta)X(\Delta t)\Delta t  \\
    &+ \dots \\
    &+ k(i\Delta t, (i-1)\Delta t;\theta)X((i-1)\Delta t)\Delta t \\
    &+ \lambda B_{i\Delta t}, \qquad i=1,2,\dots,n.\\
\end{aligned}
\end{equation}

In summary, we have outlined a numerical generation method utilizing the finite difference method to approximate solutions for the Volterra integral equations. By discretizing the interval and employing an algebraic sum to represent the integral terms, we have established a systematic approach to iteratively compute $X(i\Delta t)$, and $X(T)$ can be approximated as
\begin{equation}
    X(T)\approx f(T) + \sum_{i=1}^{n}\sum_{j=0}^{i-1}k_{ij}X_j\Delta t+ \lambda B_T.
\end{equation}
This method not only enhances computational efficiency but also provides a flexible framework for addressing various scenarios dictated by the kernel function $k(t,s;\theta)$. Ultimately, this technique serves as a robust tool for further exploration of the integral equations at hand.

\subsection{The improved DNNs framework}
\label{sec:PINN}
DNNs can combine deep learning with physical laws, integrating physical information directly into the neural network training process to more effectively approximate the exact solutions of differential equations. It has been proven that DNNs can address the inverse problems of both linear and nonlinear PDEs \cite{pinn1}. Consider a PDE with Dirichlet boundary condition, given by
\begin{equation}
    \begin{aligned}
        &\frac{\partial ^n u(x,t;\theta)}{\partial x^n}+f(x,t)=0, \\
        &u(0,t)=\varphi (0,t), \\
        &u(x,0)=\psi (x,0),
    \end{aligned}
    \label{example}
\end{equation}
where $(x,t)$ is the input of the neural network. In inverse problems, discovering unknown parameters $\theta$ from measurement data, and precise initial and boundary conditions can be challenging to determine; typically, we only have the governing equation and measurement data. In these instances, the physical information comprises two components: the residuals of the governing equation and the residuals of the measurement data. So the loss function $\mathrm{MSE}_{total}$ is formulated as:
\begin{equation}
    \mathrm{MSE}_{total} = \mathrm{MSE}_m + \mathrm{MSE}_g\,,
\end{equation}
where
\begin{equation*}
    \mathrm{MSE}_m=\frac{1}{N_m}\sum_{i=1}^{N_m}\left|u_{pred}(x_i^m,t_i^m;\gamma,\theta)-u_m(x_i^m,t_i^m)\right|^2 \,,
\end{equation*}
\begin{equation*}
    \mathrm{MSE}_g=\frac{1}{N_g}\sum_{i=1}^{N_g}\left|\frac{\partial ^n u_{pred}(x_i^g,t_i^g;\gamma,\theta)}{\partial x^n}+f(x_i^g,t_i^g)\right|^2\,.
\end{equation*}
$\mathrm{MSE}_m$ and $\mathrm{MSE}_g$ represent the mean square errors of the measurement data residuals and the governing equation residuals, respectively, with $N_m$ denoting the number of measurement data points, $N_g$ denoting the number of collocation points. $u_{pred}(x_i^m,t_i^m;\gamma,\theta)$ and $u_m(x_i^m,t_i^m)$ are the predicted and measured values at the measurement points $(x_i^m,t_i^m)$. Here, $\gamma$ denotes the DNN parameters, including weights $\mathbf{W}$ and biases $\mathbf{b}$, while $\theta$ represents the unknown parameters in the PDEs.

\par

\begin{figure}[t]
    \centering
    \includegraphics[width=1\textwidth]{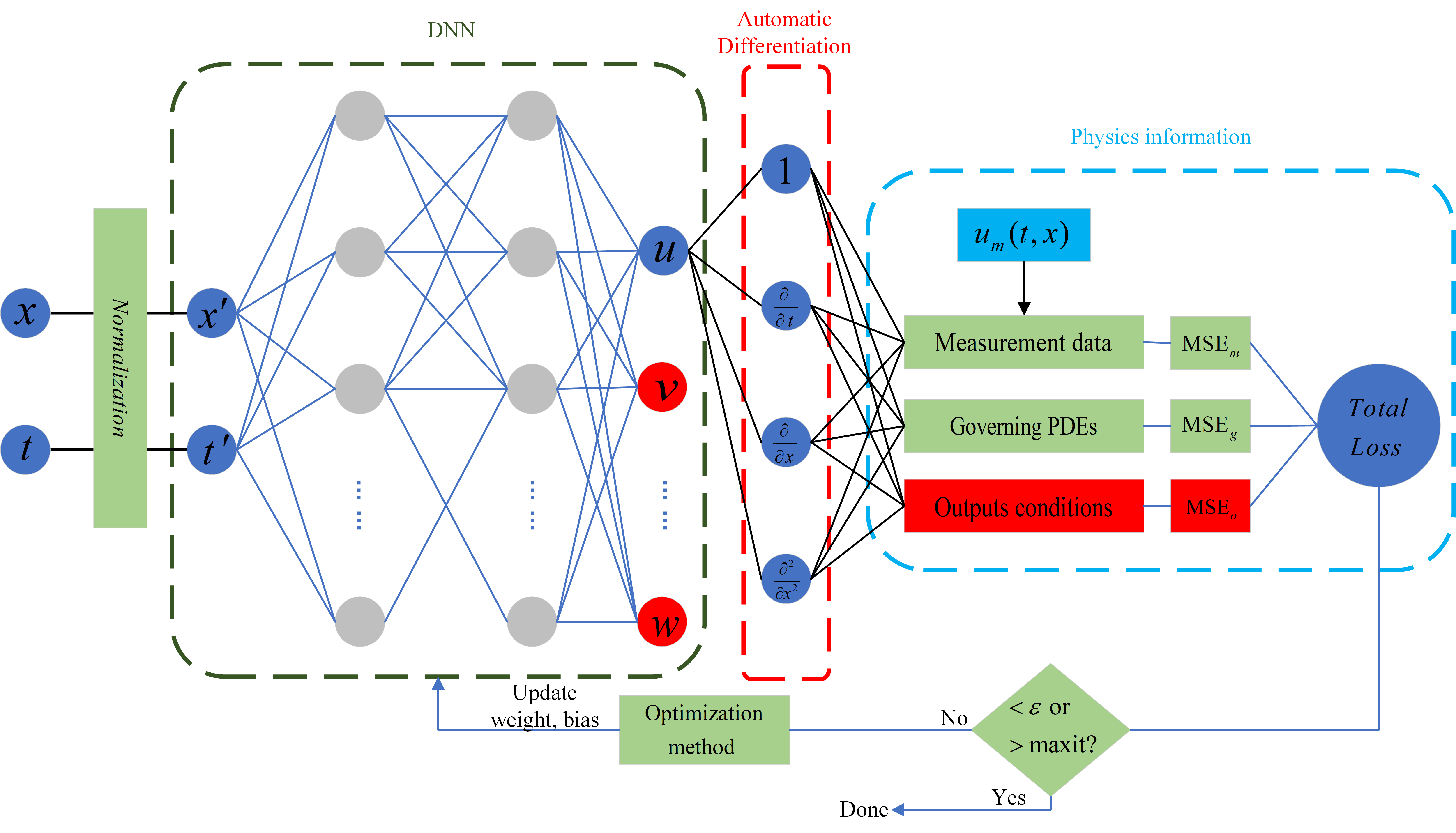}
    \caption{The framework of improved DNNs for solving the inverse problem of Volterra integral equation disturbed by Gaussian noise.}
    \label{improvedPINN}
\end{figure}

After introducing the basic framework, we propose an improved neural network framework to address the parameter identification problem of the Volterra integral equation disturbed by Gaussian noise. The framework is illustrated in Fig.\,\ref{improvedPINN}. We refer to Yuan \cite{pinn20}, utilizing a multi-output DNN to simultaneously compute the primary and auxiliary outputs which respectively represent the variables and integrals involved in the governing equation. Subsequently, the relationship between primary and auxiliary outputs is established by enforcing additional output conditions that adhere to physical laws. This can also be understood as adding extra physical law conditions to the physics-informed module in the typical DNNs framework. The following focuses on two representative cases of Eq.\,\ref{disturbed} under consideration: 
\begin{equation}
    X(t) = f(t) + \theta\int_{0}^{t}k(t,s)X(s)\:\mathrm{d}s + \lambda\int_{0}^{t}\:\mathrm{d}B_s, \qquad t\in [0,T],
    \label{scase1}
\end{equation}
\begin{equation}
    X(t) = f(t) + \theta\int_{0}^{t}k(t,s)X(s)\:\mathrm{d}s + \lambda\int_{0}^{t}B_s\:\mathrm{d}B_s, \qquad t\in [0,T],
    \label{scase2}
\end{equation}
the equations can also be equivalently expressed in the following form:
\begin{equation}
    X(t) = f(t) + \theta\int_{0}^{t}k(t,s)X(s)\:\mathrm{d}s + \lambda B_t, \qquad t\in [0,T],
    \label{simple}
\end{equation}
\begin{equation}
    X(t) = f(t) + \theta\int_{0}^{t}k(t,s)X(s)\:\mathrm{d}s + \lambda (\frac{1}{2}{B_t}^2-\frac{1}{2}t), \qquad t\in [0,T].
\end{equation}
For understanding, the notation used in the previous example is employed to describe the equation, defining the primary output $u(t)$ to represent $X(t)$ and the auxiliary output $v(t)$  to represent the integral in Eq.\,\ref{simple}. Thus, the equation can be transformed into
\begin{equation}
    \begin{aligned}
        &u(t) = f(t) + \theta\cdot v(t) + \lambda B_t, \qquad t\in [0,T], \\
        &v(t)=\int_{0}^{t}k(t,s)X(s)\:\mathrm{d}s, \\
        &u(0)=f(0).
    \end{aligned}
    \label{ezone}
\end{equation}
\par
For this problem, the variable of the equation is $t$. In addressing parameter identification problems, which involve discovering unknown parameters in governing equations with measurement data, the DNN is configured to calculate the predicted value $u_{pred}(t_i;\gamma,\theta)$ and the auxiliary output $v_{pred}(t_i;\gamma,\theta)$, which approximate the true value $u_{true}(t_i)$ and the integral $\int_{0}^{t_i}k(t,s)X(s)\:\mathrm{d}s$, respectively, where $\gamma$ is the parameters of the DNN and $\theta$ is the unknown parameter in the governing equation. To simplify programming and avoid integral manipulation, the relationship between $u(t)$ and $v(t)$ is reformulated as
\begin{equation}
    \begin{aligned}
        &\frac{\mathrm{d}v(t)}{\mathrm{d}t} = k(t,t)u(t)+ \int_{0}^{t}\frac{\partial k(t,s)}{\partial t}u(s)\:\mathrm{d}s,\\
        &v(0)=0.
    \end{aligned}
    \label{eztwo}
\end{equation}

A new output condition Eq.\,\ref{eztwo} is introduced as a physical law to constrain the relationship between the auxiliary output $v$ and the primary output $u$. The mean square error of the residuals for this new output condition is calculated as follows:
\begin{equation}
    \mathrm{MSE}_o=\frac{1}{N_g}\sum_{i=1}^{N_g}\left|\frac{\partial v_{pred}(t_i^g;\gamma,\theta)}{\partial t}-k(t_i^g,t_i^g)u_{pred}(t_i^g;\gamma,\theta)-\int_{0}^{t_i^g}\frac{\partial k(t_i^g,s)}{\partial t}u_{pred}(s;\gamma,\theta)\:\mathrm{d}s\right|^2 \,,
\end{equation}
in which $N_g$ is the number of collocation points sampled in the equation domain. The mean square error of residuals of the measurement data is formulated by
\begin{equation}
    \mathrm{MSE}_m=\frac{1}{N_m}\sum_{i=1}^{N_m}\left|u_{pred}(t_i^m;\gamma,\theta)- u_m(t_i^m)\right|^2,
\end{equation}
where $N_m$ is the number of measurement data. The mean square error of residuals of the governing equation is formulated by
\begin{equation}
    \mathrm{MSE}_g=\frac{1}{N_g}\sum_{i=1}^{N_g}\left|u_{pred}(t_i^g;\gamma,\theta)-E[f(t_i^g)+\theta\cdot v_{pred}(t_i^g;\gamma,\theta) + \lambda B_{t_i^g}]\right|^2\,,
    \label{mseg}
\end{equation}
where $E$ is the expectation operator, by averaging the simulated paths, we obtain a smoother and more representative trajectory that captures the underlying dynamics while effectively filtering out the noise, this averaged trajectory serves as an essential reference for training the improved DNNs, notice that $E(B_t)=0$ and $E(\frac{1}{2}{B_t}^2-\frac{1}{2}t)=0$. 
The total loss in the improved DNNs framework is represented as a weighted sum of the individual mean-squared errors:
\begin{equation}
    \mathrm{MSE}_{\mathrm{total}} = w_m\cdot \mathrm{MSE}_m + w_g\cdot \mathrm{MSE}_g + w_o\cdot \mathrm{MSE}_o\,.
\end{equation}

Following Yuan \cite{pinn20}, we adopt an adaptive weighting strategy to balance the different residuals automatically. In our implementation we use the following stable form of the adaptive weights:
\begin{equation}
    (w_m, w_g, w_o)
    =\Big( 1\,,\; \frac{\mathrm{MSE}_g}{\min\{\mathrm{MSE}_g,\mathrm{MSE}_o\}+\varepsilon_e},\,\;
      \frac{\mathrm{MSE}_o}{\min\{\mathrm{MSE}_g,\mathrm{MSE}_o\}+\varepsilon_e}\, \Big)\,,
\end{equation}
where \(\varepsilon_e>0\) is a small regularization constant introduced to prevent division by zero or excessively large weight values when one residual is very small. We set \(\varepsilon_e=10^{-8}\) in our experiments; this value is several orders of magnitude smaller than typical residuals and therefore does not materially change the balancing behaviour while eliminating instability. We set \(w_m=1\) because, in the presence of measurement noise, the measurement loss \(\mathrm{MSE}_m\) generally does not admit a zero minimizer and should therefore be treated differently from the physics residuals \(\mathrm{MSE}_g\) and \(\mathrm{MSE}_o\), which have zero as their ideal targets. Alternative stabilisation strategies include clipping, moving-average smoothing, softmax normalization, or trainable weights.

\par
Now we present a rigorous convergence analysis for our improved DNNs framework. The proof leverages recent advances in PINNs theory while accounting for the integral nature of our problem and the multiple constraint terms.
\begin{assumption}[Problem regularity and well-posedness]\label{ass:wellposed}
Assume the Eq.\,\ref{disturbed} satisfies conditions (A1)--(A4) from Section \ref{sec:problem setup}, ensuring existence and uniqueness of the solution. Furthermore, assume the true solution \(u^\star\) and auxiliary function \(v^\star\) possess sufficient regularity, specifically \(u^\star, v^\star \in H^r(0,T)\) for some \(r > 0\).
\end{assumption}

\begin{assumption}[Neural network approximation capability]\label{ass:approx}
Let \(\mathcal{N}\) be the class of neural networks with architecture as described in Section \ref{sec:experiment}. There exists a constant \(C_{\text{app}} > 0\) such that for any \(\varepsilon > 0\), there exists \((\tilde{u}, \tilde{v}) \in \mathcal{N}\) with
\[
\| \tilde{u} - u^\star \|_{L^2(0,T)} + \| \tilde{v} - v^\star \|_{L^2(0,T)} \leq C_{\text{app}} \varepsilon.
\]
\end{assumption}

\begin{assumption}[Training success]\label{ass:training}
The training procedure produces parameters \(\theta^*\) such that the discrete training loss satisfies
\[
\mathcal{L}_{\text{train}}(\theta^*) = \mathrm{MSE}_m + \mathrm{MSE}_g + \mathrm{MSE}_o \leq \varepsilon_{\text{opt}}^2.
\]
\end{assumption}

\begin{thm}
\label{thm:convergence}
Under Assumptions \ref{ass:wellposed}--\ref{ass:training}, there exists a constant \(C > 0\) depending on the problem data (kernel properties, time horizon \(T\), and regularity constants) such that the trained network outputs satisfy:

\begin{enumerate}
    \item \textbf{Solution convergence:}
    \[
    \| u_{\theta^*} - u^\star \|_{L^2(0,T)} \leq C\left( \sqrt{\mathcal{L}_{\text{train}}(\theta^*)} + \varepsilon_{\text{app}} + \delta_{\text{quad}} \right)
    \]
    
    \item \textbf{Parameter convergence:} Under the identification condition that the mapping \(\theta \mapsto \int_0^t k(t,s;\theta)X(s)ds\) is injective,
    \[
    | \theta^* - \theta_{\text{true}} | \leq C_\theta \left( \sqrt{\mathcal{L}_{\text{train}}(\theta^*)} + \varepsilon_{\text{app}} + \delta_{\text{quad}} \right)
    \]
\end{enumerate}

where \(\varepsilon_{\text{app}}\) is the neural network approximation error, and \(\delta_{\text{quad}}\) quantifies the numerical integration error in the output condition.
\end{thm}

The proof, provided in \ref{app:proof_convergence}, adapts the conditional stability framework of \cite{mishra2022estimates} to integral equations and incorporates the multi-residual structure following \cite{qian2023physics}.


\section{Numerical Experiments}
\label{sec:experiment}
In this section, we conduct numerical experiments to solve both the parameter identification and prediction problems of Volterra integral equations disturbed by Gaussian noise using the improved DNNs. These experiments are designed to reflect the accuracy and efficiency of the framework in both tasks. The neural network architecture consists of 4 layers: an input layer, an output layer, and 2 hidden layers, each containing 40 neurons. For the activation function, we employ the differentiable nonlinear function $\tanh{(\cdot)}$. The optimization algorithm used is L-BFGS, with a learning rate of 0.01. All models are implemented using Python and the PyTorch library. We divide the interval of integration into $n=1000$ equal parts, each of length $\Delta t=\frac{T}{n}$.

First, we validate the performance of the model by identifying the unknown parameters in the drift term of Volterra integral equations. Next, we extend the time range beyond the reconstruction domain to perform predictions, allowing us to evaluate the model's predictive capabilities. This dual-task approach provides a comprehensive assessment of the framework’s robustness and practical applicability in modeling stochastic systems.

\begin{figure}[htbp]
    \centering
    \begin{minipage}{0.33\textwidth}
        \centering
        \begin{minipage}{\textwidth}
            \includegraphics[width=\textwidth]{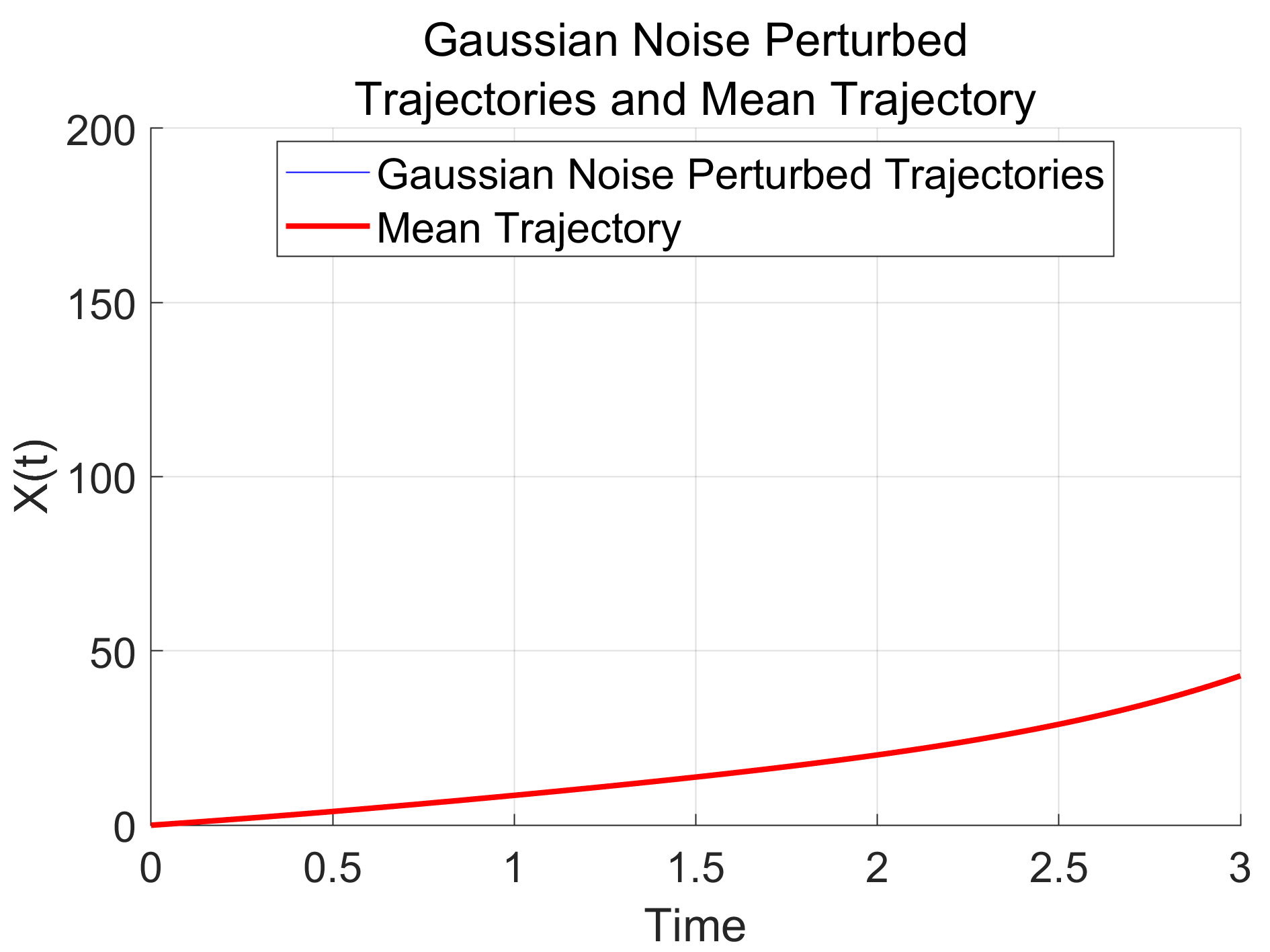}
            \caption*{(a1)$\lambda=0$}
        \end{minipage}\par
        \begin{minipage}{\textwidth}
            \includegraphics[width=\textwidth]{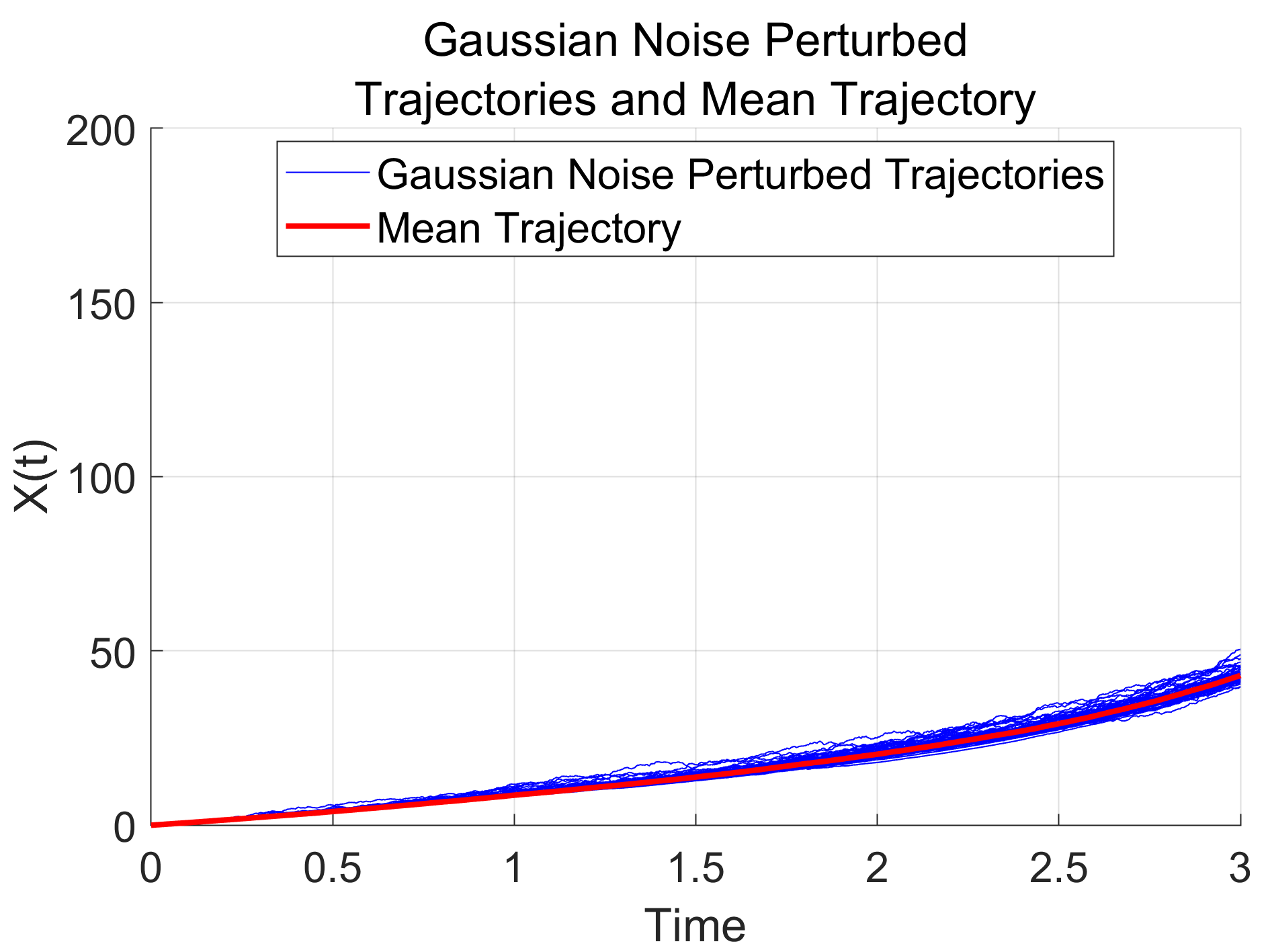}
            \caption*{(b1)$\lambda=1$}
        \end{minipage}\par
        \begin{minipage}{\textwidth}
            \includegraphics[width=\textwidth]{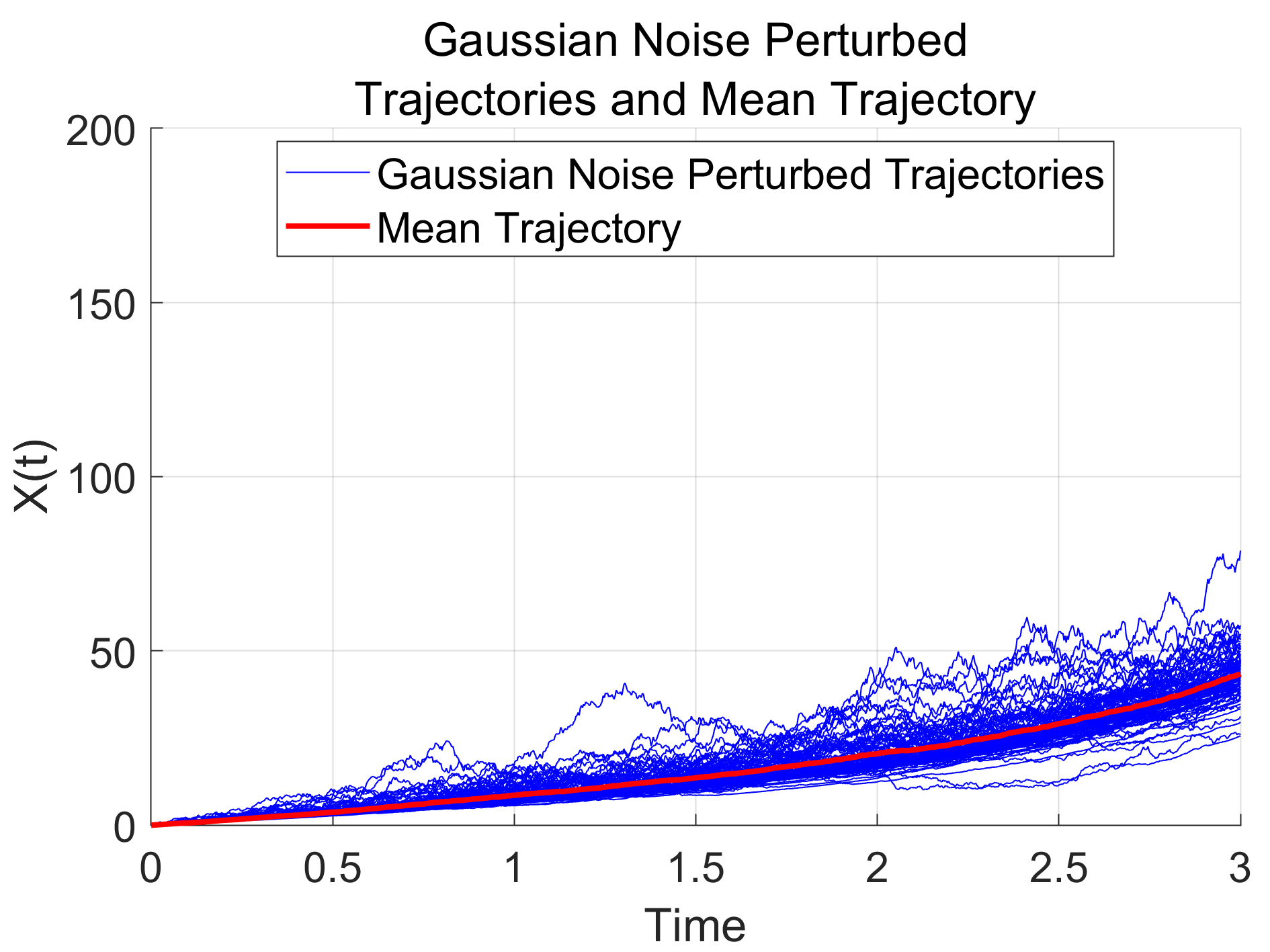}
            \caption*{(c1)$\lambda=5$}
        \end{minipage}\par
        \begin{minipage}{\textwidth}
            \includegraphics[width=\textwidth]{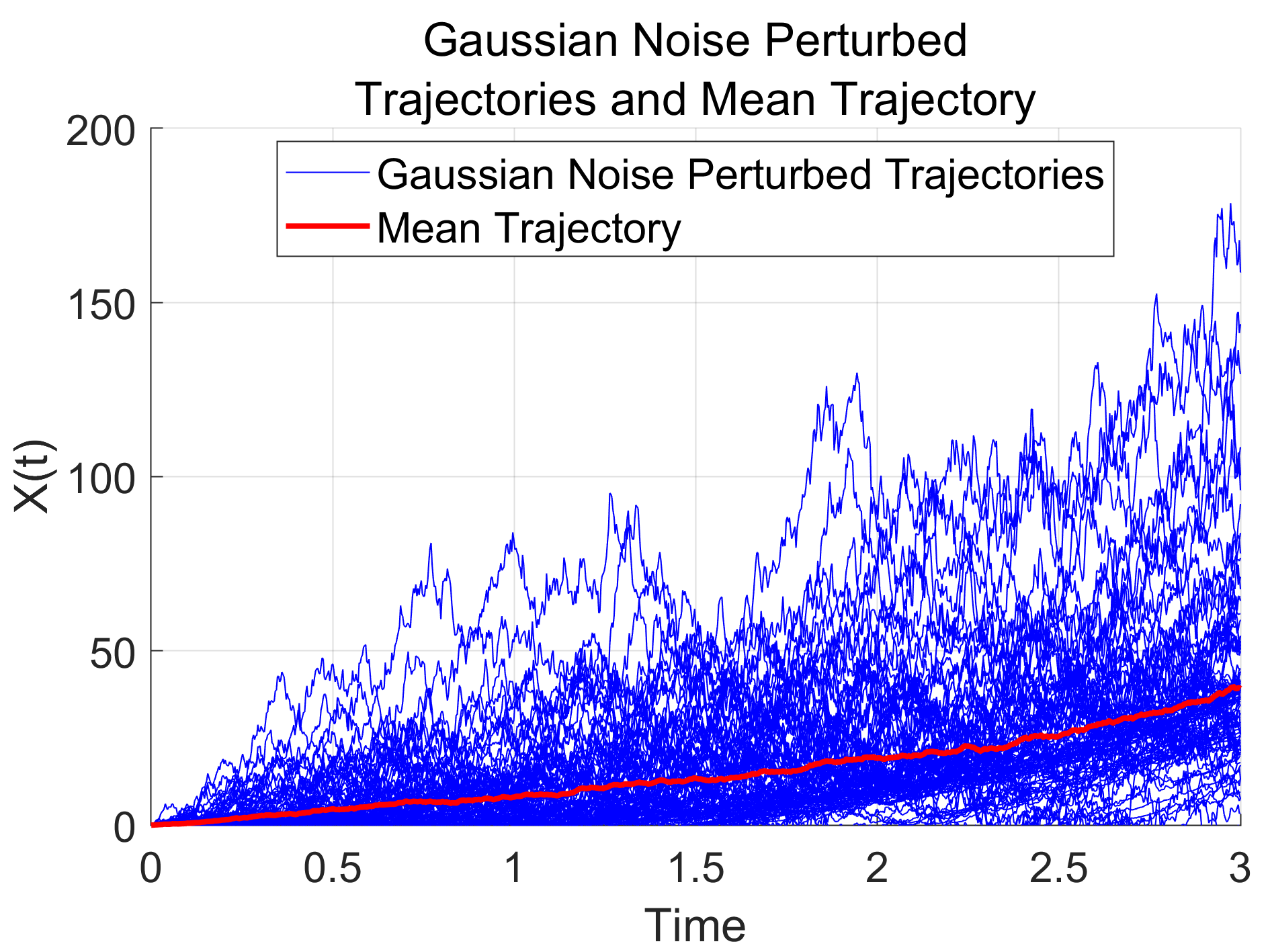}
            \caption*{(d1)$\lambda=20$}
        \end{minipage}
    \end{minipage}%
    \begin{minipage}{0.33\textwidth}
        \centering
        \begin{minipage}{\textwidth}
            \includegraphics[width=\textwidth]{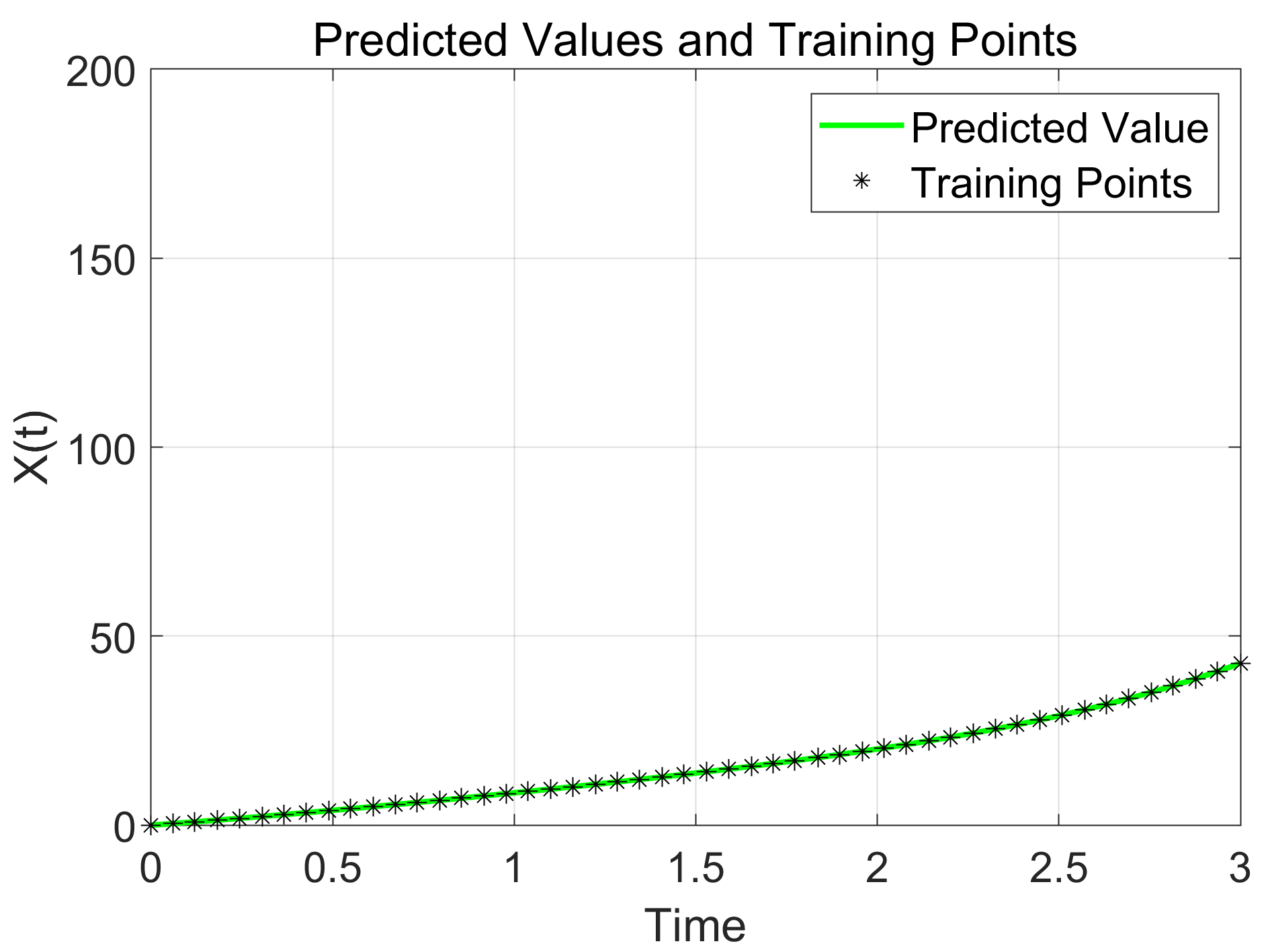}
            \caption*{(a2)$\lambda=0$}
        \end{minipage}\par
        \begin{minipage}{\textwidth}
            \includegraphics[width=\textwidth]{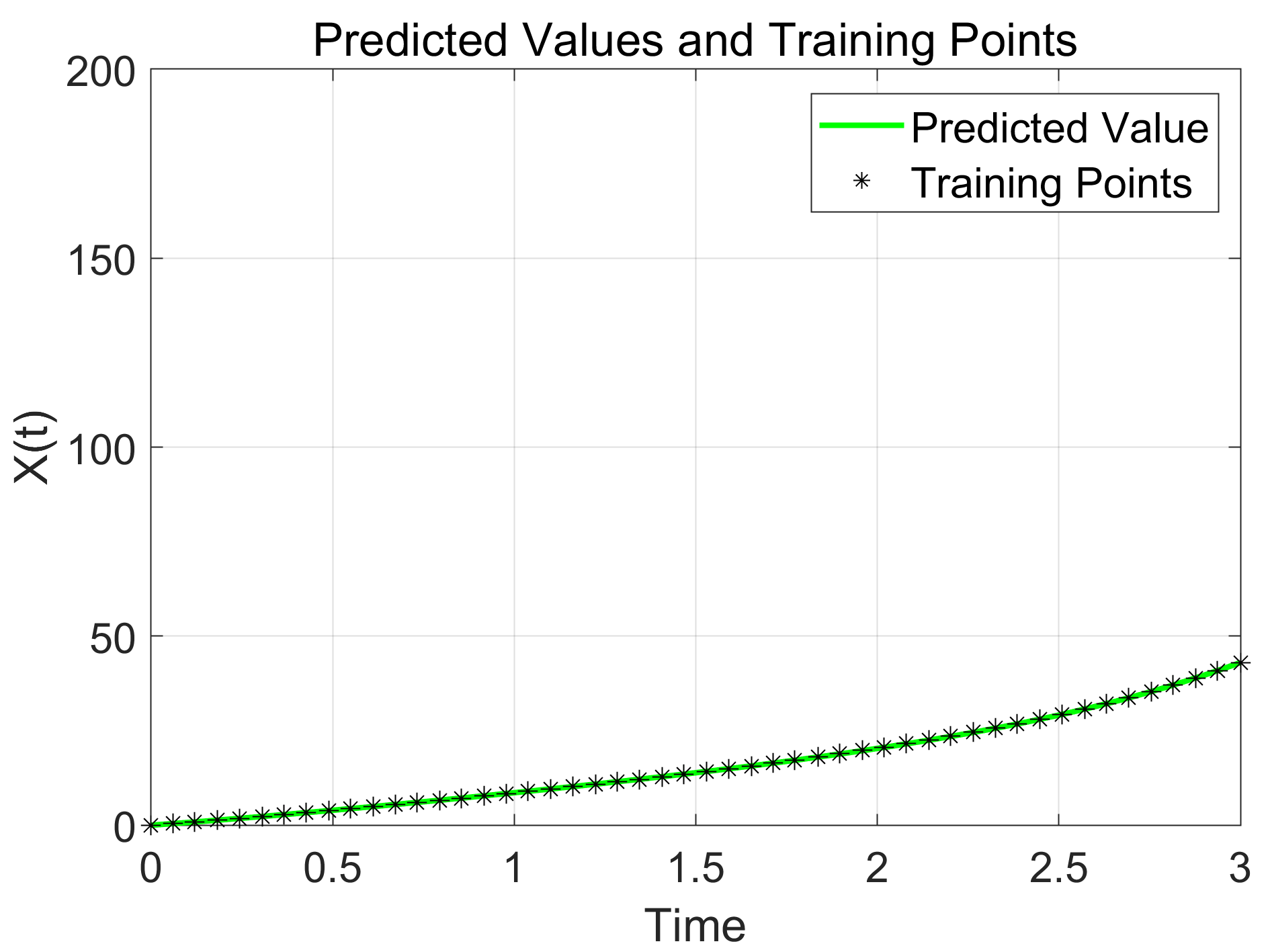}
            \caption*{(b2)$\lambda=1$}
        \end{minipage}\par
        \begin{minipage}{\textwidth}
            \includegraphics[width=\textwidth]{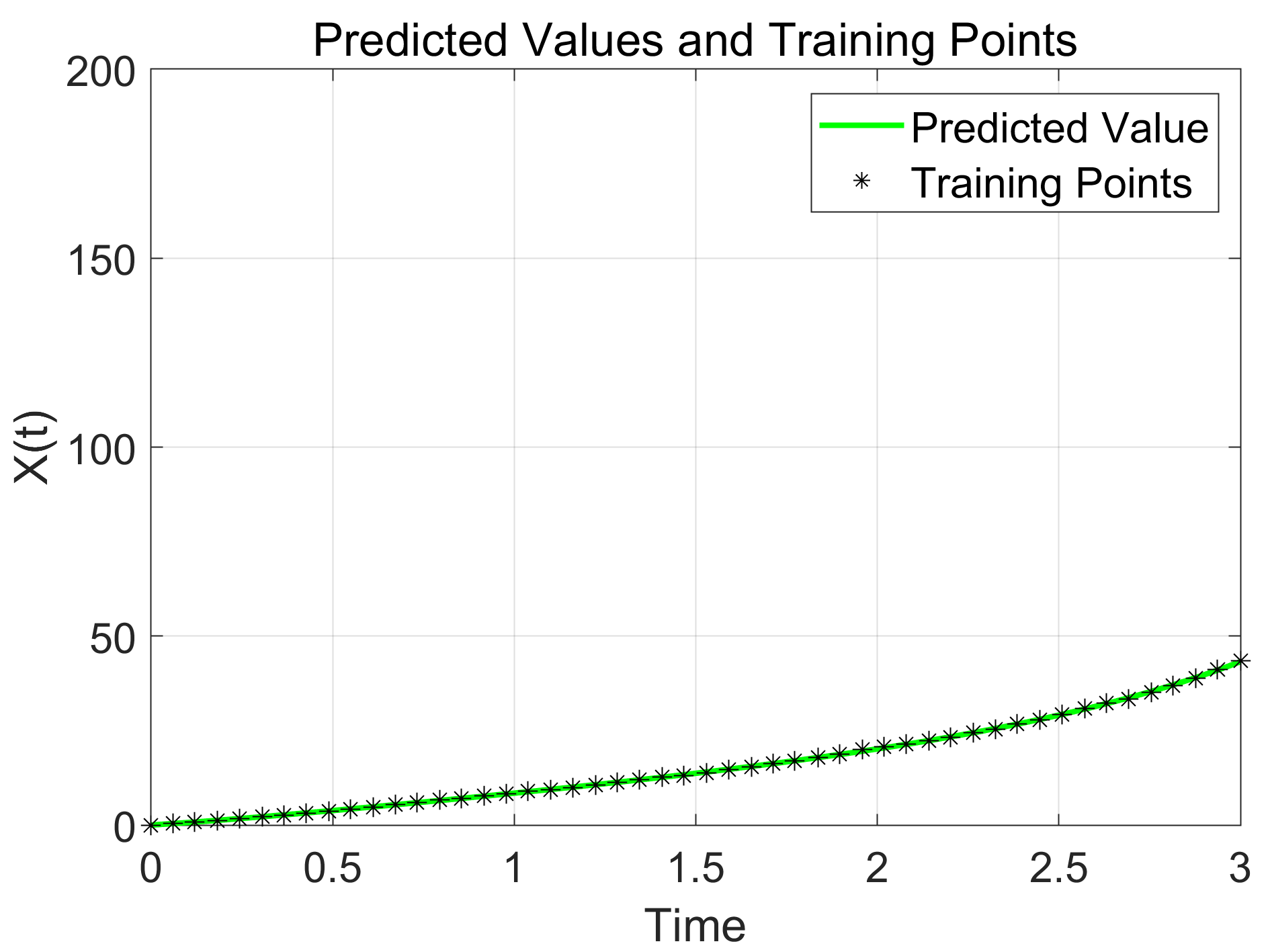}
            \caption*{(c2)$\lambda=5$}
        \end{minipage}\par
        \begin{minipage}{\textwidth}
            \includegraphics[width=\textwidth]{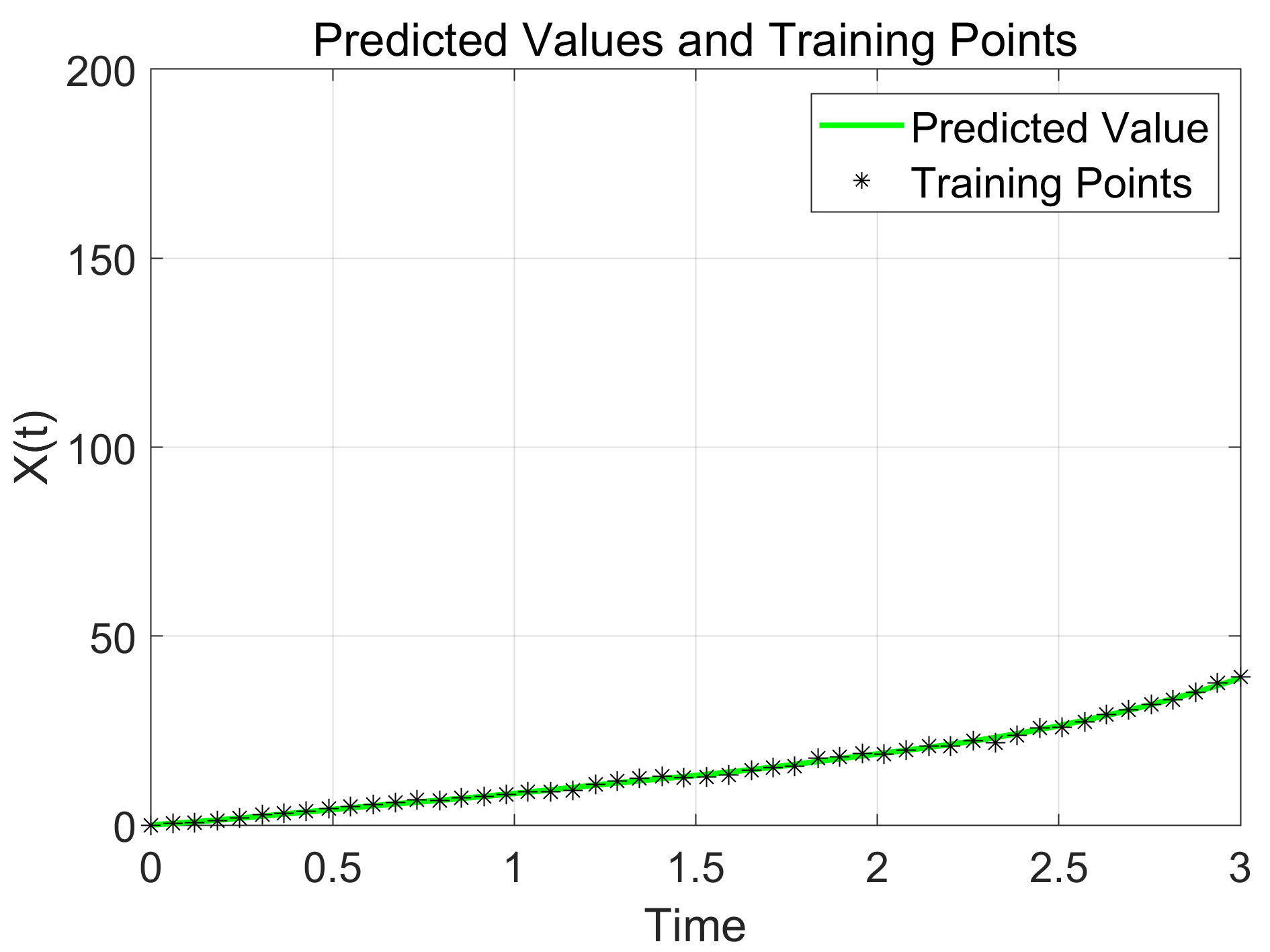}
            \caption*{(d2)$\lambda=20$}
        \end{minipage}
    \end{minipage}%
    \begin{minipage}{0.33\textwidth}
        \centering
        \begin{minipage}{\textwidth}
            \includegraphics[width=\textwidth]{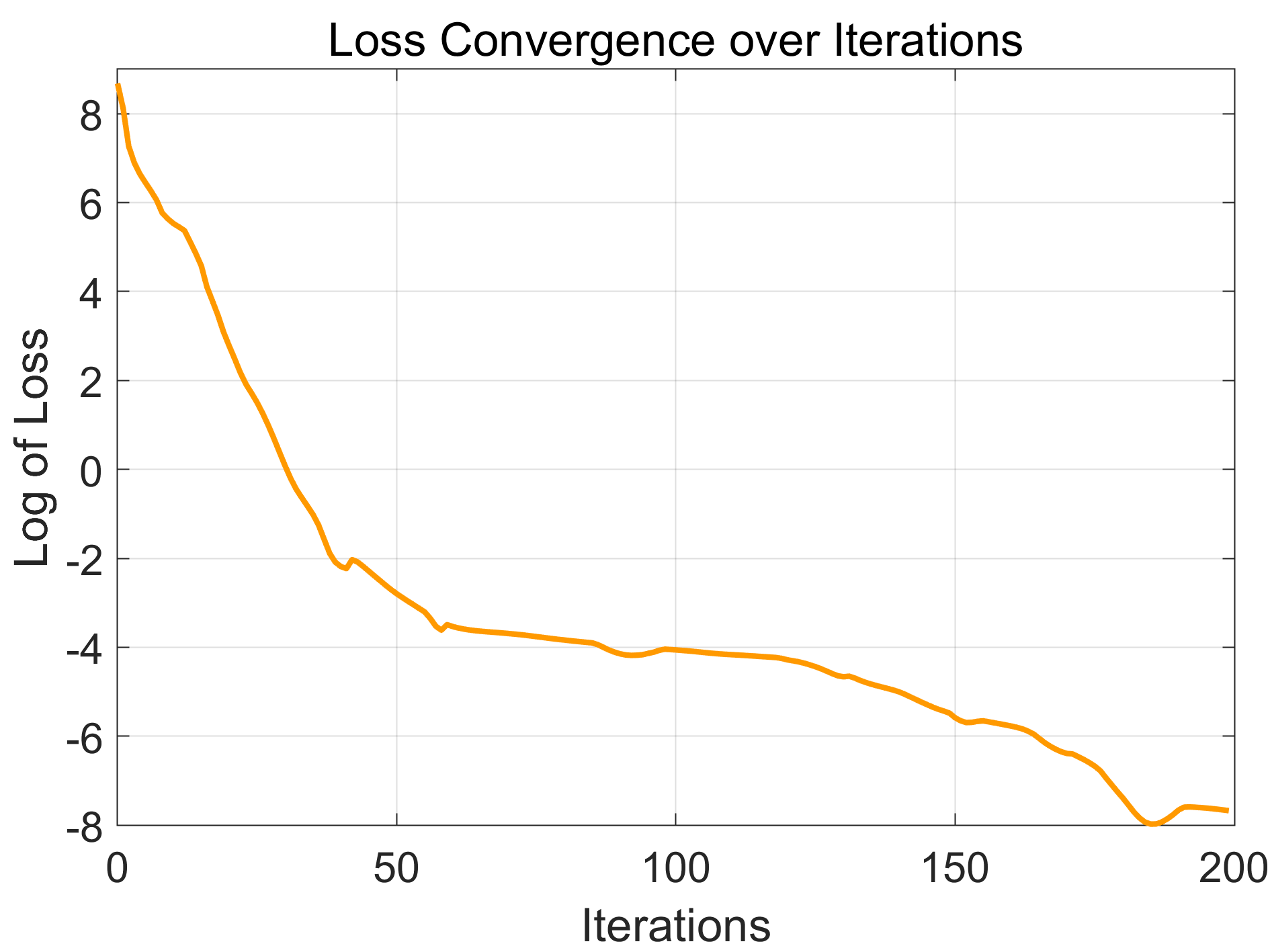}
            \caption*{(a3)$\lambda=0$}
        \end{minipage}\par
        \begin{minipage}{\textwidth}
            \includegraphics[width=\textwidth]{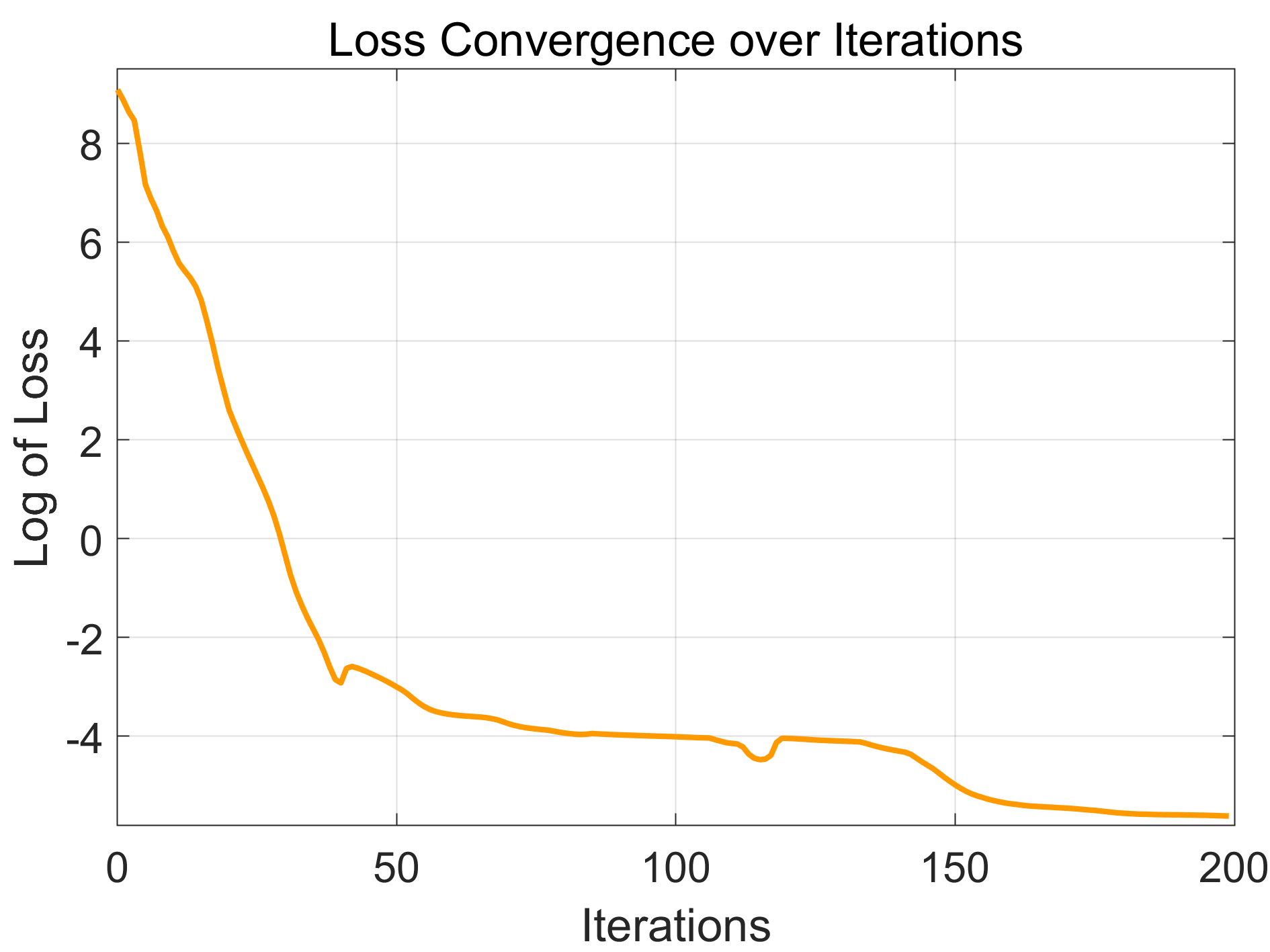}
            \caption*{(b3)$\lambda=1$}
        \end{minipage}\par
        \begin{minipage}{\textwidth}
            \includegraphics[width=\textwidth]{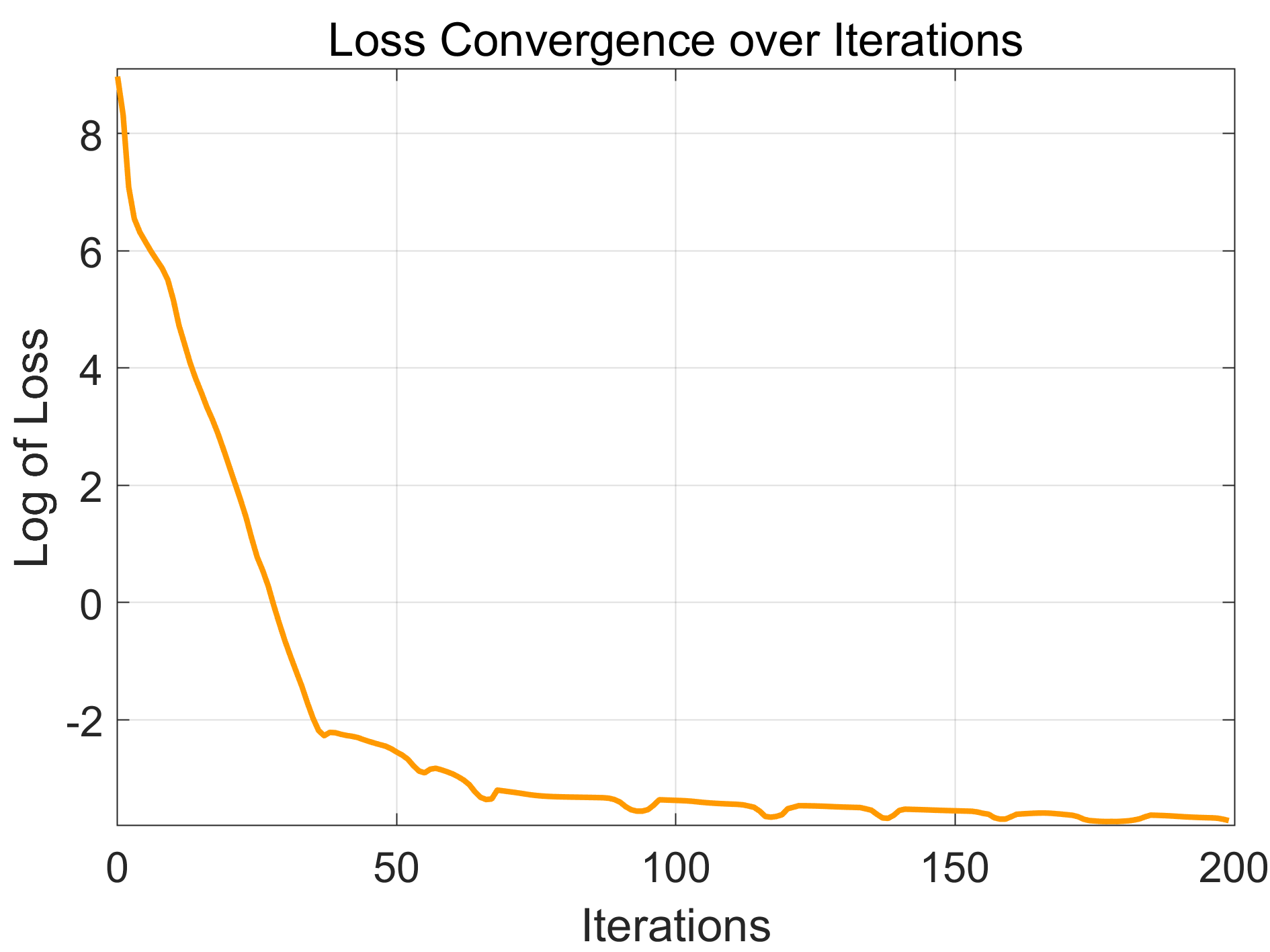}
            \caption*{(c3)$\lambda=5$}
        \end{minipage}\par
        \begin{minipage}{\textwidth}
            \includegraphics[width=\textwidth]{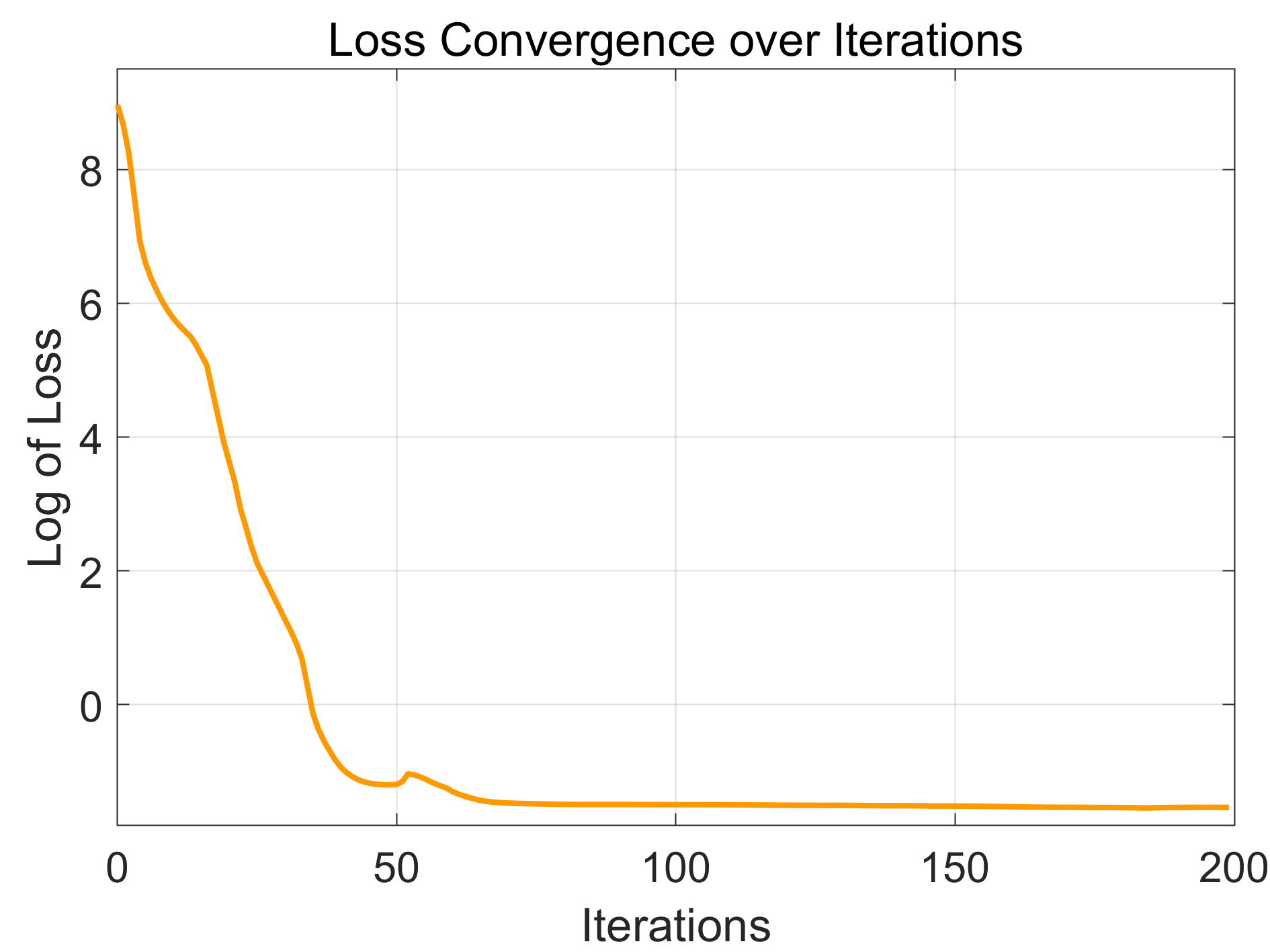}
            \caption*{(d3)$\lambda=20$}
        \end{minipage}
    \end{minipage}
    \caption{$\textbf{Case 1:}$ The length of the integration interval is $3$, with a sampling interval of $3/1000$ and a total of $50$ measurement data points. $\lambda$ is the Gaussian noise level coefficient. Blue line - the Gaussian noise perturbed trajectories, Red line - the mean trajectory. Green line - the improved DNNs predicted value, Black star - the true solution value. The Orange line illustrates the loss convergence over iterations.}
    \label{fig:3x4_case1}
\end{figure}

\subsection{Case 1}
The first experiment investigates the parameter identification problem of Volterra integral equations perturbed by Gaussian noise, characterized by a linear kernel function. The equation is expressed as

\begin{equation}
    X(t) = 4e^t+3t-4 - \int_0^t \theta(t-s)X(s)\:\mathrm{d}s + \int_0^t \lambda B_s\:\mathrm{d}B_s, \qquad t\in [0,3].
    \label{case1sto}
\end{equation}
\par
In the absence of noise disturbances, the true solution of Eq.\,\ref{case1sto} is $2e^t-2\cos{t}+5\sin{t}$ when $\theta=1$. In this case, $\theta$ is an unknown parameter, according to the improved DNNs framework Fig.\,\ref{improvedPINN}, with some measurement data acquired in the true solution, we will apply the improved DNNs to discover the value of $\theta$. Defining a primary output $u(t)$ to represent $X(t)$ and an auxiliary output $v(t)$ to 
represent the first integral in Eq.\,\ref{case1sto}, thus, it can be rewritten as
\begin{equation}
    \begin{aligned}
        &u(t) = 4e^t+3t-4 - \theta\cdot v(t) + \lambda (\frac{1}{2}{B_t}^2-\frac{1}{2}t), \qquad t\in [0,3], \\
        &v(t)=\int_0^t (t-s)u(s)\:\mathrm{d}s, \\
        &v(0)=0, \\
    \end{aligned}
\end{equation}
referring to Eq.\,\ref{mseg} and noting that $E(\frac{1}{2}{B_t}^2-\frac{1}{2}t)=0$, the mean square error of residuals of the measurement data is
\begin{equation}
    \mathrm{MSE}_m=\frac{1}{N_m}\sum_{i=1}^{N_m}\left|u_{pred}(t_i^m;\gamma,\theta)- u_m(t_i^m)\right|^2,
\end{equation}
where $N_m$ is the number of measurement data, $u_{pred}(t_i^m;\gamma,\theta)$ and $u_m(t_i^m)$ are the predicted and measured values at the measurement point $t_i^m$. As for the mean square error of residuals of the governing equation, in this context, we assume that the collocation points coincide precisely with the measurement points, the formula is as follows
\begin{equation}
    \mathrm{MSE}_g=\frac{1}{N_m}\sum_{i=1}^{N_m}\left|u_{pred}(t_i^m;\gamma,\theta)-[4e^{t_i^m}+3{t_i^m}-4-\theta\cdot v_{pred}(t_i^m;\gamma,\theta)]\right|^2\,,
\end{equation}
and the mean square error of residuals of the constraint between the primary output and the auxiliary output is
\begin{equation}
    \mathrm{MSE}_o = \frac{1}{N_m} \sum_{i=1}^{N_m} \left| \frac{\partial^2 v_{pred}(t_i^m; \gamma,\theta)}{\partial t^2} - u_{pred}(t_i^m; \gamma,\theta) \right|^2 \,,
\end{equation}
the mean square error of residuals of the initial condition of the auxiliary output is
\begin{equation}
    \mathrm{MSE}_i=\left| v_{pred}(0;\gamma)-0 \right|^2.
\end{equation}
\par
The loss function $\mathrm{MSE}_{total}$ is expressed as
\begin{equation}
    \mathrm{MSE}_{total}=w_m\cdot \mathrm{MSE}_m + w_g\cdot \mathrm{MSE}_g + w_i\cdot \mathrm{MSE}_i + w_o\cdot \mathrm{MSE}_o.
\end{equation}
As mentioned in Section \ref{sec:PINN}, in the presence of measurement noise, the optimal value of $\mathrm{MSE}_m$ would not be $0$, so we set $w_m=1$ and compute $w_g$, $w_i$ and $w_o$ in line with the adaptive weighting strategy. We evenly sample $50$ measurement data points from the expected values of $100$ simulated trajectories, using these as noise-free training data. In this scenario, we obtain a highly accurate value for $\theta$ after $200$ iterations, which closely matches the true value. The relative error between the predicted and the true parameter values, and the absolute error between the predicted values and noise-free training data with different noise levels are shown in Table \ref{tab:comparison}, the simulated trajectories, the training data and the predicted values of $X(t)$, the convergence of the iterations are shown in Fig.\,\ref{fig:3x4_case1}.

\begin{figure}[htbp]
    \centering
    \begin{minipage}{0.33\textwidth}
        \centering
        \begin{minipage}{\textwidth}
            \includegraphics[width=\textwidth]{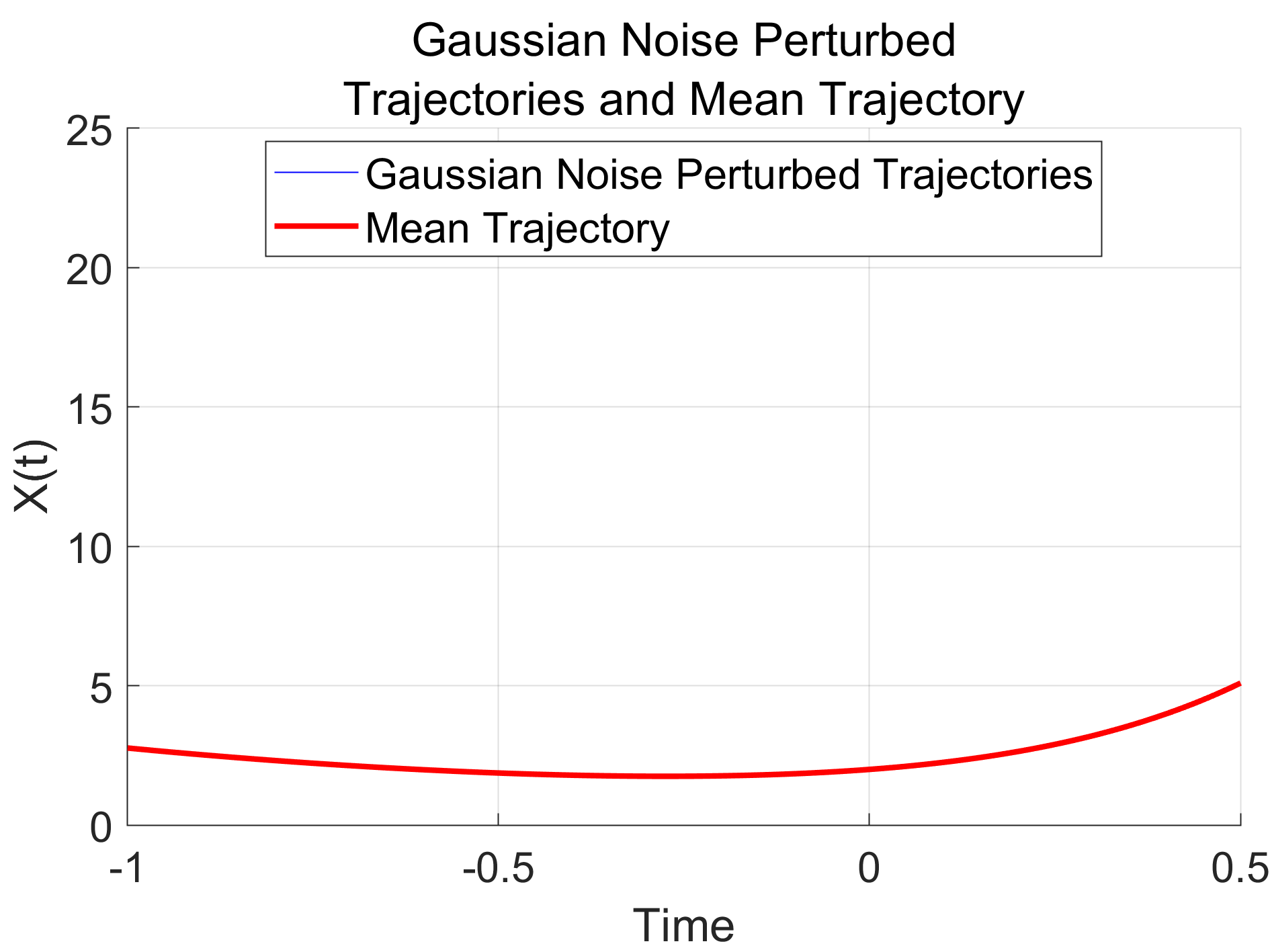}
            \caption*{(a1)$\lambda=0$}
        \end{minipage}\par
        \begin{minipage}{\textwidth}
            \includegraphics[width=\textwidth]{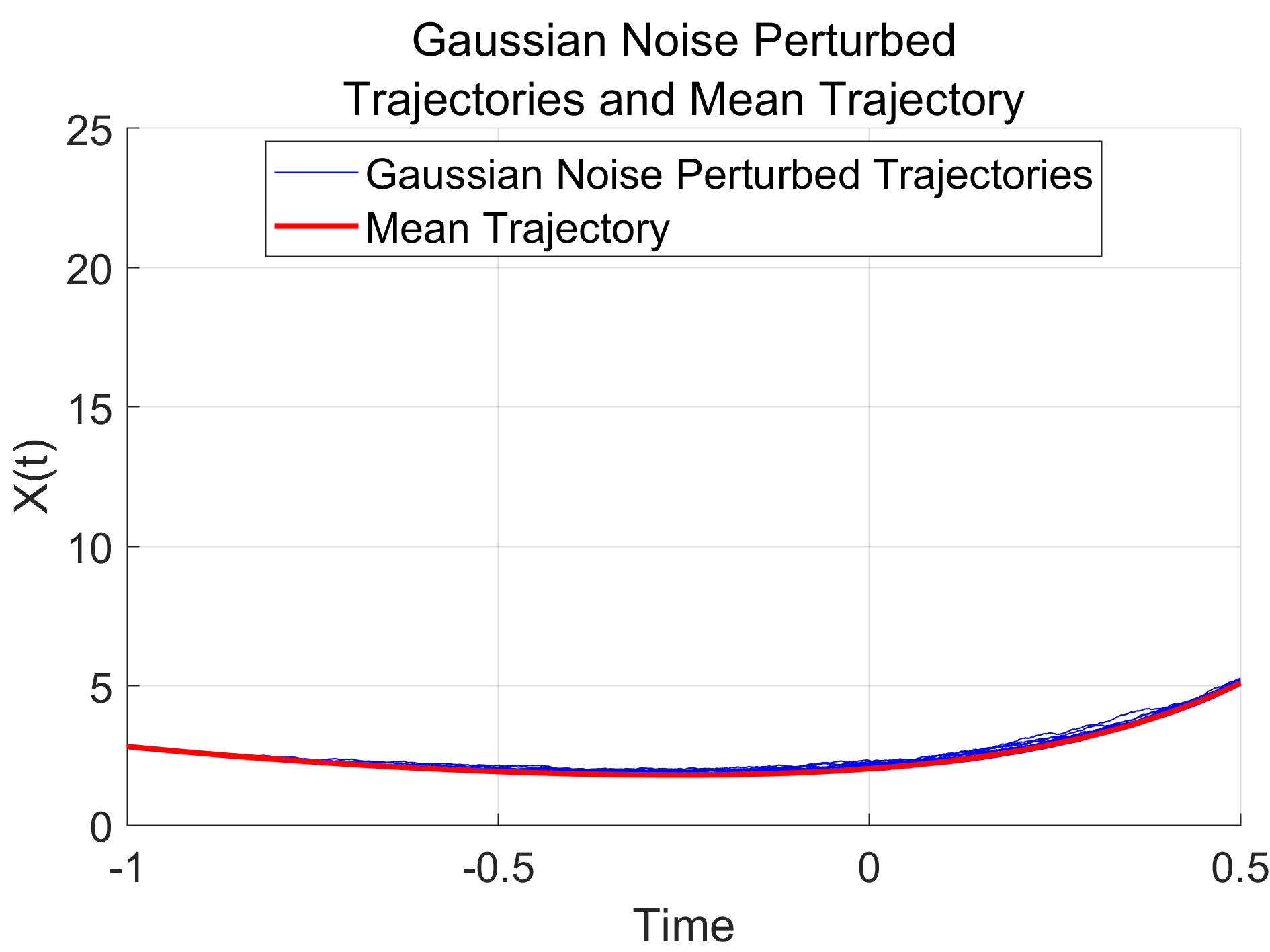}
            \caption*{(b1)$\lambda=0.1$}
        \end{minipage}\par
        \begin{minipage}{\textwidth}
            \includegraphics[width=\textwidth]{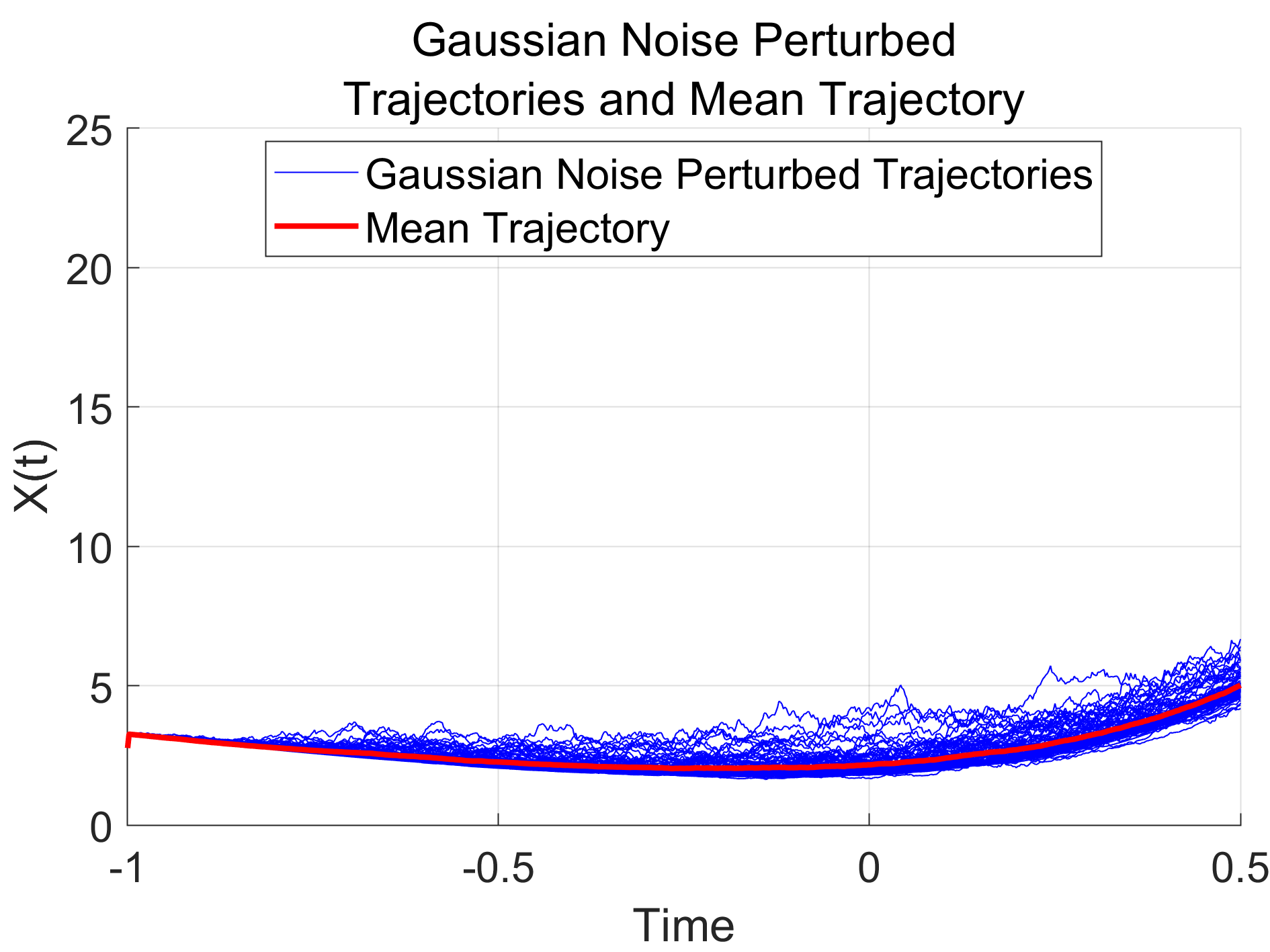}
            \caption*{(c1)$\lambda=1$}
        \end{minipage}\par
        \begin{minipage}{\textwidth}
            \includegraphics[width=\textwidth]{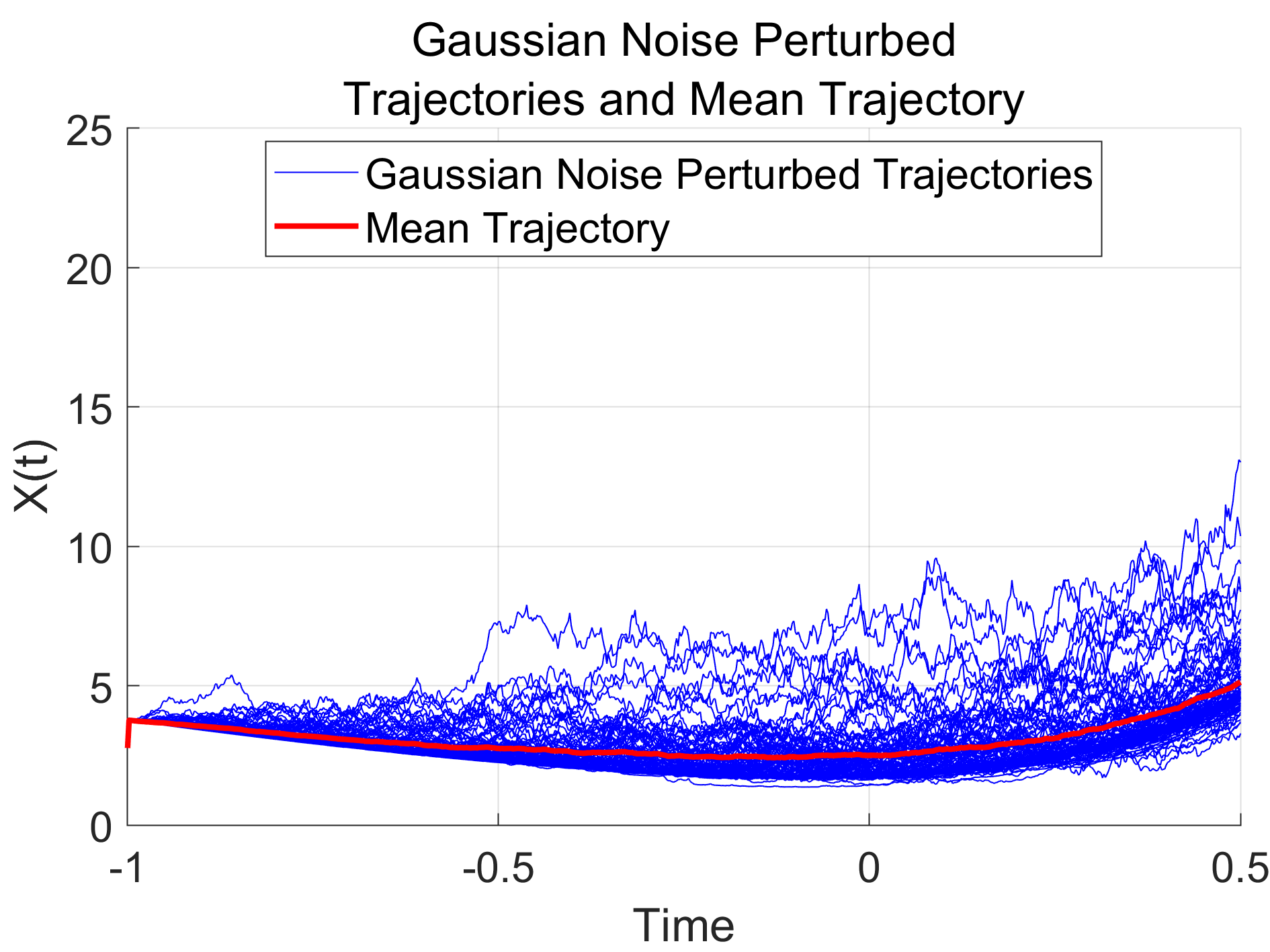}
            \caption*{(d1)$\lambda=2$}
        \end{minipage}
    \end{minipage}%
    \begin{minipage}{0.33\textwidth}
        \centering
        \begin{minipage}{\textwidth}
            \includegraphics[width=\textwidth]{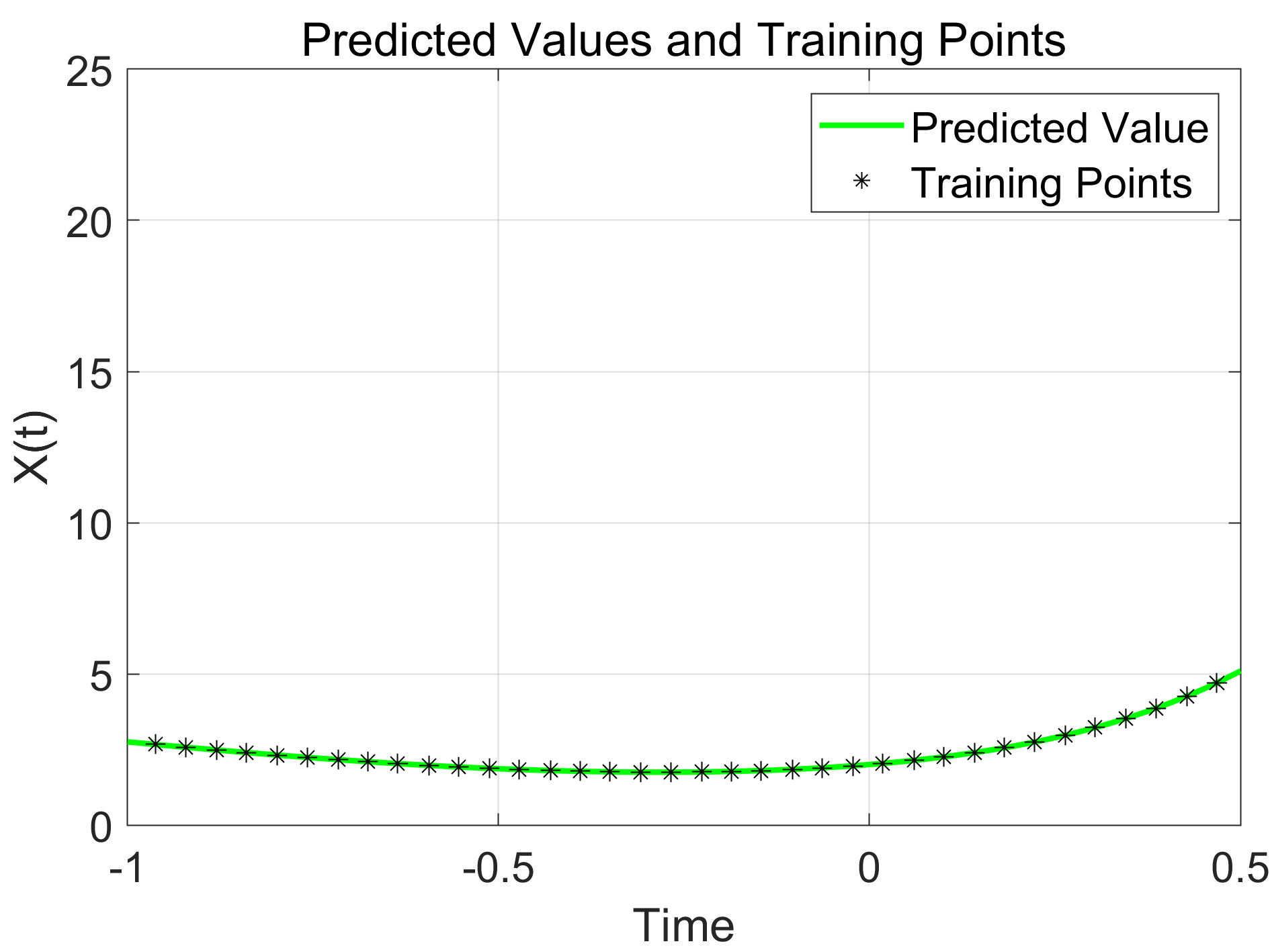}
            \caption*{(a2)$\lambda=0$}
        \end{minipage}\par
        \begin{minipage}{\textwidth}
            \includegraphics[width=\textwidth]{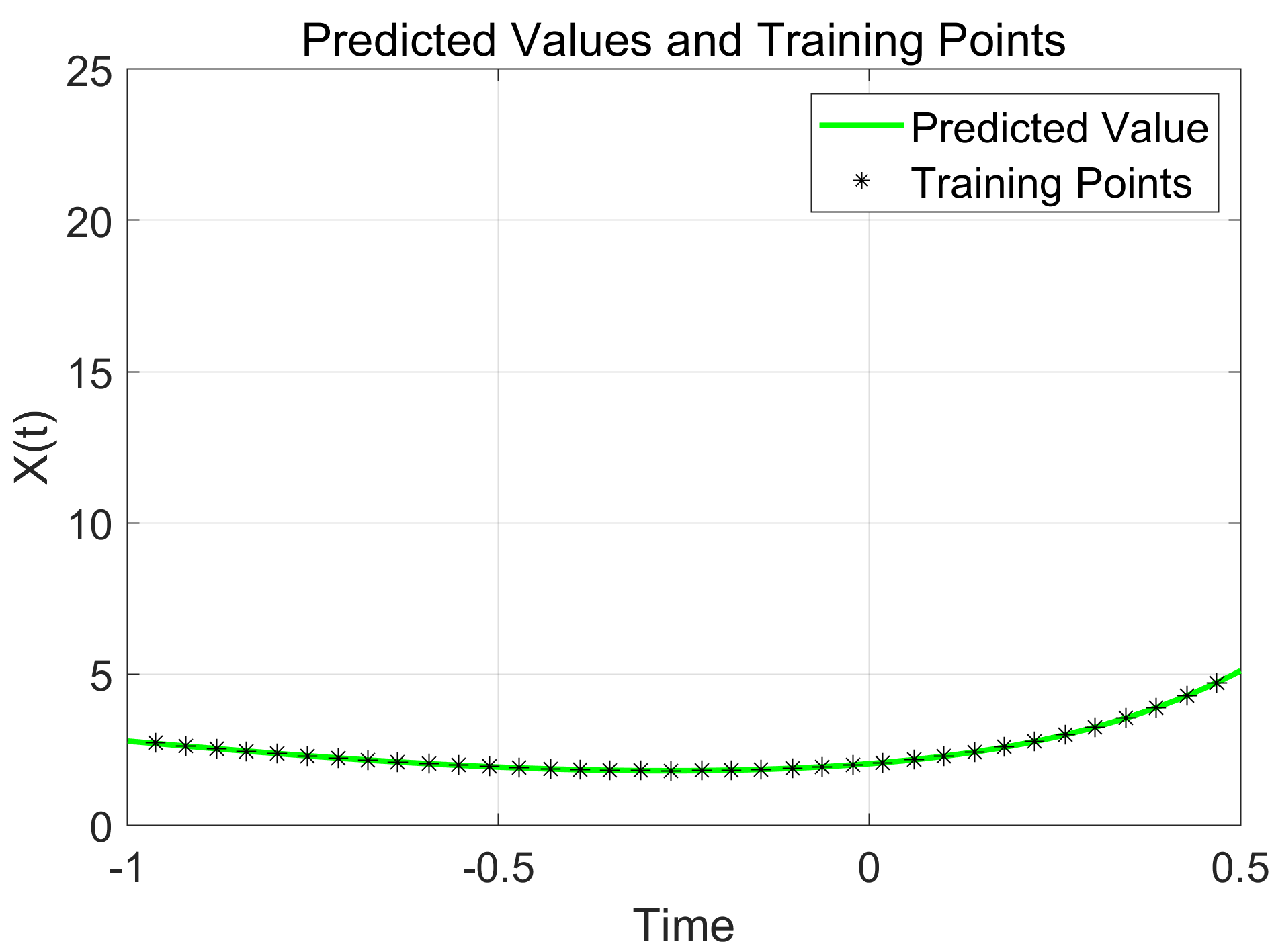}
            \caption*{(b2)$\lambda=0.1$}
        \end{minipage}\par
        \begin{minipage}{\textwidth}
            \includegraphics[width=\textwidth]{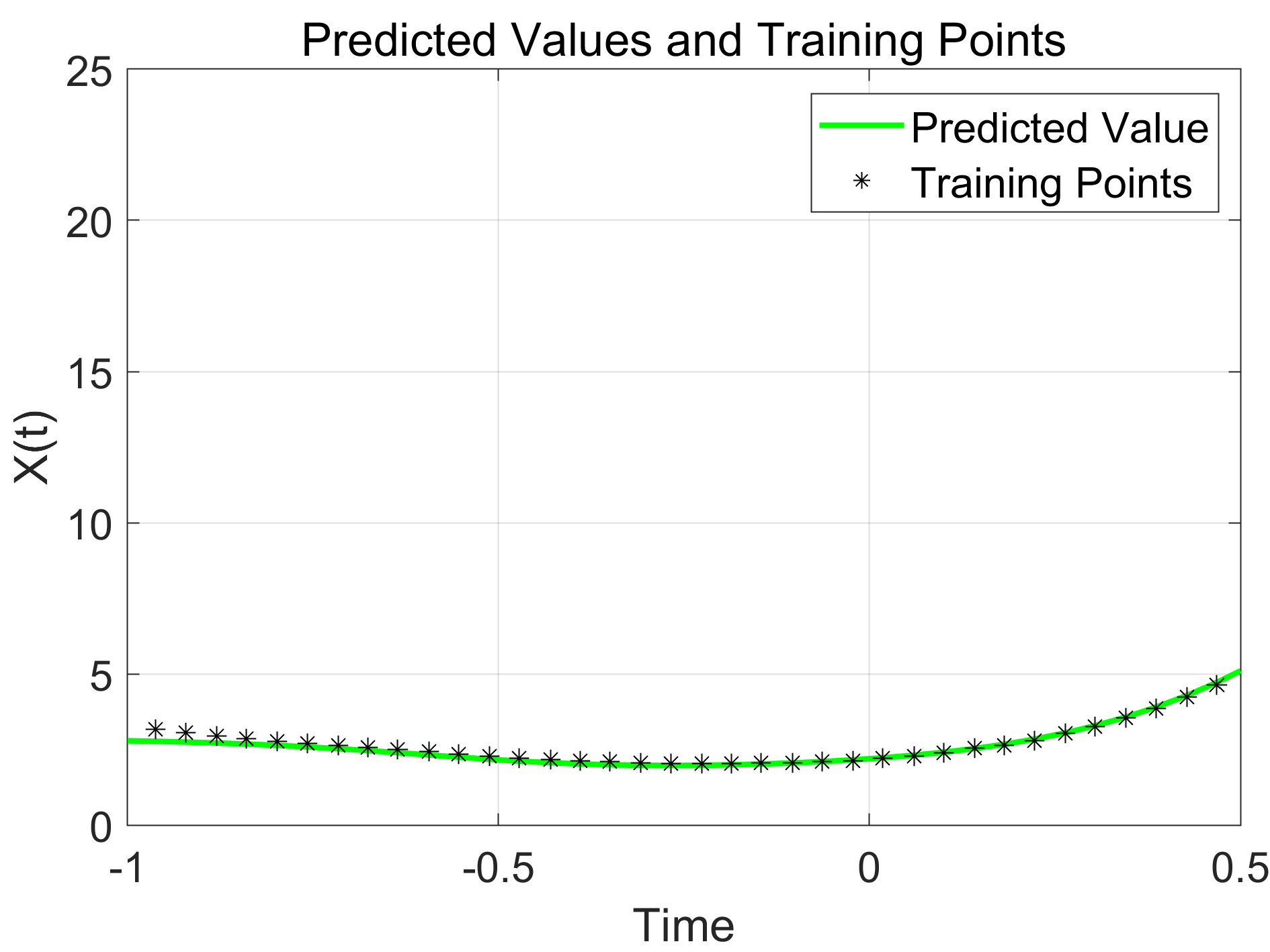}
            \caption*{(c2)$\lambda=1$}
        \end{minipage}\par
        \begin{minipage}{\textwidth}
            \includegraphics[width=\textwidth]{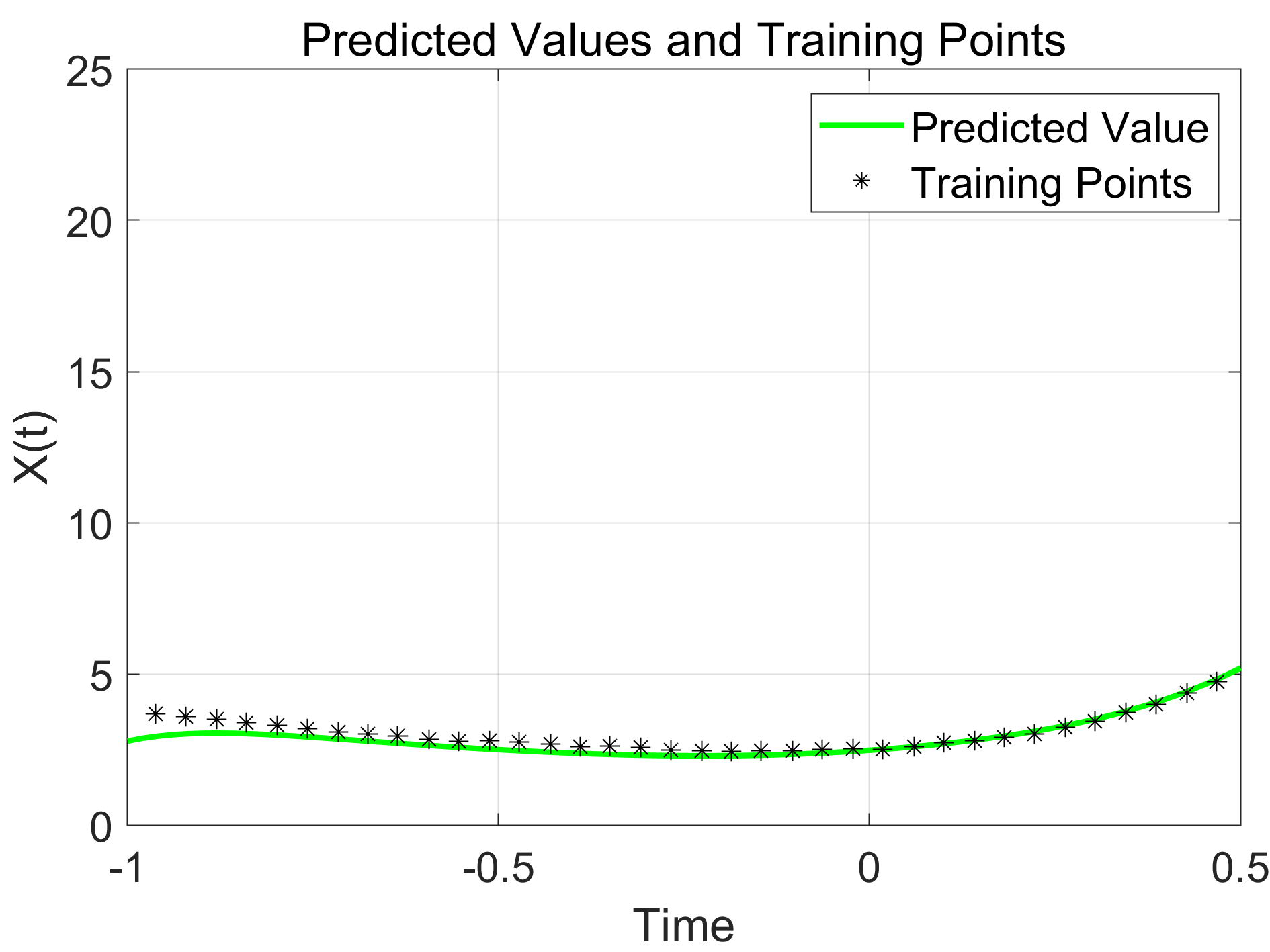}
            \caption*{(d2)$\lambda=2$}
        \end{minipage}
    \end{minipage}%
    \begin{minipage}{0.33\textwidth}
        \centering
        \begin{minipage}{\textwidth}
            \includegraphics[width=\textwidth]{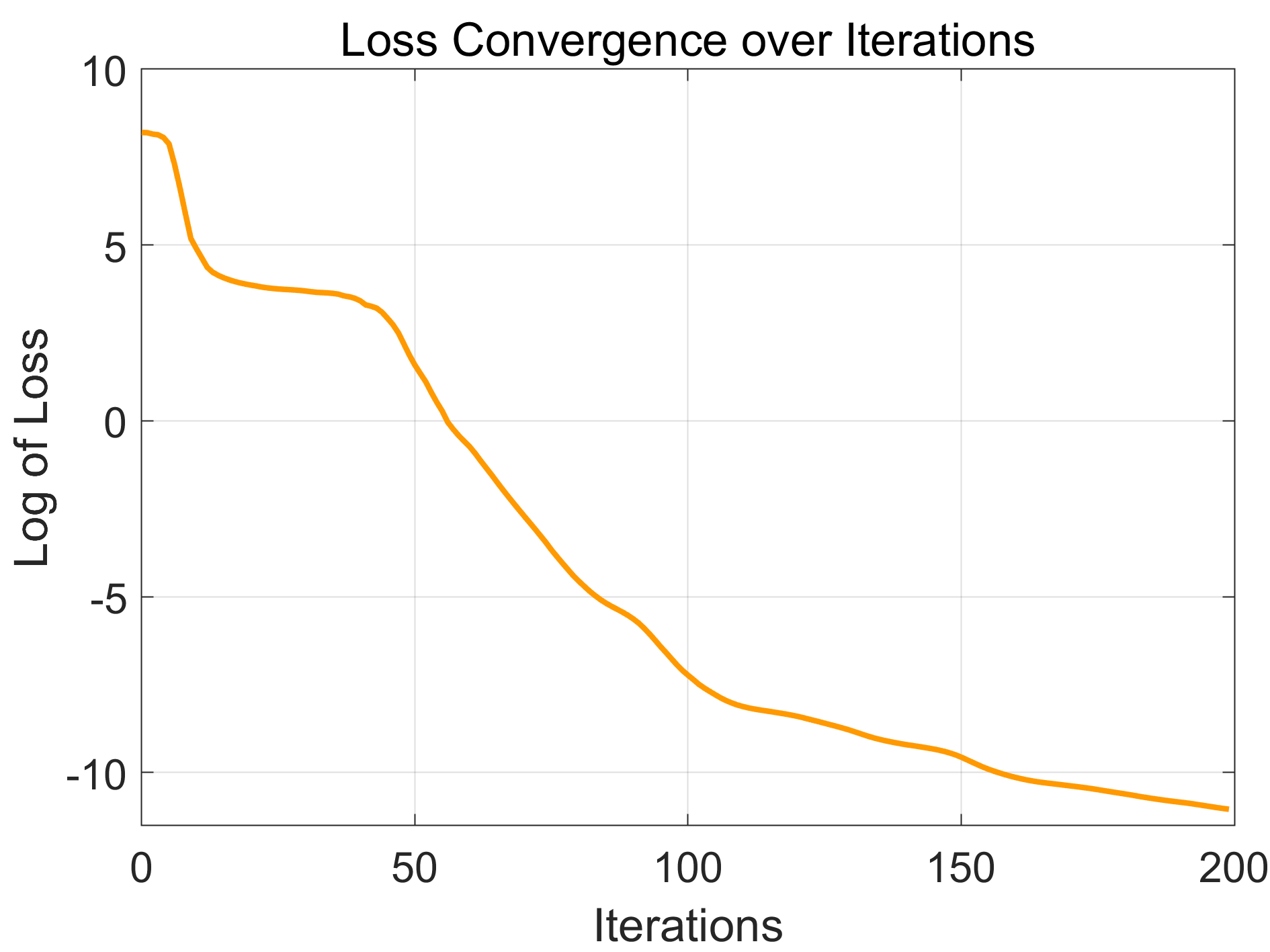}
            \caption*{(a3)$\lambda=0$}
        \end{minipage}\par
        \begin{minipage}{\textwidth}
            \includegraphics[width=\textwidth]{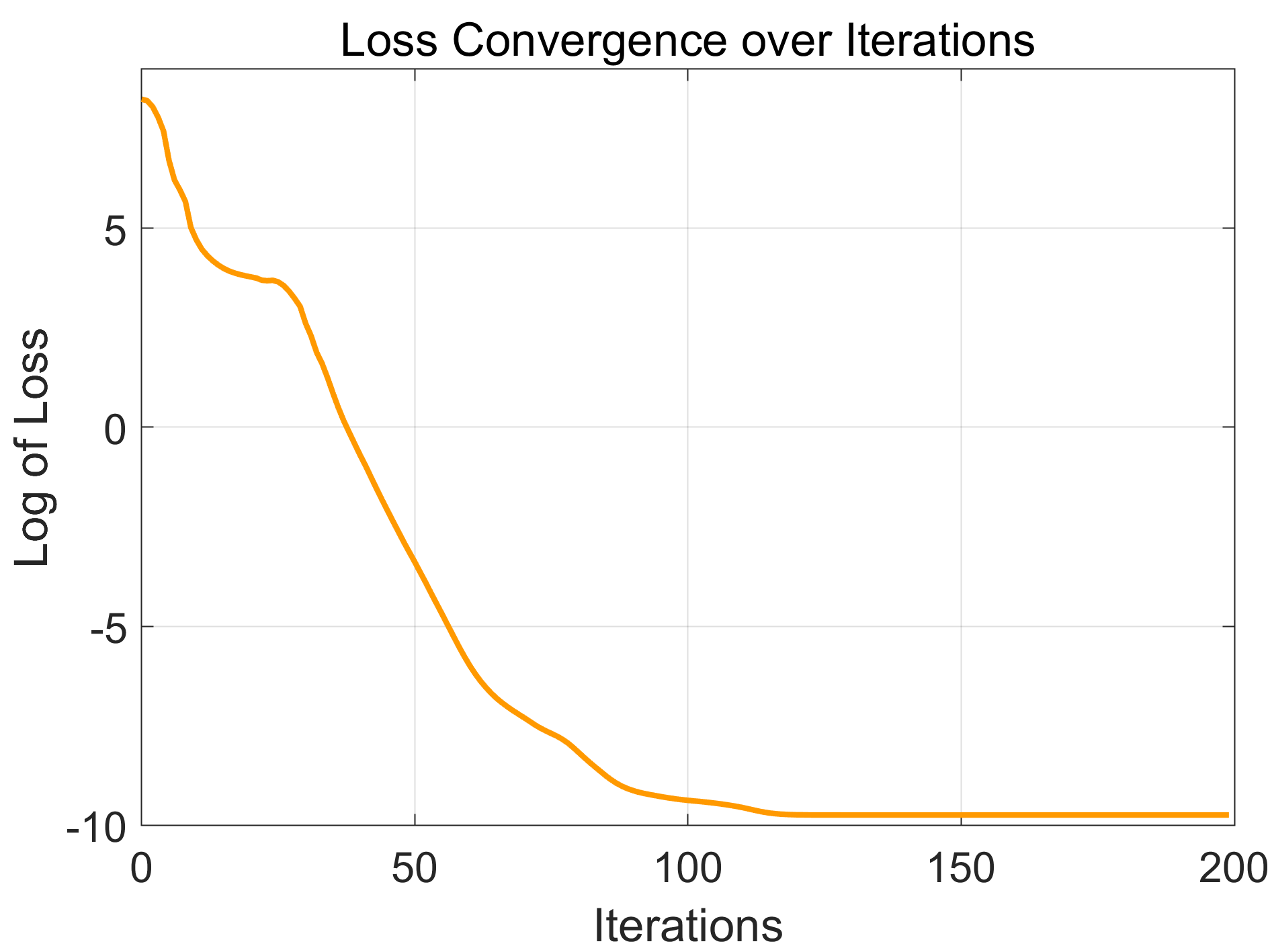}
            \caption*{(b3)$\lambda=0.1$}
        \end{minipage}\par
        \begin{minipage}{\textwidth}
            \includegraphics[width=\textwidth]{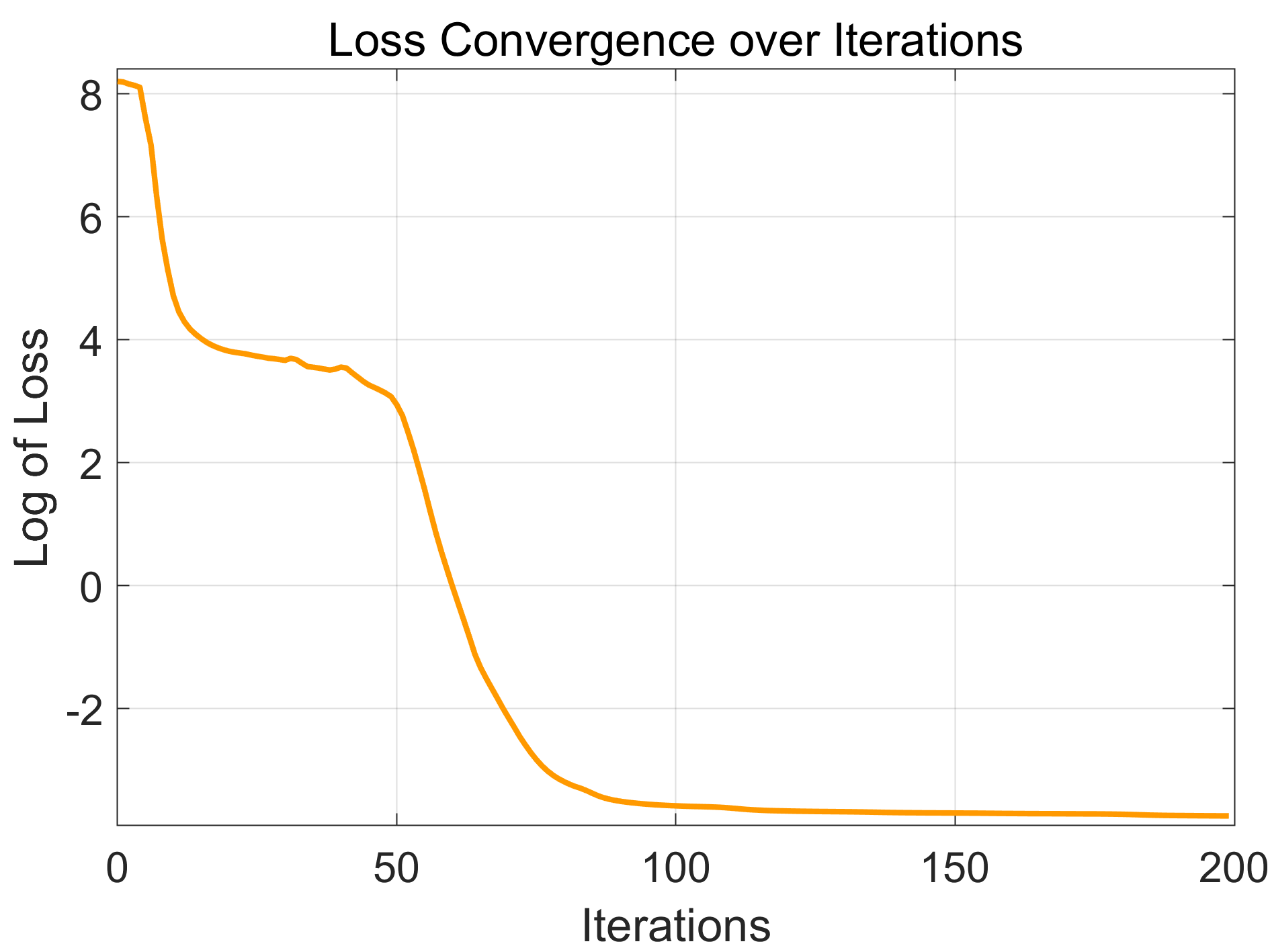}
            \caption*{(c3)$\lambda=1$}
        \end{minipage}\par
        \begin{minipage}{\textwidth}
            \includegraphics[width=\textwidth]{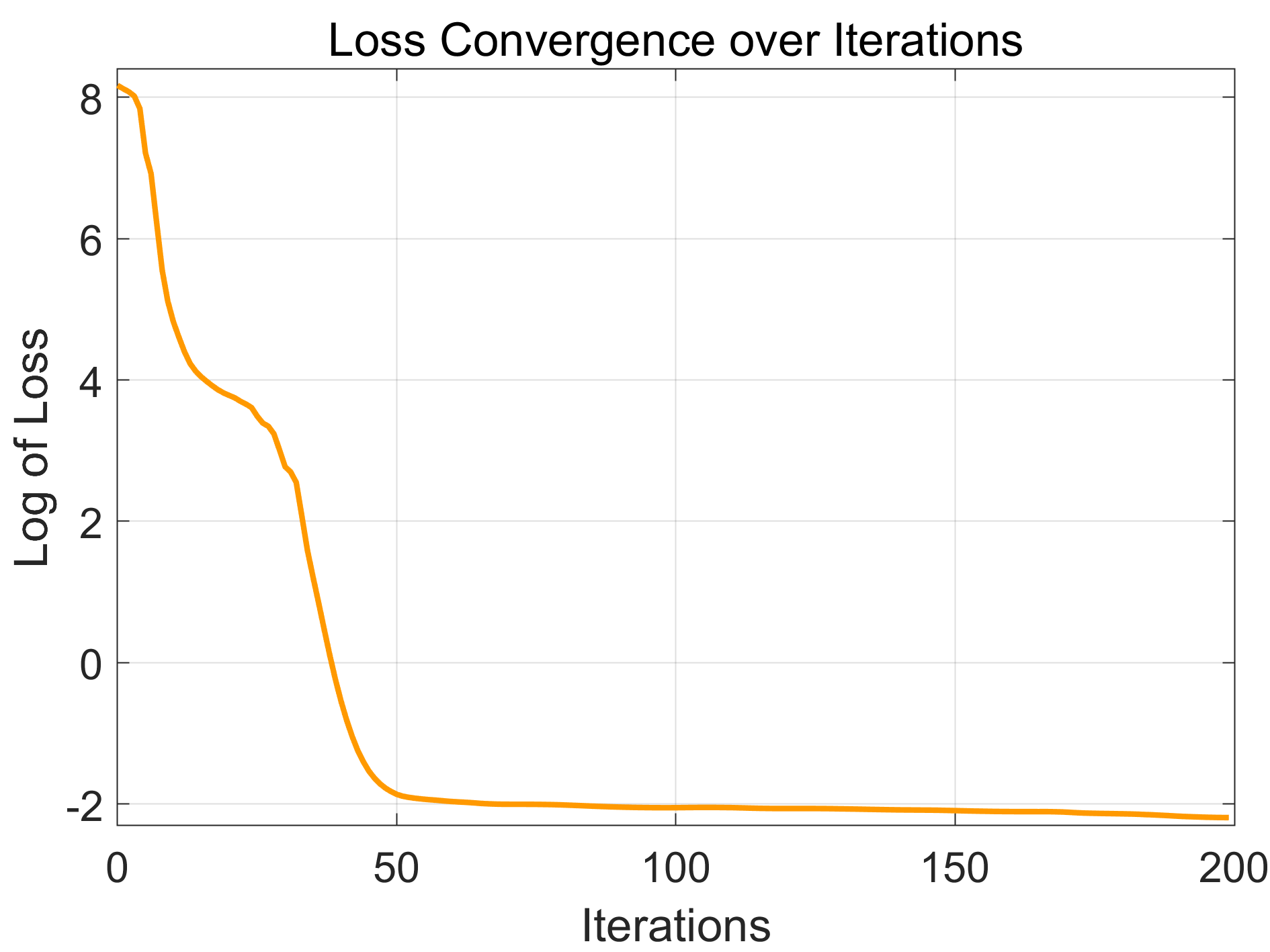}
            \caption*{(d3)$\lambda=2$}
        \end{minipage}
    \end{minipage}
    \caption{$\textbf{Case 2:}$ The length of the integration interval is $\frac{3}{2}$, with a sampling interval of $3/2000$ and a total of $50$ measurement data points. $\lambda$ is the Gaussian noise level coefficient. Blue line - the Gaussian noise perturbed trajectories, Red line - the mean trajectory. Green line - the improved DNNs predicted value, Black star - the true solution value. The Orange line illustrates the loss convergence over iterations.}
    \label{fig:3x4_case2}
\end{figure}

\subsection{Case 2}
The second experiment investigates the parameter identification problem of Volterra integral equations perturbed by Gaussian noise, characterized by a nonlinear kernel function. The equation is expressed as
\begin{equation}
    X(t)=e^{-t}+e^{3t}+e^t (t+1)+\frac{1}{4}e^t (e^{4t}-e^{-4}) - \int_{0}^{t+1} \theta e^{t+s-1}X(s-1)\:\mathrm{d}s + \int_{0}^{t+1} \lambda B_s\:\mathrm{d}B_s, \,\,\,\, t\in [-1,\frac{1}{2}].
    \label{case2sto}
\end{equation}
\par
In the absence of noise disturbances, the true solution of Eq.\,\ref{case2sto} is $e^{-t}+e^{3t}$ when $\theta=1$. In contrast to Case 1, the kernel function in Case 2 is nonlinear, with the variable $t$ constrained to the range $[-1,\frac{1}{2}]$ and the upper limit of the integral is $t+1$. We continue to use the improved DNNs to address this problem and assess its generalizability. Similar to Case 1, we introduce a primary output $u(t)$ to represent $X(t)$ and an auxiliary output $v(t)$ to 
represent the first integral in Eq.\,\ref{case2sto}, so the equation would be represented as
\begin{equation}
    \begin{aligned}
        &u(t) = e^{-t}+e^{3t}+e^t (t+1)+\frac{1}{4}e^t (e^{4t}-e^{-4}) - \theta\cdot v(t) + \lambda (\frac{1}{2}{B^2_{t+1}}-\frac{1}{2}t-\frac{1}{2}), \qquad t\in [-1,\frac{1}{2}], \\
        &v(t)=\int_{0}^{t+1} e^{t+s-1}u(s-1)\:\mathrm{d}s, \\
        &v(-1)=0, \\
    \end{aligned}
\end{equation}
the mean square error of residuals of the measurement data is
\begin{equation}
    \mathrm{MSE}_m=\frac{1}{N_m}\sum_{i=1}^{N_m}\left|u_{pred}(t_i^m;\gamma,\theta)- u_m(t_i^m)\right|^2,
\end{equation}
and again we simply assume that the collocation points coincide precisely with the measurement points, the mean square error of residuals of the governing equation is
\begin{equation}
    \mathrm{MSE}_g=\frac{1}{N_m}\sum_{i=1}^{N_m} \left|u_{pred}(t_i^m;\gamma,\theta)-[e^{-t_i^m}+e^{3t_i^m}+e^{t_i^m} (t_i^m+1)+\frac{1}{4}e^{t_i^m} (e^{4t_i^m}-e^{-4})-\theta\cdot v_{pred}(t_i^m;\gamma,\theta)] \right|^2\,,
\end{equation}
the mean square error of residuals of the constraint between the primary output and the auxiliary output is represented as
\begin{equation}
    \mathrm{MSE}_o = \frac{1}{N_m} \sum_{i=1}^{N_m} \left| \frac{\partial v_{pred}(t_i^m; \gamma,\theta)}{\partial t} - e^{2t_i^m} * u_{pred}(t_i^m; \gamma,\theta) - v_{pred}(t_i^m; \gamma,\theta) \right|^2 \,,
\end{equation}
the mean square error of residuals of the initial condition of the auxiliary output is
\begin{equation}
    \mathrm{MSE}_i=\left| v_{pred}(-1;\gamma)-0 \right|^2.
\end{equation}
\par
The loss function $\mathrm{MSE}_{total}$ is expressed as
\begin{equation}
    \mathrm{MSE}_{total}=w_m\cdot \mathrm{MSE}_m + w_g\cdot \mathrm{MSE}_g + w_i\cdot \mathrm{MSE}_i + w_o\cdot \mathrm{MSE}_o.
\end{equation}
The same adaptive weighting strategy as in Case 1 is adopted, we evenly sample $50$ measurement data points from the expected values of $100$ simulated trajectories, using these as noise-free training data. By the improved DNNs, the obtained value for $\theta$ is still quite close to the true value, showing good accuracy after $200$ iterations. The relative error between the predicted and the true parameter values, and the absolute error between the predicted values and noise-free training data with different noise levels are shown in Table \ref{tab:comparison}, the simulated trajectories, the training data and the predicted values of $X(t)$, along with the convergence of the iterations are shown in Fig.\,\ref{fig:3x4_case2}.

\begin{figure}[htbp]
    \centering
    \begin{minipage}{0.33\textwidth}
        \centering
        \begin{minipage}{\textwidth}
            \includegraphics[width=\textwidth]{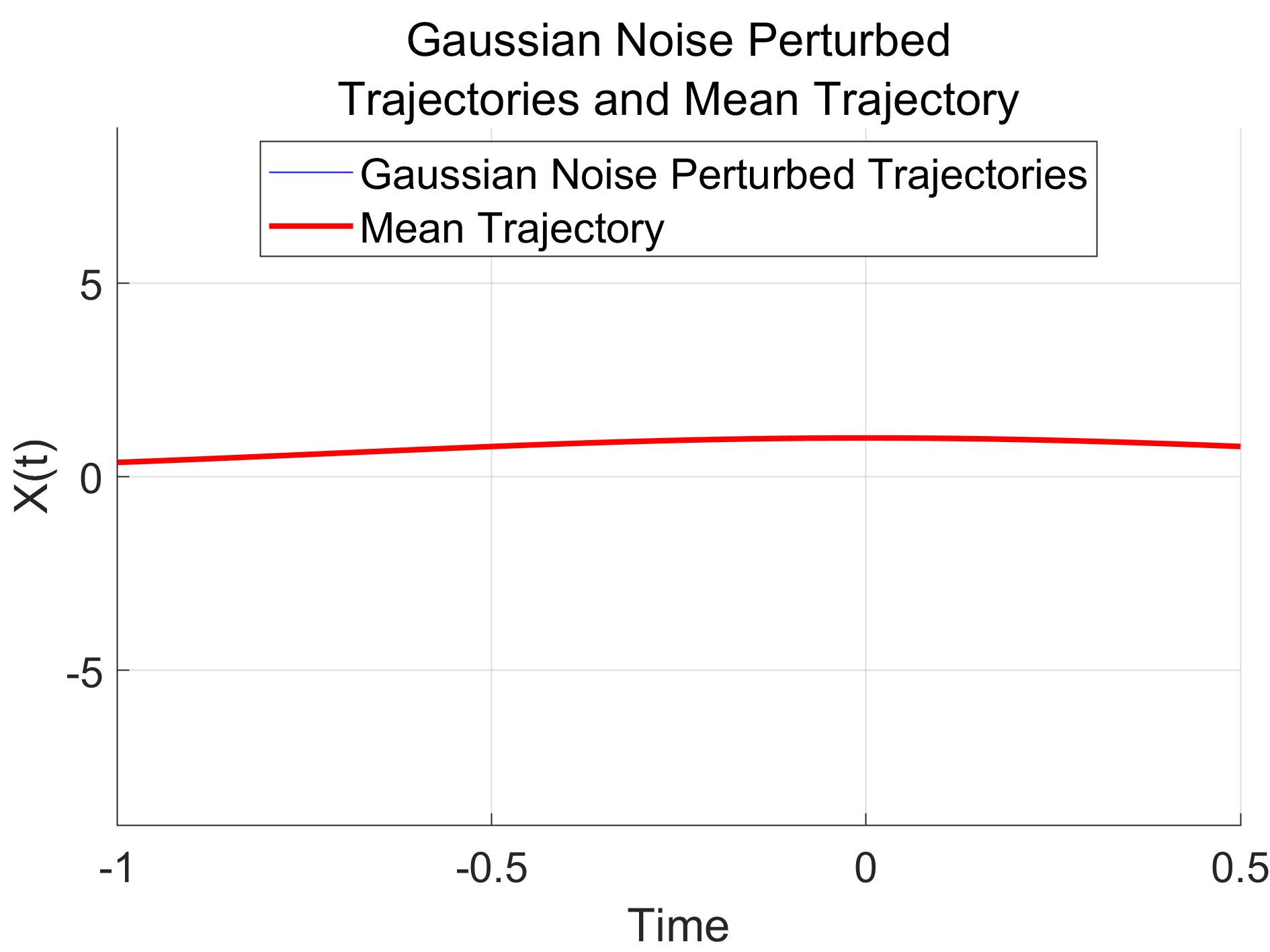}
            \caption*{(a1)$\lambda=0$}
        \end{minipage}\par
        \begin{minipage}{\textwidth}
            \includegraphics[width=\textwidth]{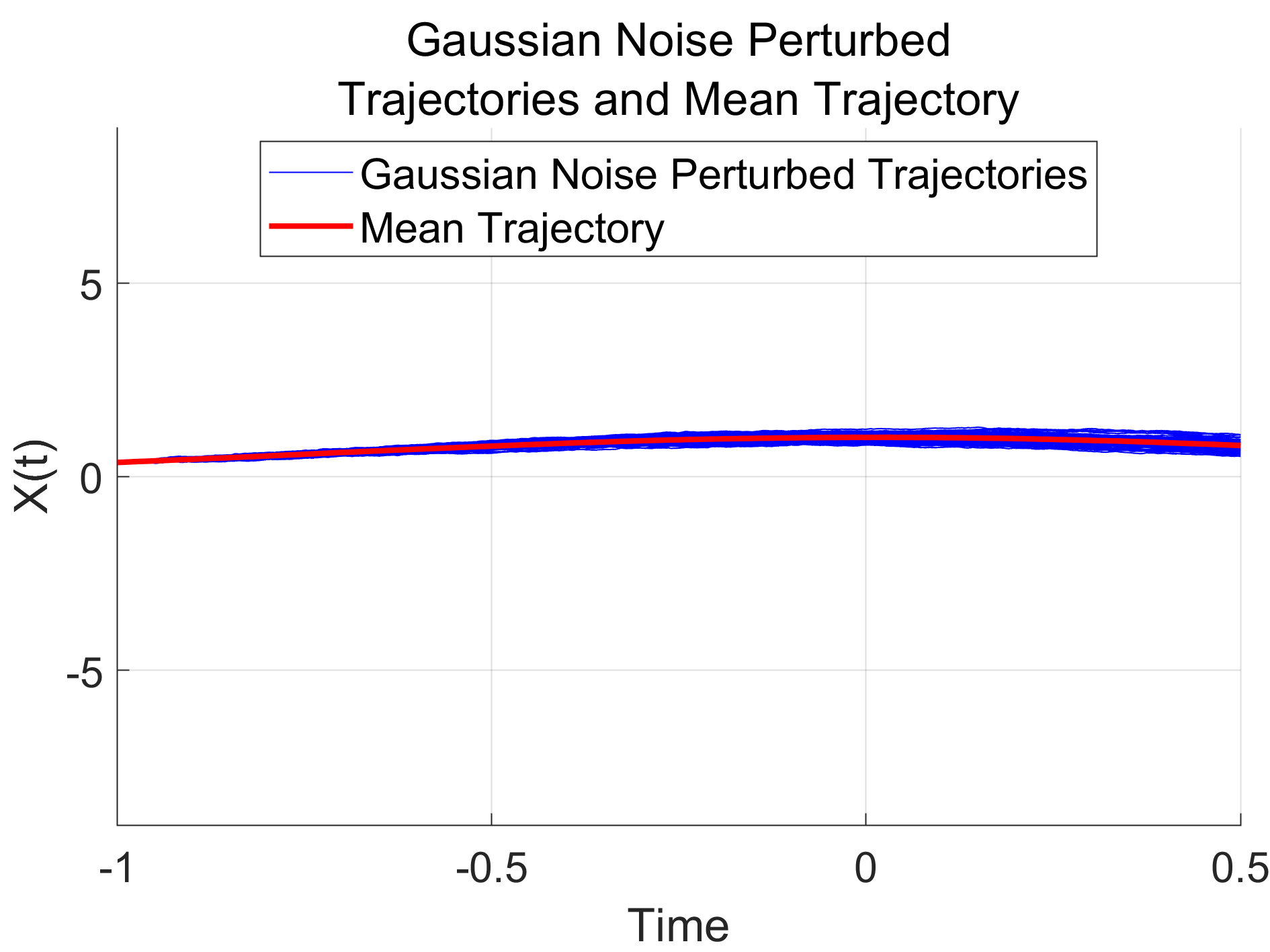}
            \caption*{(b1)$\lambda=0.1$}
        \end{minipage}\par
        \begin{minipage}{\textwidth}
            \includegraphics[width=\textwidth]{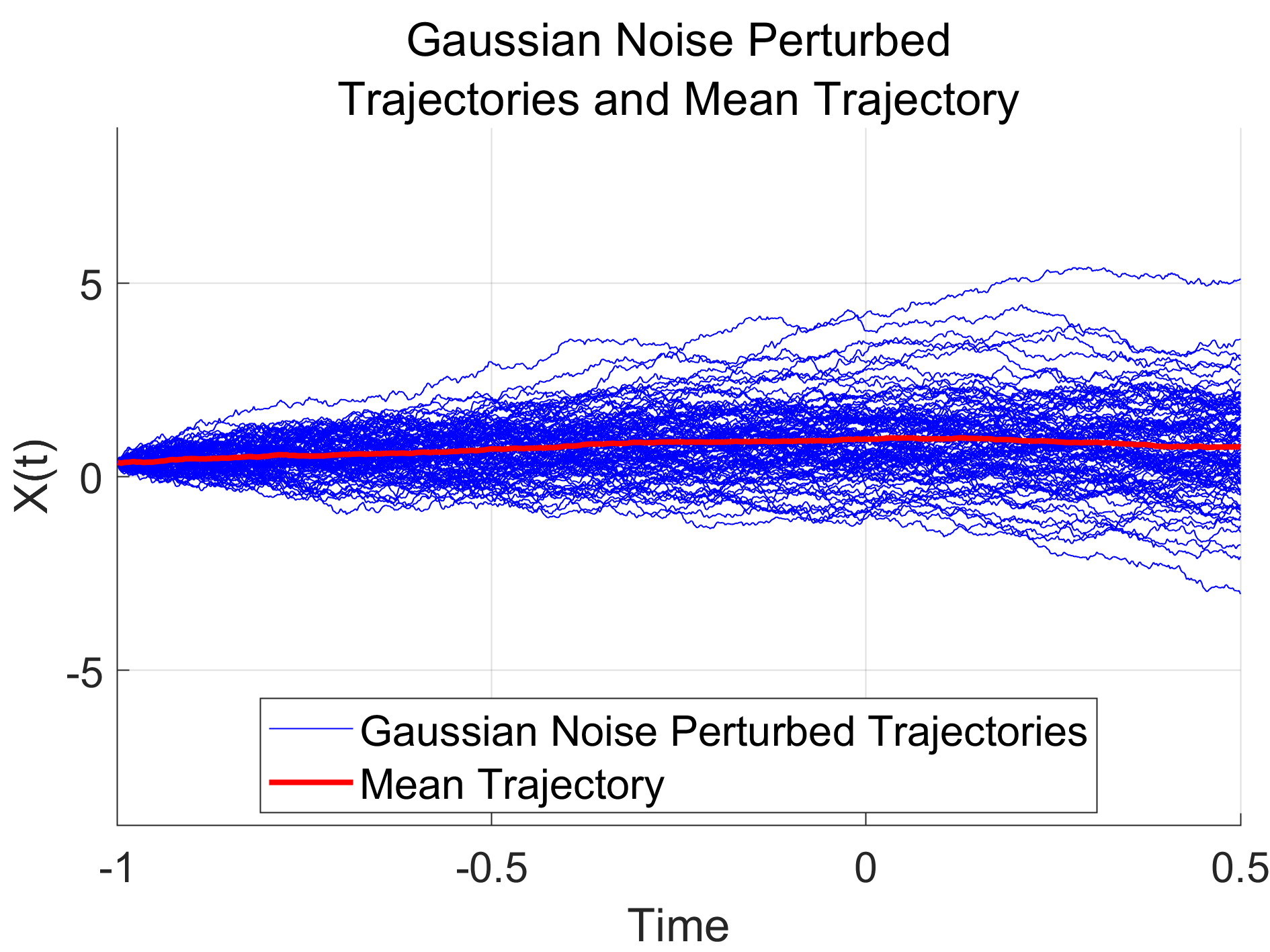}
            \caption*{(c1)$\lambda=1$}
        \end{minipage}\par
        \begin{minipage}{\textwidth}
            \includegraphics[width=\textwidth]{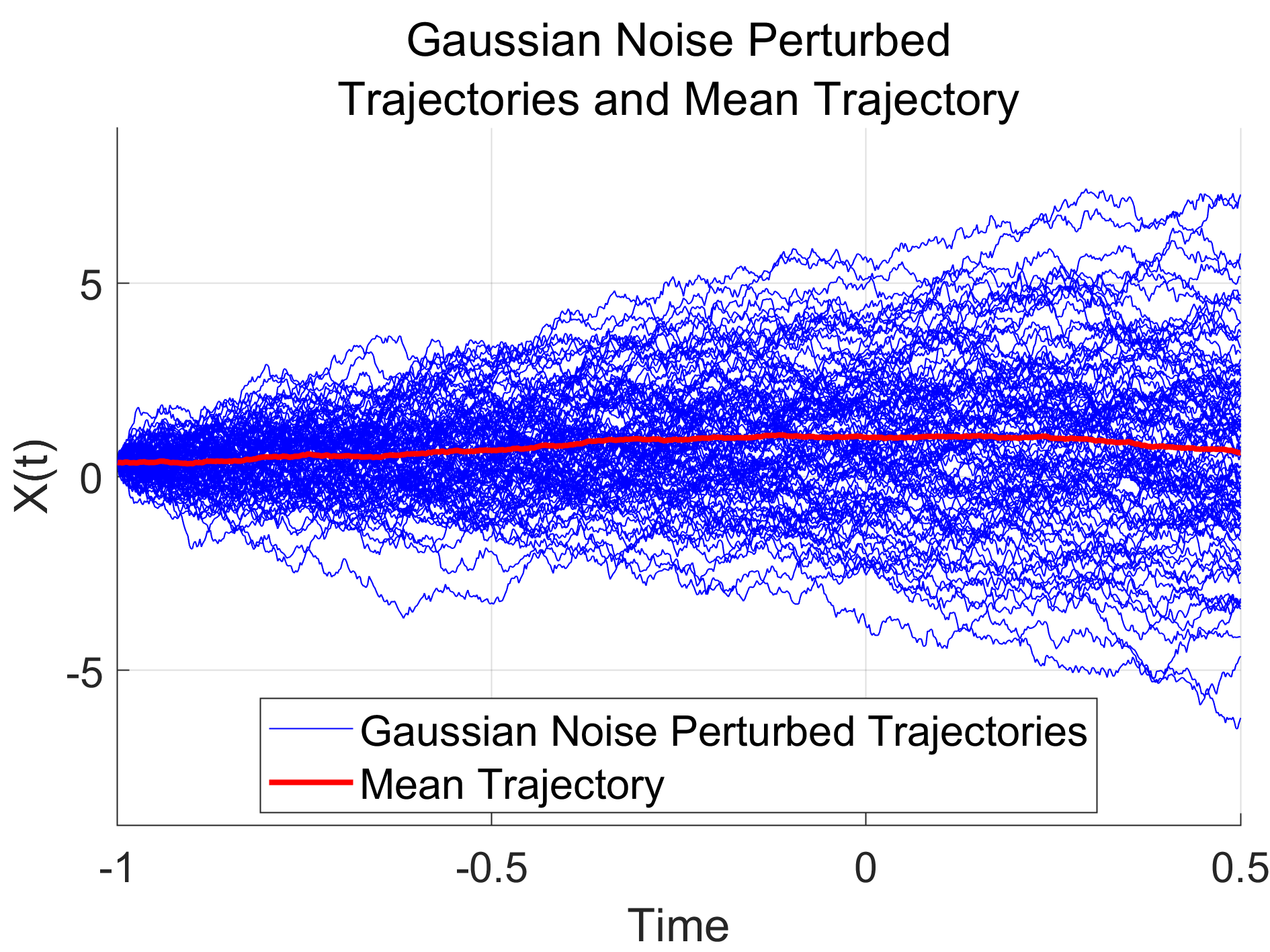}
            \caption*{(d1)$\lambda=2$}
        \end{minipage}
    \end{minipage}%
    \begin{minipage}{0.33\textwidth}
        \centering
        \begin{minipage}{\textwidth}
            \includegraphics[width=\textwidth]{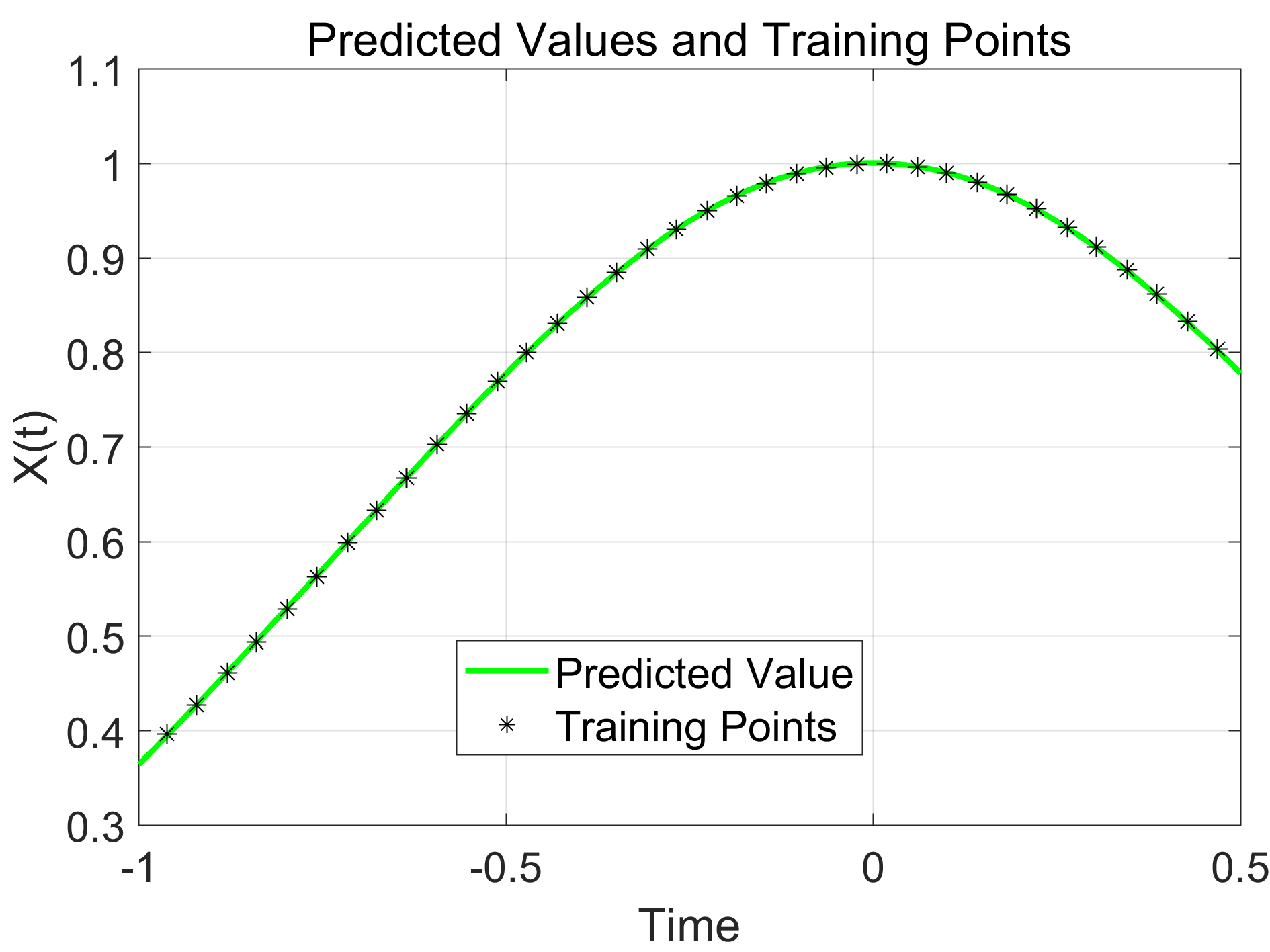}
            \caption*{(a2)$\lambda=0$}
        \end{minipage}\par
        \begin{minipage}{\textwidth}
            \includegraphics[width=\textwidth]{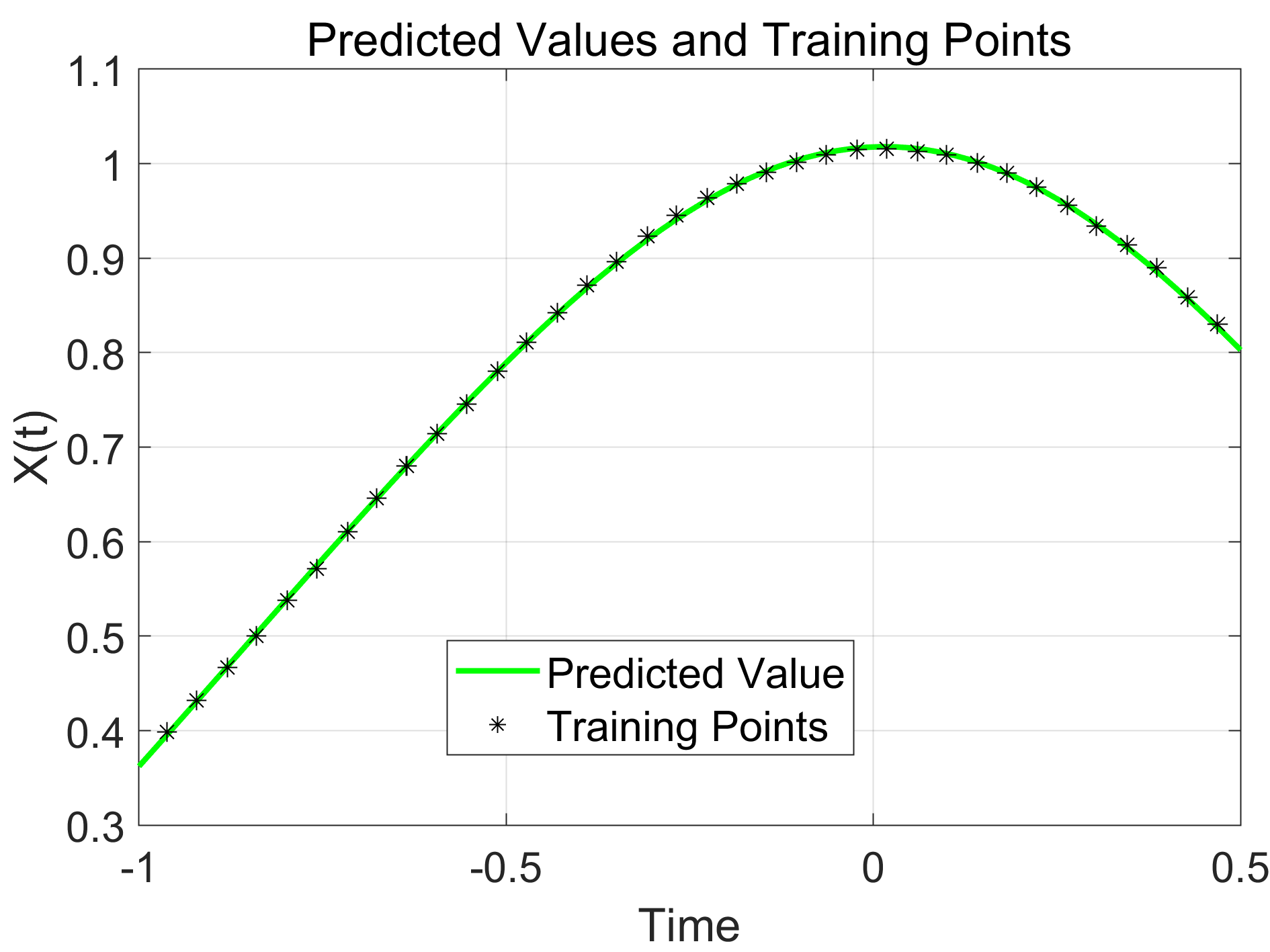}
            \caption*{(b2)$\lambda=0.1$}
        \end{minipage}\par
        \begin{minipage}{\textwidth}
            \includegraphics[width=\textwidth]{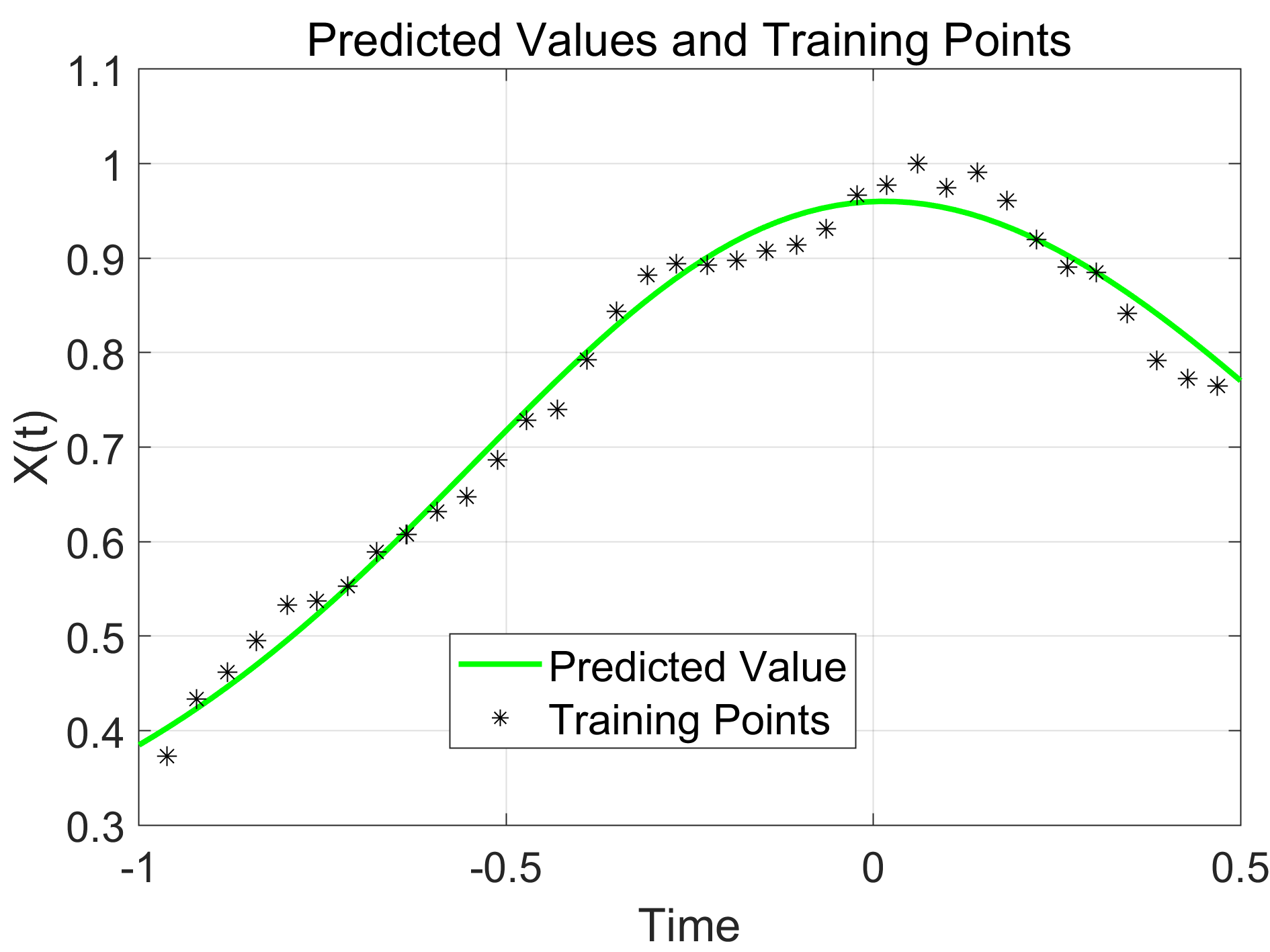}
            \caption*{(c2)$\lambda=1$}
        \end{minipage}\par
        \begin{minipage}{\textwidth}
            \includegraphics[width=\textwidth]{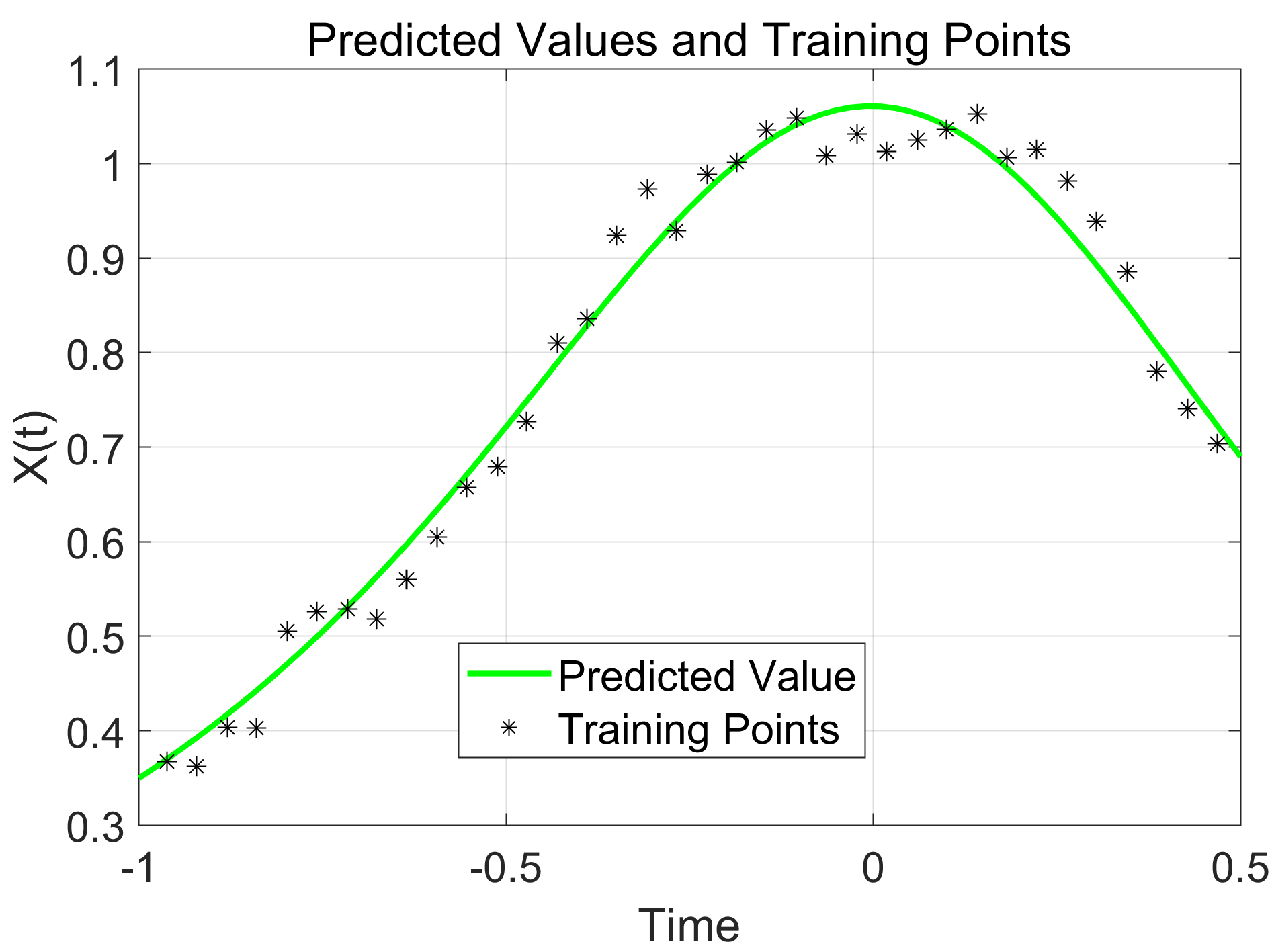}
            \caption*{(d2)$\lambda=2$}
        \end{minipage}
    \end{minipage}%
    \begin{minipage}{0.33\textwidth}
        \centering
        \begin{minipage}{\textwidth}
            \includegraphics[width=\textwidth]{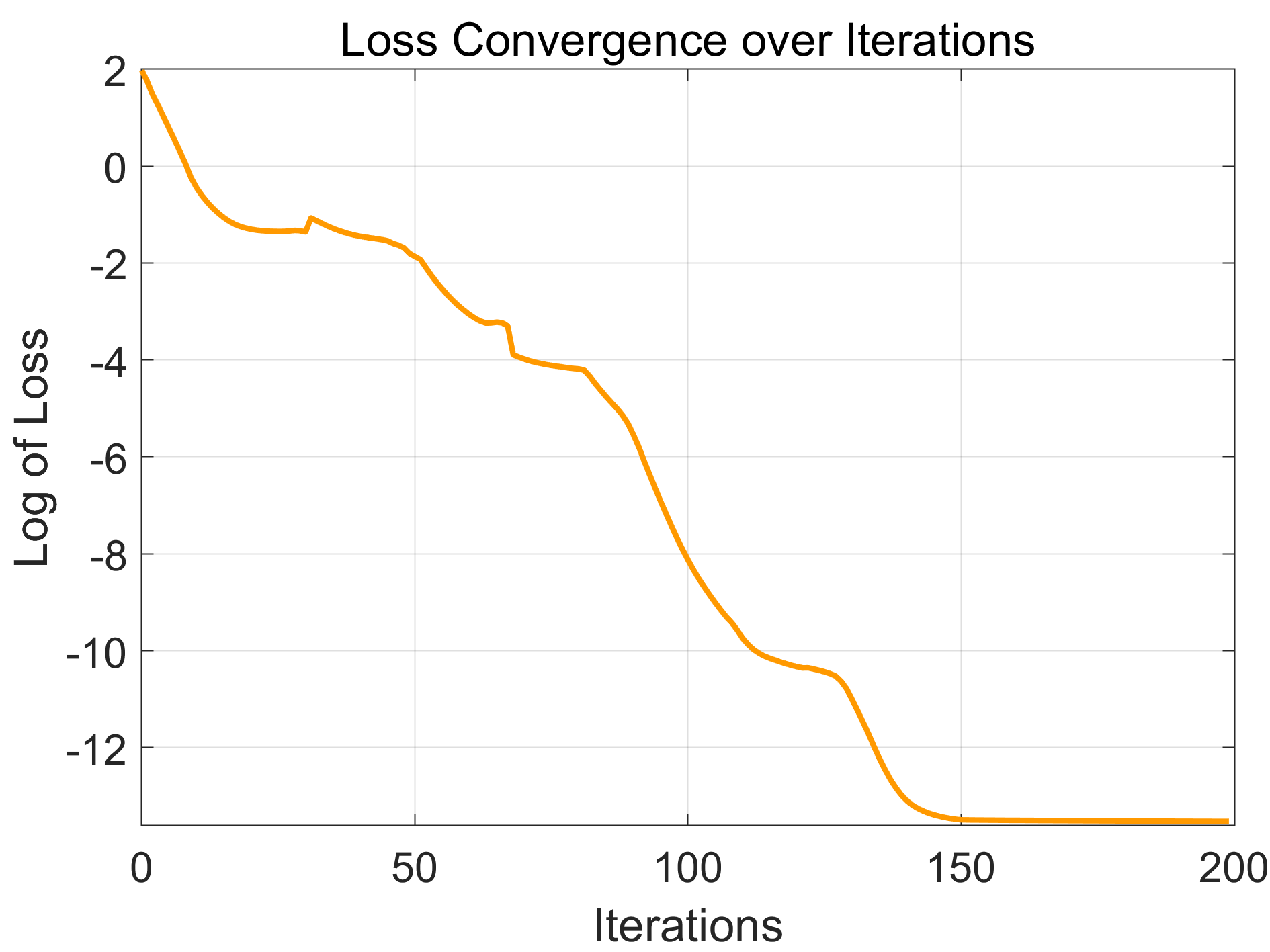}
            \caption*{(a3)$\lambda=0$}
        \end{minipage}\par
        \begin{minipage}{\textwidth}
            \includegraphics[width=\textwidth]{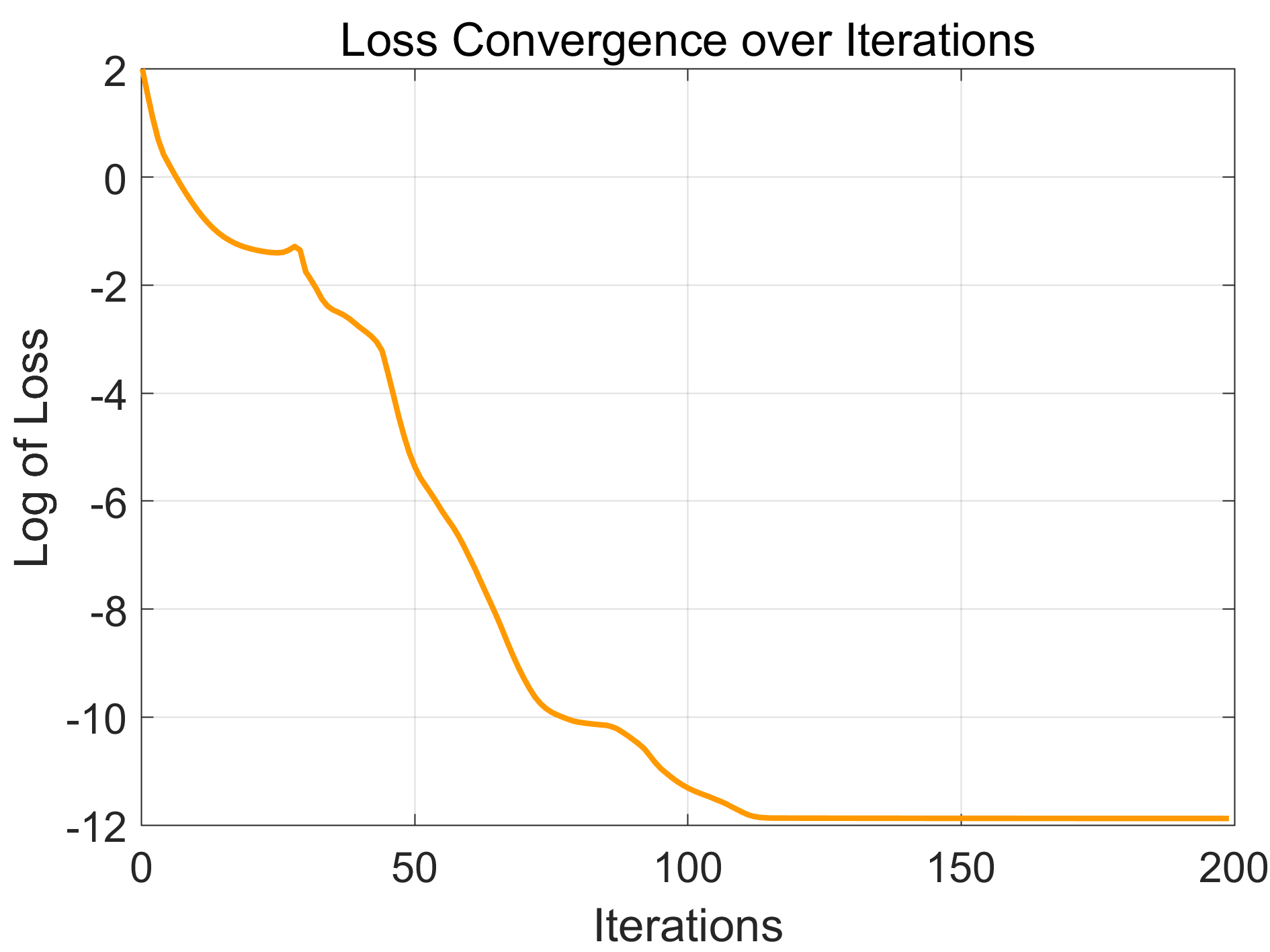}
            \caption*{(b3)$\lambda=0.1$}
        \end{minipage}\par
        \begin{minipage}{\textwidth}
            \includegraphics[width=\textwidth]{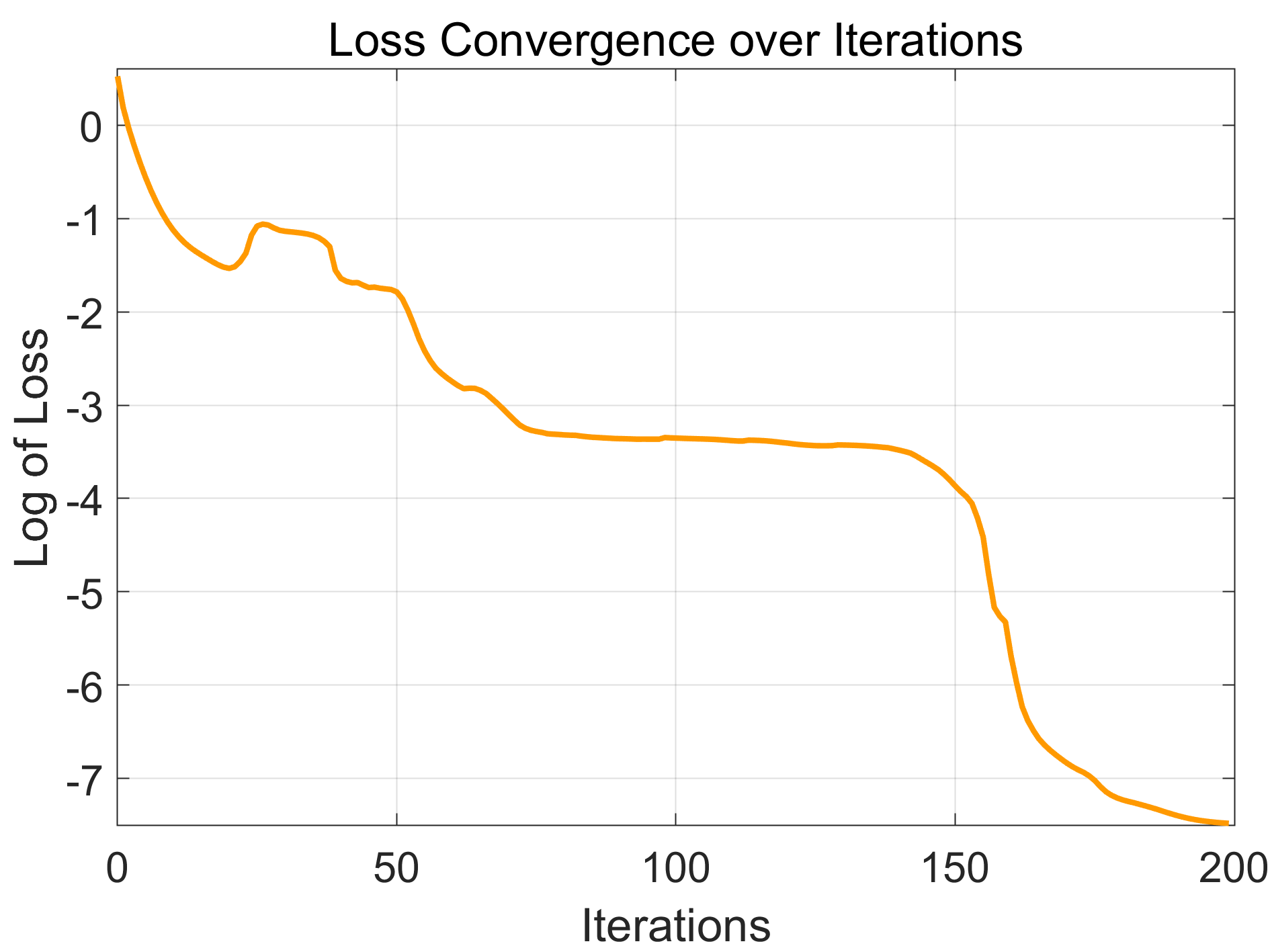}
            \caption*{(c3)$\lambda=1$}
        \end{minipage}\par
        \begin{minipage}{\textwidth}
            \includegraphics[width=\textwidth]{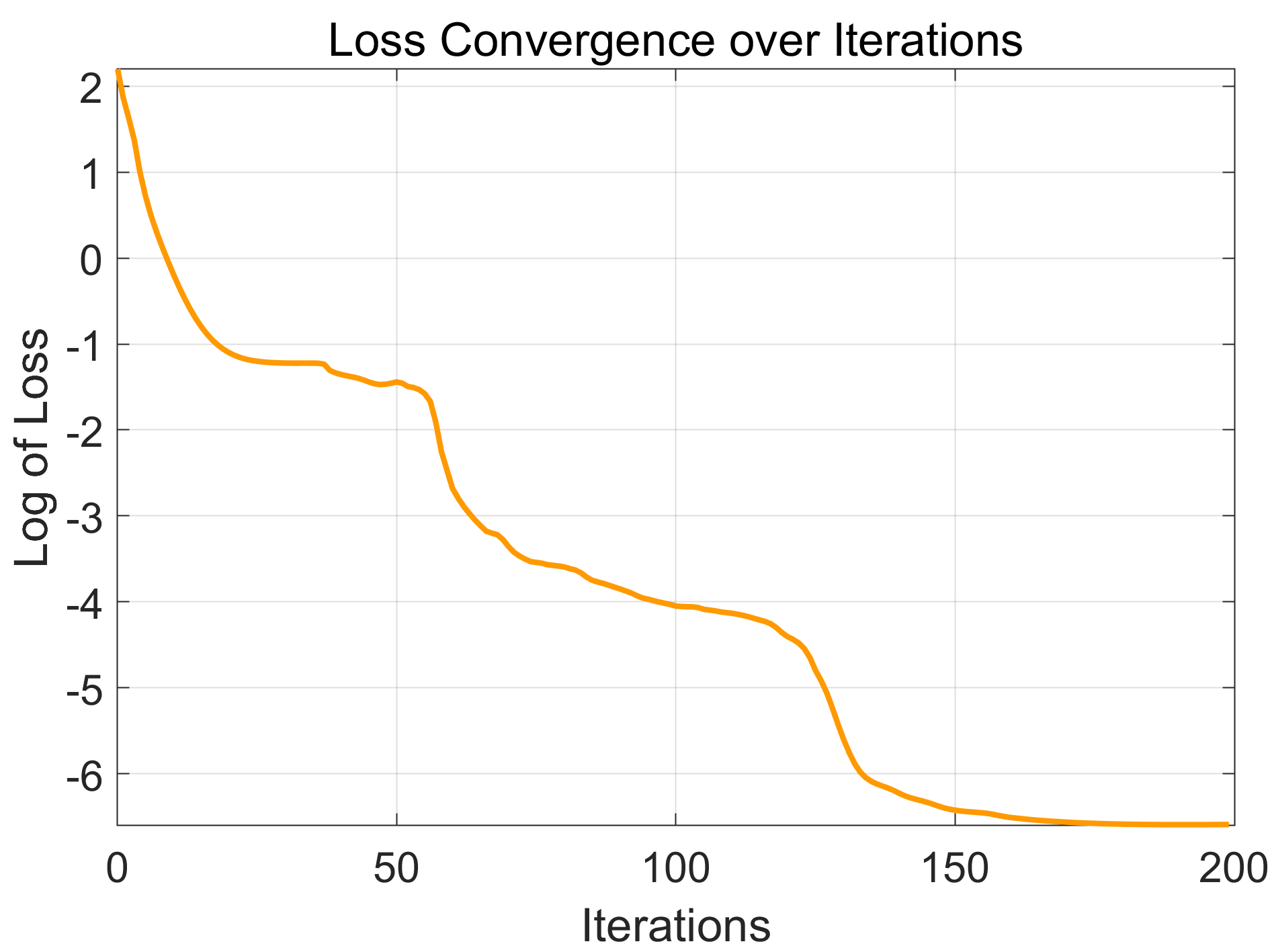}
            \caption*{(d3)$\lambda=2$}
        \end{minipage}
    \end{minipage}
    \caption{$\textbf{Case 3:}$ The length of the integration interval is $\frac{3}{2}$, with a sampling interval of $3/2000$ and a total of $50$ measurement data points. $\lambda$ is the Gaussian noise level coefficient. Blue line - the Gaussian noise perturbed trajectories, Red line - the mean trajectory. Green line - the improved DNNs predicted value, Black star - the true solution value. The Orange line illustrates the loss convergence over iterations.}
    \label{fig:3x4_case3}
\end{figure}

\subsection{Case 3}
In the third experiment, the equation is expressed as
\begin{equation}
    X(t)=e^{-t^2}+\frac{1}{2}te^{-1}-\frac{1}{2}te^{-t^2} - \int_{0}^{t+1} 
    \theta t(s-1)X(s-1)\:\mathrm{d}s + \int_{0}^{t+1} \lambda \:\mathrm{d}B_s, \qquad t\in [-1,\frac{1}{2}].
    \label{case3sto}
\end{equation}
In the absence of noise disturbances, the true solution of Eq.\,\ref{case3sto} is $e^{-t^2}$ when $\theta=1$. Similar to Case 2, the kernel function in Case 3 is nonlinear, with the variable $t$ constrained to the range $[-1,\frac{1}{2}]$ and the upper limit of the integral is $t+1$. The improved DNNs are used to address this problem, the main difference between Case 3 and Case 2 is that the true solution is $e^{-t^2}$, which is a higher-order function. This presents a greater challenge to the neural network's fitting capability. We introduce a primary output $u(t)$ to represent $X(t)$ and an auxiliary output $v(t)$ to represent the first integral in Eq.\,\ref{case3sto}, so the equation would be represented as
\begin{equation}
    \begin{aligned}
        &u(t) = e^{-t^2}+\frac{1}{2}te^{-1}-\frac{1}{2}te^{-t^2} - \theta\cdot v(t) + \lambda B_{t+1}, \qquad t\in [-1,\frac{1}{2}], \\
        &v(t)=\int_{0}^{t+1} \theta t(s-1)u(s-1)\:\mathrm{d}s, \\
        &v(-1)=0, \\
    \end{aligned}
\end{equation}
noting that $E(B_{t+1})=0$, the mean square error of residuals of the measurement data is
\begin{equation}
    \mathrm{MSE}_m=\frac{1}{N_m}\sum_{i=1}^{N_m}\left|u_{pred}(t_i^m;\gamma,\theta)- u_m(t_i^m)\right|^2,
\end{equation}
and again we simply assume that the collocation points coincide precisely with the measurement points, the mean square error of residuals of the governing equation is
\begin{equation}
    \mathrm{MSE}_g=\frac{1}{N_m}\sum_{i=1}^{N_m} \left|u_{pred}(t_i^m;\gamma,\theta)-[e^{-{t_i^m}^2}+\frac{1}{2}{t_i^m}e^{-1}-\frac{1}{2}{t_i^m}e^{-{(t_i^m)}^2}-\theta\cdot v_{pred}(t_i^m;\gamma,\theta)] \right|^2\,,
\end{equation}
the mean square error of residuals of the constraint between the primary output and the auxiliary output is represented as
\begin{equation}
    \mathrm{MSE}_o = \frac{1}{N_m} \sum_{i=1}^{N_m} \left| \frac{\partial v_{pred}(t_i^m; \gamma,\theta)}{\partial t} - {(t_i^m)}^2 * u_{pred}(t_i^m; \gamma,\theta) - \frac{1}{t_i^m}*v_{pred}(t_i^m; \gamma,\theta) \right|^2 \,,
\end{equation}
the mean square error of residuals of the initial condition of the auxiliary output is
\begin{equation}
    \mathrm{MSE}_i=\left| v_{pred}(-1;\gamma)-0 \right|^2.
\end{equation}
\par
The loss function $\mathrm{MSE}_{total}$ is expressed as
\begin{equation}
    \mathrm{MSE}_{total}=w_m\cdot \mathrm{MSE}_m + w_g\cdot \mathrm{MSE}_g + w_i\cdot \mathrm{MSE}_i + w_o\cdot \mathrm{MSE}_o.
\end{equation}
In Case 3, we employed the same adaptive weighting strategy as in Case 2, evenly sampling $50$ measurement data points from the expected values of $100$ simulated trajectories, using these as noise-free training data. By the improved DNNs, the value of $\theta$ converges to a solution that is reasonably close to the true value, demonstrating satisfactory performance after $200$ iterations. The relative error between the predicted and the true parameter values, and the absolute error between the predicted values and noise-free training data with different noise levels are shown in Table \ref{tab:comparison}, the simulated trajectories, the training data and the predicted values of $X(t)$, along with the convergence of the iterations are shown in Fig.\,\ref{fig:3x4_case3}.

\begin{table}[!h]
    \centering
    \begin{tabular}{cccccc}
        \toprule
        Case & Noise $\lambda$ & $\theta_{true}$ & $\theta_{pred}$ & Relative error in $\theta$ & Absolute error in $u(t)$ \\
        \midrule
        \multirow{4}{*}{Case 1} & 0   & 1.0000 & 1.0000   & 2.5988e-5 & 8.9899 \\
         & 1   & 1.0000 & 0.9931  & -6.8706e-3 & 9.2988 \\
         & 5   & 1.0000 & 0.9871  & -1.2934e-2 & 11.2232 \\
         & 20  & 1.0000 & 1.1475   & 1.4754e-1 & 18.0493 \\
        \midrule
        \multirow{4}{*}{Case 2} & 0   & 1.0000 & 0.9991   & -8.9019e-4 & 20.1727 \\
         & 0.1 & 1.0000 & 0.9986   & -1.3791e-3 & 20.1510 \\
         & 1   & 1.0000 & 1.0088   & 8.8235e-3 & 23.4212 \\
         & 2   & 1.0000 & 1.0302   & 3.0182e-2 & 26.4390 \\
        \midrule
        \multirow{4}{*}{Case 3} & 0   & 1.0000 & 1.0011   & 1.1481e-3 & 0.3789 \\
         & 0.1 & 1.0000 & 1.0912   & 9.1228e-2 & 0.3903 \\
         & 1   & 1.0000 & 1.2194   & 2.1941e-1 & 0.9764 \\
         & 2   & 1.0000 & 1.2530   & 2.5302e-1 & 1.4708 \\
        \bottomrule
    \end{tabular}
    \caption{Comparison of Gaussian noise level coefficient $\lambda$, true parameter $\theta_{true}$, the improved DNNs predicted parameter $\theta_{pred}$, relative error between the predicted and true parameter values, and absolute error between predicted values and noise-free training data for different experimental cases.}
    \label{tab:comparison}
\end{table}

\begin{table}[!h]
    \centering
    
    \begin{tabular}{ccccccc}
        \toprule
        Case & Depth & $\theta_{true}$ & $\theta_{pred}$ & Relative error in $\theta$ & Absolute error in $u(t)$  & Training time\\
        \midrule
        \multirow{4}{*}{Case 3} &  
           1 & 1.0000 & 1.4528 & 4.5277e-1 & 5.2930 & 25.1101\,s \\
         & 2 & 1.0000 & 1.2530 & 2.5302e-1 & 1.4708 & 27.7917\,s \\
         & 3 & 1.0000 & 1.1409 & 1.4091e-1 & 0.6596 & 33.4449\,s \\
         & 4 & 1.0000 & 1.0465 & 4.6529e-2 & 0.3843 & 56.2898\,s \\
        \bottomrule
    \end{tabular}
    \caption{Comparison of network depth for Case 3 under fixed noise level $\lambda=2$ and fixed hidden-layer width$=40$. The table lists the true parameter $\theta_{true}$, the predicted parameter $\theta_{pred}$, relative error between the predicted and true parameter values, absolute error between predicted values and noise-free training data, and training time. Only the number of hidden layers was varied, all other experimental conditions were held constant.}
    \label{tab:depth}
    
\end{table}

\paragraph{Effect of network depth and width}
To investigate the influence of network architecture on parameter identification and solution reconstruction for Case 3, we performed a controlled study in which only the network structure was varied while all other settings were kept fixed (noise level \(\lambda=2\), training procedure, optimizer, learning rate, data sampling and preprocessing). In the depth study we varied the number of hidden layers $(1, 2, 3, 4)$ with a fixed layer width of $40$. In the width study we varied the hidden-layer width $(10, 20, 40, 80)$ with a fixed depth of two hidden layers. Reported metrics are the relative error in $\theta$, the absolute error in $u(t)$, and training time. Results in Table \ref{tab:depth} and Table \ref{tab:width} show that a single hidden layer strongly underfits Case 3 (large parameter and solution errors). Increasing depth to three or four layers yields improved accuracy relative to the two-layer baseline, but the marginal gains are small compared with the rise in computational cost. Likewise, very narrow networks (width = $10$) underfit and intermediate width ($20$) can be unstable under the present hyperparameter choices, while a larger width ($80$) yields only modest improvements over the baseline width ($40$) at substantially greater training cost. These controlled experiments quantify the capacity–cost trade-off and support using a two hidden-layer network with width $40$ as a practical compromise for Case 3 under the present training protocol; if higher accuracy is required and computational resources permit, modest increases in depth or width may be employed together with more extensive hyperparameter tuning and regularization.

\begin{table}[!h]
    \centering
    
    \begin{tabular}{ccccccc}
        \toprule
        Case & Width & $\theta_{true}$ & $\theta_{pred}$ & Relative error in $\theta$ & Absolute error in $u(t)$  & Training time\\
        \midrule
        \multirow{4}{*}{Case 3} &  
           10 & 1.0000 & 1.6056 & 6.0566e-1 & 13.5727 & 24.5913\,s \\
         & 20 & 1.0000 & -0.8662 & 1.8662e+0 & 0.3427 & 27.0879\,s \\
         & 40 & 1.0000 & 1.2530 & 2.5302e-1 & 1.4708 & 27.7917\,s \\
         & 80 & 1.0000 & 1.2280 & 4.6529e-2 & 0.2759 & 38.1601\,s \\
        \bottomrule
    \end{tabular}
    \caption{Comparison of network width for Case 3 under fixed noise level $\lambda=2$ and fixed depth$=2$ hidden layers. The table lists the true parameter $\theta_{true}$, the predicted parameter $\theta_{pred}$, relative error between the predicted and true parameter values, absolute error between predicted values and noise-free training data, and training time. Only the hidden-layer width was varied, all other experimental conditions were held constant.}
    \label{tab:width}
    
\end{table}

\subsection{Prediction experiments}
\begin{figure}[htbp]
    \centering
    \begin{minipage}{0.33\textwidth}
        \centering
        \begin{minipage}{\textwidth}
            \includegraphics[width=\textwidth]{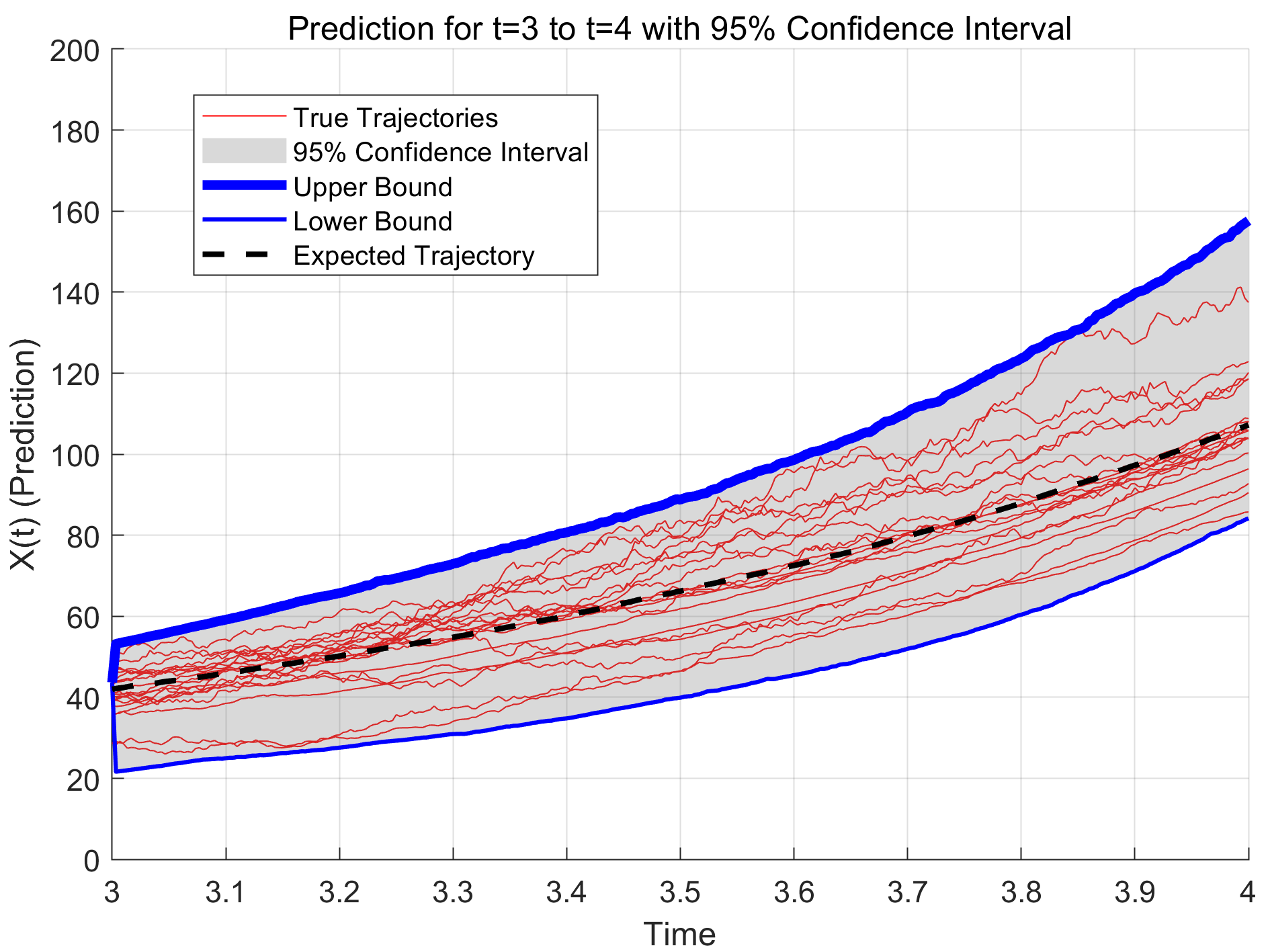}
            \caption*{(a1)$\lambda=5$}
        \end{minipage}\par
    \end{minipage}%
    \begin{minipage}{0.33\textwidth}
        \centering
        \begin{minipage}{\textwidth}
            \includegraphics[width=\textwidth]{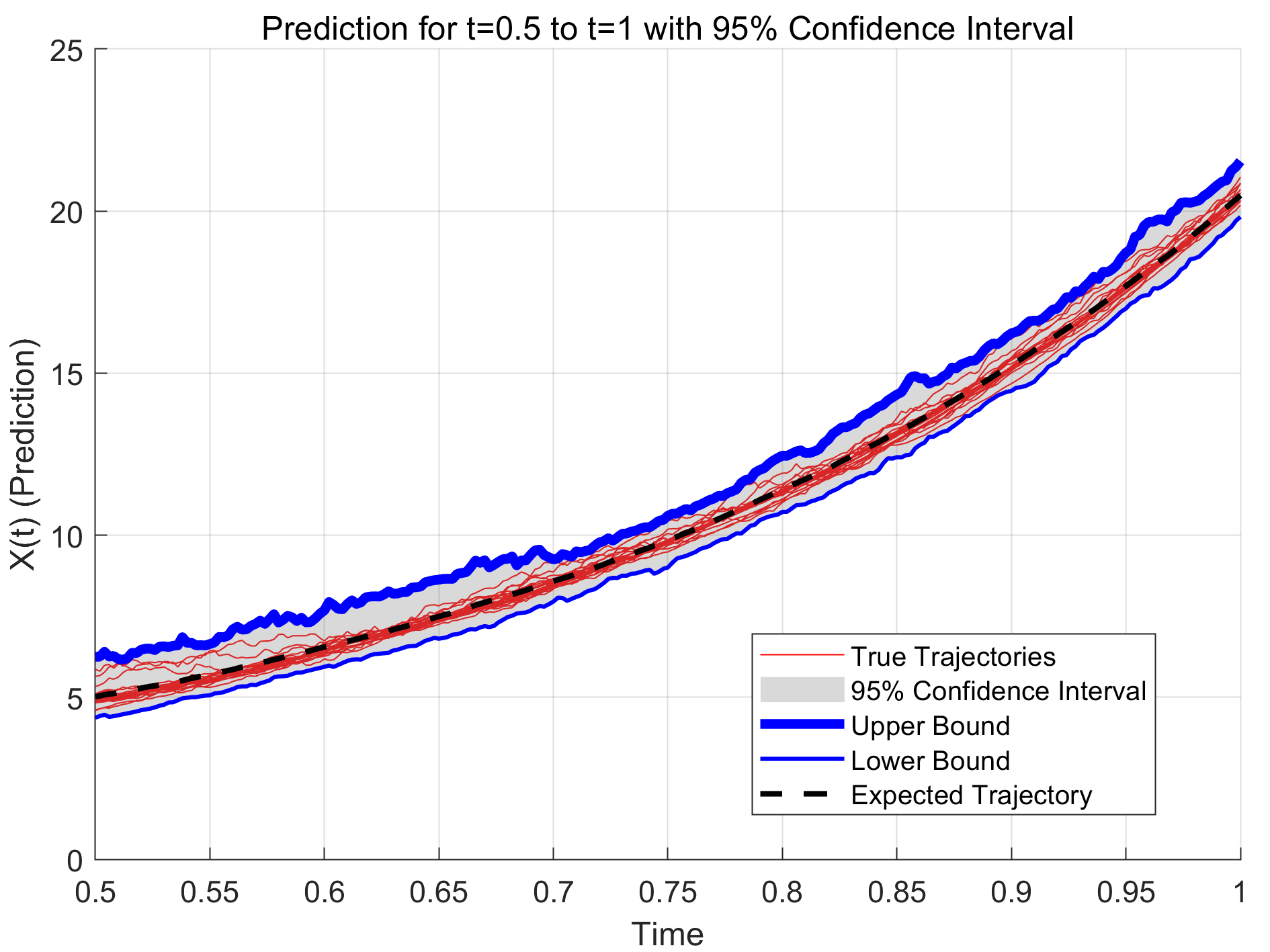}
            \caption*{(a2)$\lambda=1$}
        \end{minipage}\par
    \end{minipage}%
    \begin{minipage}{0.33\textwidth}
        \centering
        \begin{minipage}{\textwidth}
            \includegraphics[width=\textwidth]{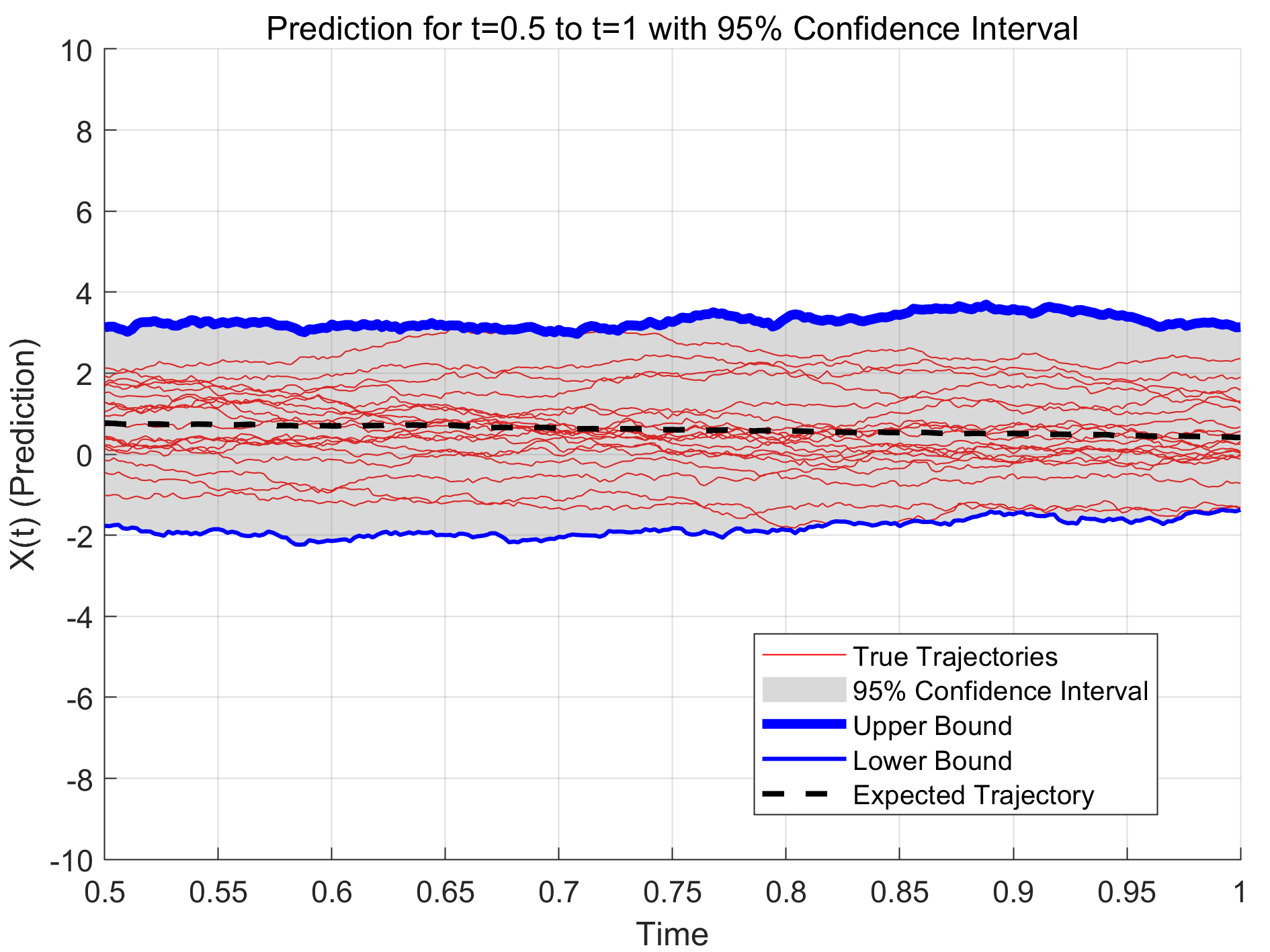}
            \caption*{(a3)$\lambda=1$}
        \end{minipage}\par
    \end{minipage}
    \caption{$\textbf{Prediction:}$  In Case 1, the Gaussian noise level coefficient $\lambda$ is set to $5$, and the prediction interval is from $t=3$ to $t=4$. In Case 2 and Case 3, the Gaussian noise level coefficient $\lambda$ is set to $1$, and the prediction intervals span from $t=0.5$ to $t=1$. The values of $\lambda$ are chosen based on the reconstruction experiments, where these noise levels were found to have an appropriate impact on the model's performance.}
    \label{fig:pred}
\end{figure}

To validate the proposed methodology, we conduct numerical experiments for both linear and nonlinear kernel functions, demonstrating that the improved DNNs can effectively estimate parameters even under varying levels of Gaussian noise. Following the parameter identification phase, the experiment aims to validate the performance of the reconstructed Volterra integral equations by extending the time horizon beyond the reconstruction period. The prediction performance is validated by comparing the predicted trajectories with the true trajectories, using a $95\%$ confidence interval for the comparison. This confidence interval is constructed by substituting the parameters $\theta$ obtained from the reconstruction experiment into the integral equation, simulating $1000$ random trajectories, and sampling over an integration interval divided into $250$ equal parts. Specifically, $20$ random trajectories from the true data are compared against the predicted trajectories, and if all of these trajectories fall within the confidence interval, we consider both the parameter identification and prediction tasks to have been successfully addressed. This approach not only confirms the accuracy of the parameter estimates but also demonstrates the robustness of the improved DNNs framework in handling predictions over extended time horizons. The experimental results are shown in Fig.\,\ref{fig:pred}.

The results show that the true trajectories fall within the $95\%$ confidence interval of the simulated trajectories, thereby confirming the effectiveness of the parameter identification process. These findings support the robustness of the model and its applicability for forecasting in extended time periods.

\section{Discussion and Conclusion}

\paragraph{Discussion: effect of noise on parameter identification}
Table~\ref{tab:comparison} shows that parameter estimation error increases markedly
as the Gaussian noise coefficient \(\lambda\) becomes large. This behavior is expected.
When the observational noise is comparable to or dominates the deterministic drift,
the data carry less information about the drift parameter; consequently the estimator
tends to display larger variance and possibly bias.

The measurement process may be viewed as \(y(t)=u(t;\theta)+\eta(t)\), where \(u(\cdot;\theta)\)
is the deterministic component determined by \(\theta\) and \(\eta\) is the stochastic perturbation.
As the magnitude of \(\eta\) increases (lower signal-to-noise ratio), the parameter estimation
problem becomes effectively more ill-conditioned and estimation uncertainty increases. Pragmatic strategies to mitigate this effect include:

\begin{itemize}
  \item \textbf{Multi-trajectory averaging.} Averaging observations across independent realizations
    reduces the effective noise variance and improves identification.
  \item \textbf{Explicit noise modeling and regularization.} Jointly modeling the noise (or placing
    informative priors on parameters) and adding regularization stabilizes estimation under high noise.
  \item \textbf{Uncertainty quantification.} Reporting parameter standard errors or credible intervals
    makes the increased estimation uncertainty explicit and aids interpretation.
\end{itemize}

We defer implementation and systematic numerical study of these mitigation strategies to future work.

\paragraph{Conclusion}
In this paper, we have proposed a new DNNs framework to learn the unknown parameters of the drift term in Volterra integral equations perturbed by Gaussian noise. In the improved DNNs, a multi-output neural network is designed to concurrently represent both the primary variables and the integrals involved in the governing equations. To address the parameter identification problem under perturbations, we extend the basic DNNs framework by leveraging the multi-output feature. We establish connections between the various outputs of the neural network and incorporate these constraints into the loss function. This approach enhances the network's convergence during the iterative process, leading to more accurate predictions. The improved DNNs demonstrate excellent performance in solving the parameter identification problem for perturbed Volterra integral equations, indicating their ability to handle a broader range of problems, extending beyond deterministic equations.

Through a series of numerical experiments, the improved DNNs have been shown to effectively solve the parameter identification problem for perturbed Volterra integral equations with both linear and nonlinear kernel functions. The experimental results indicate that the improved DNNs exhibit superior performance with linear kernel functions, showcasing their robustness and versatility in handling different problem settings. Additionally, we have validated the performance of the reconstructed equations by predicting the evolution of the system beyond the integration interval. The predicted trajectories, compared with the true trajectories, confirm the model's capability in handling long-term predictions, demonstrating its robustness in extended time horizons.

While the improved DNNs have demonstrated promising results, there remain areas for further investigation and refinement. For example, the complexity of the true solution in undisturbed cases appears to influence the accuracy of the results, particularly in cases where the true solution has a higher order. Future work will focus on addressing these challenges and exploring more advanced techniques for improving the accuracy of the model in the presence of highly complex true solutions. Additionally, extending the framework to handle other types of perturbations and exploring its performance in higher-dimensional problems will be important directions for future research.

\appendix
\section{Proof of Theorem \ref{thm:convergence}}
\label{app:proof_convergence}

\subsection*{Preliminaries}
Let \(u^\star\) be the true solution of Eq.\,\ref{disturbed} and \(v^\star(t) = \int_0^t k(t,s)u^\star(s)ds\) be the corresponding integral term. Define the error functions:
\[
e_u(t) = u_{\theta^*}(t) - u^\star(t), \quad e_v(t) = v_{\theta^*}(t) - v^\star(t).
\]

The residuals are defined as:
\begin{align*}
R_g(t) &= u_{\theta^*}(t) - \mathbb{E}[f(t) + \theta^* v_{\theta^*}(t) + \lambda B_t] \\
R_o(t) &= \frac{\partial v_{\theta^*}}{\partial t} - k(t,t)u_{\theta^*}(t) - \int_0^t \frac{\partial k(t,s)}{\partial t} u_{\theta^*}(s) ds \\
R_m(t_i) &= u_{\theta^*}(t_i) - u_m(t_i)
\end{align*}

\subsection*{Step 1: Conditional Stability Estimate}

Following the approach in \cite{mishra2022estimates} for inverse problems, we establish a conditional stability estimate for the Volterra integral equation. Under Assumption \ref{ass:wellposed}, there exists a constant \(C_{\text{stab}} > 0\) such that:
\begin{lem}[Stability Estimate]
\label{lem:stability}
For the Eq.\,\ref{disturbed}, the following stability bound holds:
\[
\| e_u \|_{L^2(0,T)} \leq C_{\text{stab}} \left( \| R_g \|_{L^2(0,T)} + \| R_o \|_{L^2(0,T)} + \| R_m \|_{L^2_{\text{data}}} \right)
\]
where \(\| \cdot \|_{L^2_{\text{data}}}\) denotes the \(L^2\)-norm over the measurement points.
\end{lem}

\subsection*{Step 2: Neural Network Approximation}

Under Assumption \ref{ass:approx}, there exists a network \((\tilde{u}, \tilde{v}) \in \mathcal{N}\) such that:
\[
\| \tilde{u} - u^\star \|_{L^2} + \| \tilde{v} - v^\star \|_{L^2} \leq \varepsilon_{\text{app}}
\]
This follows from the neural network approximation theory for Sobolev spaces, as established in \cite{qian2023physics} (Lemma A.3).

\medskip

\subsection*{Step 3: Training Error to Residual Bound}

The discrete training loss controls the continuous residuals up to quadrature error:

\begin{lem}[Residual Control]
\label{lem:residual}
There exists a constant \(C_{\text{disc}} > 0\) such that:
\[
\| R_g \|_{L^2(0,T)}^2 + \| R_o \|_{L^2(0,T)}^2 + \| R_m \|_{L^2_{\text{data}}}^2 \leq C_{\text{disc}} \left( \mathcal{L}_{\text{train}}(\theta^*) + \delta_{\text{quad}}^2 \right)
\]
where \(\delta_{\text{quad}}\) bounds the numerical integration error in the output condition.
\end{lem}

\begin{proof}
This follows by:
\begin{enumerate}
    \item Relating discrete MSE to continuous \(L^2\) norms via quadrature error estimates
    \item Using the regularity of network outputs (bounded derivatives due to tanh activation)
    \item Applying the numerical integration error bound for the output condition integral
\end{enumerate}
The multi-term structure is handled similarly to \cite{qian2023physics} Theorem 3.5.
\end{proof}

\subsection*{Step 4: Parameter Identifiability}

For parameter convergence, we require the identification condition that the operator:
\[
\mathcal{F}(\theta) = \int_0^t k(t,s;\theta)X(s)ds
\]
is injective in a neighborhood of \(\theta_{\text{true}}\). Under this condition, standard inverse problem theory \cite{mishra2022estimates} guarantees that solution convergence implies parameter convergence.

\subsection*{Final Combination}

Combining Lemma \ref{lem:exist-uniq}, Lemma \ref{lem:stability} and Lemma \ref{lem:residual}:
\[
\| e_u \|_{L^2} \leq C_{\text{stab}} C_{\text{disc}}^{1/2} \left( \sqrt{\mathcal{L}_{\text{train}}(\theta^*)} + \delta_{\text{quad}} \right)
\]
Adding the approximation error and setting \(C = \max(C_{\text{stab}} C_{\text{disc}}^{1/2}, 1)\) yields the solution convergence result.

The parameter convergence follows similarly, with constant \(C_\theta\) depending on the identification condition.

This completes the appendix proof.

\section*{Acknowledgements}
This work was supported by the NSFC grant 12371198.


\begin{thebibliography}{0}
\bibitem{ahmed2015parameter}
S. Ahmed, Parameter and delay estimation of fractional order models from step response, {\it IFAC-PapersOnLine}
{\bf 48} (2015) 942--947.

\bibitem{fermo2016scattering}
L. Fermo, C. van der Mee and S. Seatzu,
Scattering data computation for the Zakharov-Shabat system,
{\it Calcolo}
{\bf 53} (2016) 487--520.

\bibitem{mohammadian2018integral}
B. Mohammadian, R. Mohammad and S. Saramad,
Integral-equation based methods for parameter estimation in output pulses of radiation detectors: Application in nuclear medicine and spectroscopy,
{\it Nucl. Instrum. Methods Phys. Res. Sect. A}
{\bf 887} (2018) 7--12.

\bibitem{song2005integral}
Z. Song,
Integral-equation-based continuous-time model identification of a magnetostrictive actuator,
{\it IEE Proc.-Control Theory Appl.}
{\bf 152} (2005) 85--89.

\bibitem{li2024parameter}
A. Li and Y. Xia,
Parameter estimation of uncertain differential equations with estimating functions,
{\it Soft Comput.}
{\bf 28} (2024) 77--86.

\bibitem{li2022numerical}
M. Li, C. Huang and Y. Hu,
Numerical methods for stochastic Volterra integral equations with weakly singular kernels,
{\it IMA J. Numer. Anal.}
{\bf 42} (2022) 2656--2683.

\bibitem{ref4}
M. A. Berger and V. J. Mizel,
Volterra equations with It\^{o} integrals---I,
{\it J. Integral Equations}
(1980) 187--245.

\bibitem{17}
K. It\^{o},
{\it On Stochastic Differential Equations},
Vol. 4 (American Mathematical Soc., 1951).

\bibitem{dingxing2}
T. A. Burton,
{\it Volterra Integral and Differential Equations}
(Elsevier, 2005).

\bibitem{dingxing3}
M. Islam and J. T. Neugebauer,
Qualitative properties of nonlinear Volterra integral equations,
{\it Electron. J. Qual. Theory Differ. Equ.}
{\bf 12} (2008).

\bibitem{shuzhi8}
A. Pedas and G. Vainikko,
Smoothing transformation and piecewise polynomial collocation for weakly singular Volterra integral equations,
{\it Computing}
{\bf 73} (2004) 271--293.

\bibitem{shuzhi16}
D. Hou, Y. Lin, M. Azaiez and C. Xu,
A M\"{u}ntz-collocation spectral method for weakly singular Volterra integral equations,
{\it J. Sci. Comput.}
{\bf 81} (2019) 2162--2187.

\bibitem{shuzhi17}
S. Chen, J. Shen and L. Wang,
Generalized Jacobi functions and their applications to fractional differential equations,
{\it Math. Comp.}
{\bf 85} (2016) 1603--1638.

\bibitem{canshu1}
D. Conte, G. Pagano and B. Paternoster,
Two-step peer methods with equation-dependent coefficients,
{\it Comput. Appl. Math.}
{\bf 41} (2022) 140.

\bibitem{canshu2}
I. Dattner, H. Ship and E. O. Voit,
Separable nonlinear least-squares parameter estimation for complex dynamic systems,
{\it Complexity}
{\bf 2020} (2020) 6403641.

\bibitem{canshu3}
H. Wu,
Statistical methods for HIV dynamic studies in AIDS clinical trials,
{\it Stat. Methods Med. Res.}
{\bf 14} (2005) 171--192.

\bibitem{canshu4}
W. Bradley and F. Boukouvala,
Two-stage approach to parameter estimation of differential equations using neural odes,
{\it Ind. Eng. Chem. Res.}
{\bf 60} (2021) 16330--16344.

\bibitem{boulier2014algorithm}
F. Boulier, A. Korporal, F. Lemaire, W. Perruquetti, A. Poteaux and R. Ushirobira,
An algorithm for converting nonlinear differential equations to integral equations with an application to parameter estimation from noisy data,
in {\it International Workshop on Computer Algebra in Scientific Computing}
(2014) 28--43.

\bibitem{pinn2}
M. Raissi, P. Perdikaris and G. E. Karniadakis,
Physics informed deep learning (part i): Data-driven solutions of nonlinear partial differential equations,
{\it arXiv preprint arXiv:1711.10561}
(2017).

\bibitem{pinn20}
L. Yuan, Y. Ni, X. Deng and S. Hao,
A-PINN: Auxiliary physics informed neural networks for forward and inverse problems of nonlinear integro-differential equations,
{\it J. Comput. Phys.}
{\bf 462} (2022) 111260.

\bibitem{pinn28}
L. Yang, D. Zhang and G. E. Karniadakis,
Physics-informed generative adversarial networks for stochastic differential equations,
{\it SIAM J. Sci. Comput.}
{\bf 42} (2020) A292--A317.

\bibitem{pinn29}
D. Zhang, L. Lu, L. Guo and G. E. Karniadakis,
Quantifying total uncertainty in physics-informed neural networks for solving forward and inverse stochastic problems,
{\it J. Comput. Phys.}
{\bf 397} (2019) 108850.

\bibitem{pinn32}
D. Zhang, L. Guo and G. E. Karniadakis,
Learning in modal space: Solving time-dependent stochastic PDEs using physics-informed neural networks,
{\it SIAM J. Sci. Comput.}
{\bf 42} (2020) A639--A665.

\bibitem{pinn1}
M. Raissi, P. Perdikaris and G. E. Karniadakis,
Physics-informed neural networks: A deep learning framework for solving forward and inverse problems involving nonlinear partial differential equations,
{\it J. Comput. Phys.}
{\bf 378} (2019) 686--707.

\bibitem{pinn4}
S. Cai, Z. Mao, Z. Wang, M. Yin and G. E. Karniadakis,
Physics-informed neural networks (PINNs) for fluid mechanics: A review,
{\it Acta Mech. Sin.}
{\bf 37} (2021) 1727--1738.

\bibitem{pinn9} 
A. Arzani, J. Wang and R. M. D'Souza,
Uncovering near-wall blood flow from sparse data with physics-informed neural networks,
{\it Phys. Fluids}
{\bf 33} (2021).

\bibitem{pinn10}
S. Cai, Z. Wang, S. Wang, P. Perdikaris and G. E. Karniadakis,
Physics-informed neural networks for heat transfer problems,
{\it J. Heat Transfer}
{\bf 143} (2021) 060801.

\bibitem{pinn11}
D. Jalili, S. Jang, M. Jadidi, G. Giustini, A. Keshmiri and Y. Mahmoudi,
Physics-informed neural networks for heat transfer prediction in two-phase flows,
{\it Int. J. Heat Mass Transfer}
{\bf 221} (2024) 125089.

\bibitem{pinn3}
A. G. Baydin, B. A. Pearlmutter, A. A. Radul and J. M. Siskind,
Automatic differentiation in machine learning: a survey,
{\it J. Mach. Learn. Res.}
{\bf 18} (2018) 1--43.

\bibitem{pinn34}
X. Chen, J. Duan, J. Hu and D. Li,
Data-driven method to learn the most probable transition pathway and stochastic differential equation,
{\it Physica D}
{\bf 443} (2023) 133559.

\bibitem{oksendal2013stochastic}
B. Oksendal,
{\it Stochastic Differential Equations: An Introduction with Applications}
(Springer, 2013).




\bibitem{ito1979existence}
I. It\^{o},
On the existence and uniqueness of solutions of stochastic integral equations of the Volterra type,
{\it Kodai Math. J.}
{\bf 2} (1979) 158--170.

\bibitem{jin2022existence}
S. Jin,
Existence and uniqueness of solutions for the stochastic Volterra-Levin equation with variable delays,
{\it Open Math.}
{\bf 20} (2022) 769--782.

\bibitem{fdm}
J. W. Thomas,
{\it Numerical Partial Differential Equations: Finite Difference Methods},
Vol. 22 (Springer, 2013).

\bibitem{mishra2022estimates}
S. Mishra and R. Molinaro,
Estimates on the generalization error of physics-informed neural networks for approximating a class of inverse problems for PDEs,
{\it IMA J. Numer. Anal.}
{\bf 42} (2022) 981--1022.

\bibitem{qian2023physics}
Y. Qian, Y. Zhang, Y. Huang and S. Dong,
Physics-informed neural networks for approximating dynamic (hyperbolic) PDEs of second order in time: Error analysis and algorithms,
{\it J. Comput. Phys.}
{\bf 495} (2023) 112527.

\end{thebibliography}


\end{document}